\documentclass[10pt,onecolumn,letterpaper]{article}

\usepackage{geometry}
\usepackage{times}
\usepackage{epsfig}
\usepackage{graphicx}
\usepackage{amsmath}
\usepackage{amssymb}

\usepackage{subfigure}
\usepackage{paralist}
\usepackage{xcolor}
\usepackage{booktabs}  
\usepackage{multirow} 
\usepackage{multicol} 
\usepackage{arydshln}
\usepackage{wrapfig}
\usepackage{comment}
\usepackage{natbib}

\usepackage[linesnumbered,ruled,lined]{algorithm2e}

\newtheorem{proposition}{Proposition}

\newtheorem{definition}{Definition}
\newtheorem{assumption}{Assumption}
\newtheorem{corollary}{Corollary}

\newsavebox\CBox
\def\textBF#1{\sbox\CBox{#1}\resizebox{\wd\CBox}{\ht\CBox}{\textbf{#1}}}

\usepackage[pagebackref=true,breaklinks=true,letterpaper=true,colorlinks,bookmarks=false,linkcolor=black,citecolor=blue]{hyperref}

\begin{document}
	
	\title{Supplemental Material~: Lifelong Generative Modelling Using Dynamic Expansion Graph Model}
	\author{
		Fei Ye and Adrian G. Bors
		\\
		Department of Computer Science, University of York, York YO10 5GH, UK\\
		fy689@york.ac.uk, adrian.bors@york.ac.uk
	}

	\maketitle

	\setlength{\textfloatsep}{8pt}
	\setlength{\abovedisplayskip}{8pt} 
	\setlength{\belowdisplayskip}{8pt}
	\setlength{\abovecaptionskip}{8pt}

	\appendix
	\renewcommand{\appendixname}{Appendix~\Alph{section}}

	\tableofcontents
	
	\clearpage
	
	\section{The proof of Theorem 1}
	
	\begin{assumption}
		Let $\mathcal{X}$ be a metric space that satisfies $\mathcal{L}(a,b) \le \mathcal{L}(a,c) + \mathcal{L}(c,b) $ where the loss function $\mathcal{L}(\cdot)$ is a metric and $a, b, c \in \mathcal{X}$.
		\label{assumuption1}
	\end{assumption}
	
	Based on Assumption~\ref{assumuption1}, we provide the detailed proof as follows~:
	\vspace{10pt}
	
	\textBF{Proof.}
	Let $\mathcal{P}_i$ and ${\tilde {\mathcal{P}}}_i$ be two domains over $\mathcal{X}$. Then for $h_{{\mathcal{P}_i}}^{\star} = \arg {\min _{h \in \mathcal{H}}}{{\mathcal{R}}_{{\mathcal{P}_i}}}(h,f_{{\mathcal{P}_i}})$ and $h_{\tilde{\mathcal{P}}_i}^{\star} = \arg {\min _{h \in \mathcal{H}}}{{\mathcal{R}}_{\tilde{\mathcal{P}}_i}}(h,f_{\tilde{\mathcal{P}}_i})$ where $f_{\tilde{\mathcal{P}}_i} \in \mathcal{H}$ \vspace{-2pt}is the ground truth function (identity function under the encoder-decoding process) for 
	$\tilde{\mathcal{P}}_i$.
	
	Then according to the triangle inequality property of $\mathcal{L}$, applied twice, we have~: 
	\begin{equation}
	\begin{aligned}
	{{\mathcal{R}}_{{{\mathcal P}_i}}}\big( h,{f_{\mathcal{P}_i}} \big) &\le {{\mathcal{R}}_{{{\mathcal P}_i}}}\big( {h,h^*_{{\tilde{\mathcal{P}}}_i}} \big) + {{\mathcal{R}}_{{{\mathcal P}_i}}}\big( {h^*_{{\tilde{\mathcal{P}}}_i},h_{{\mathcal{P}_i}}^*} \big) + {{\mathcal{R}}_{{{\mathcal P}_i}}}\big( {h_{{\mathcal{P}_i}}^*},{f_{\mathcal{P}_i}} \big)
	\label{theorem1_equ1}
	\end{aligned}
	\end{equation}
	
	Eq.~\eqref{theorem1_equ1} holds because, after applying twice the triangle inequality, $\mathcal{L}(a,b) \le \mathcal{L}(a,c) + \mathcal{L}(c,d) + L(d,b)$ where $a,b,c,d$ are $h({\bf x}), {f_{\mathcal{P}_i}}({\bf x}), {h^*_{{\tilde{\mathcal{P}}}_i}}({\bf x}),h_{{\mathcal{P}_i}}^*({\bf x})$ and ${\bf x}$ is sampled from the same domain ${\mathcal{P}}_i$. According to the definition of discrepancy distance (See Definition 2 of the paper), we define the discrepancy distance between ${\mathcal{P}}_i$ and ${\tilde {\mathcal{P}}}_i$ as~:
	
	\begin{equation}
	\begin{aligned}
	disc_{\mathcal{L}}\big({\mathcal{P}_i},{\tilde{\mathcal{P}}}_i \big) =  {\sup _{\left( {h,h'} \right) \in {\cal H}}}\Big| {{{\mathbb E}_{{\bf{x}} \sim {\mathcal{P}_i} }}\left[ {{\cal L}\left( {h'\left( {\bf{x}} \right),h\left( {\bf{x}} \right)} \right)} \right] - {{\mathbb E}_{{\bf{x}} \sim {\tilde {\mathcal{P}}_i} }}\left[ {{\cal L}\left( {h'\left( {\bf{x}} \right),h\left( {\bf{x}} \right)} \right)} \right]} \Big|.
	\end{aligned}
	\end{equation}
	
	We rewrite the above equation as~:
	\begin{equation}
	\begin{aligned}
	disc_{\mathcal{L}}\big({\mathcal{P}_i},{\tilde{\mathcal{P}}}_i \big) =  {\sup _{\left( {h,h'} \right) \in {\cal H}}}\big|
	{{\mathcal{R}}_{{{\mathcal P}_i}}}\left( h,h' \right) - {{\mathcal{R}}_{{\tilde{ {\mathcal P}}_i}}}\left( h,h' \right)
	\big|.
	\label{theorem1_disc1}
	\end{aligned}
	\end{equation}
	
	We consider $h'$ to be $h^*_{{\tilde{\mathcal{P}}}_i}$ in Eq.~\eqref{theorem1_disc1} and we have~:
	\begin{equation}
	\begin{aligned}
	{\sup _{\left( {h,h'} \right) \in H}}\Big|
	{{\mathcal{R}}_{{{\mathcal P}_i}}}\left( h,h' \right) - {{\mathcal{R}}_{{\tilde{ {\mathcal P}}_i}}}\left( h,h' \right)
	\Big| \ge \Big|
	{{\mathcal{R}}_{{{\mathcal P}_i}}}\big( h,h^*_{{\tilde{\mathcal{P}}}_i} \big) - {{\mathcal{R}}_{{\tilde{ {\mathcal P}}_i}}}\big( h,h^*_{{\tilde{\mathcal{P}}}_i} \big)
	\Big|
	\label{theorem1_disc2}
	\end{aligned}
	\end{equation}
	We also know that~:
	\begin{equation}
	\begin{aligned}
	{{\mathcal{R}}_{{{\mathcal P}_i}}}\big( h,h^*_{{\tilde{\mathcal{P}}}_i} \big) \le  \Big|
	{{\mathcal{R}}_{{{\mathcal P}_i}}}\big( h,h^*_{{\tilde{\mathcal{P}}}_i} \big) - {{\mathcal{R}}_{{\tilde{ {\mathcal P}}_i}}}\big( h,h^*_{{\tilde{\mathcal{P}}}_i} \big)
	\Big| + {{\mathcal{R}}_{{\tilde{ {\mathcal P}}_i}}}\big( h,h^*_{{\tilde{\mathcal{P}}}_i} \big)
	\label{theorem1_disc3}
	\end{aligned}
	\end{equation}
	
	Therefore, we can replace the first term of the right hand side of Eq.~\eqref{theorem1_equ1} by the right hand side of Eq.~\eqref{theorem1_disc3}, resulting in~:
	\begin{equation}
	\begin{aligned}
	\vspace{-30pt}
	{{\mathcal{R}}_{{{\mathcal P}_i}}}\left( h,{f_{\mathcal{P}_i}} \right) &\le
	{{\mathcal{R}}_{{\tilde{\mathcal{P}}}_i}}\big( {h,h^*_{{\tilde{\mathcal{P}}}_i}} \big) + \Big|
	{{\mathcal{R}}_{{{\mathcal P}_i}}}\big( h,h^*_{{\tilde{\mathcal{P}}}_i} \big) - {{\mathcal{R}}_{{\tilde{ {\mathcal P}}_i}}}\big( h,h^*_{{\tilde{\mathcal{P}}}_i} \big)
	\Big| + 
	{{\mathcal{R}}_{{{\mathcal P}_i}}}\big( h^*_{{\mathcal{P} }_i},h^*_{{\tilde{\mathcal{P}}}_i} \big)+
	{{\mathcal{R}}_{{{\mathcal P}_i}}}\big( h^*_{{\mathcal P}_i},{f_{\mathcal{P}_i}} \big)
	\label{theorem1_eq4}
	\end{aligned}
	\end{equation}
	
	Then the second term, representing the absolute value of the difference, in RHS of Eq.~\eqref{theorem1_eq4} can be replaced by $disc_{\mathcal{L}}({\mathcal{P}_i},{\tilde{\mathcal{P}}}_i )$  from  Eq.~\eqref{theorem1_disc2}), since the discrepancy distance between two distributions is an upper bound to this absolute value, resulting in~:
	\begin{equation}
	\begin{aligned}
	\vspace{-30pt}
	{{\mathcal{R}}_{{{\mathcal P}_i}}}\big( h,{f_{\mathcal{P}_i}} \big) &\le
	{{\mathcal{R}}_{{\tilde{\mathcal{P}}}_i}}\big( {h,h^*_{{\tilde{\mathcal{P}}}_i}} \big) + disc_{\mathcal{L}}({\mathcal{P}_i},{\tilde{\mathcal{P}}}_i ) + 
	{{\mathcal{R}}_{{{\mathcal P}_i}}}\big( h^*_{{\mathcal{P} }_i},h^*_{{\tilde{\mathcal{P}}}_i} \big)+
	{{\mathcal{R}}_{{{\mathcal P}_i}}}\big( h^*_{{\mathcal P}_i},{f_{\mathcal{P}_i}} \big)
	\label{theorem1_eq5}
	\end{aligned}
	\end{equation}
	
	From {\bf Definition 3} from the paper, we know that $disc_{\mathcal{L}}({\mathcal{P}}_i,{\tilde{\mathcal{P}}}_i) \le disc^\star_{\mathcal{L}}({\mathcal{P}}_i,{\tilde{\mathcal{P}}}_i)$. This allows use to replace $disc_{\mathcal{L}}({\mathcal{P}_i},{\tilde{\mathcal{P}}}_i )$ by using $disc^\star_{\mathcal{L}}({\mathcal{P}}_i,{\tilde{\mathcal{P}}}_i)$ in Eq.~\eqref{theorem1_eq5}, resulting in~:
	\begin{equation}
	\begin{aligned}
	\vspace{-30pt}
	{{\mathcal{R}}_{{{\mathcal P}_i}}}\big( h,{f_{\mathcal{P}_i}} \big)& \le
	{{\mathcal{R}}_{{\tilde{\mathcal{P}}}_i}}\big( {h,h^*_{{\tilde{\mathcal{P}}}_i}} \big) + disc^\star_{\mathcal{L}}\big({\mathcal{P}_i},{\tilde{\mathcal{P}}}_i \big) + 
	{{\mathcal{R}}_{{{\mathcal P}_i}}}\big( h^*_{{\mathcal{P} }_i},h^*_{{\tilde{\mathcal{P}}}_i} \big)+
	{{\mathcal{R}}_{{{\mathcal P}_i}}}\big( h^*_{{\mathcal P}_i},{f_{\mathcal{P}_i}} \big)
	\label{theorem1_eq6}
	\end{aligned}
	\end{equation}
	
	Eq.~\eqref{theorem1_eq6} proves {\bf Theorem 1} and a similar proof can be found in Theorem 8 from \cite{domainTheory}.
	
	\section{The proof of Theorem 2}
	
	\noindent \textBF{Theorem 2.}
	For a given sequence of tasks $\{\mathcal{T}_1,\dots,\mathcal{T}_t \}$, we derive a GB between the target distribution and the evolved source distribution during the $t$-th task learning~:
	\begin{equation}
	\begin{aligned}
	\frac{1}{t} \sum\nolimits_{i = 1}^t {{{\mathcal{R}}_{{\mathcal{P}_i}}}} \left( {h,{f_{{\mathcal{P}_i}}}} \right)&\le {{\mathcal{R}}_{{{\mathbb{P}}^{t - 1}} \otimes {{\tilde {\mathcal{P}}}_t}}}\left( {h,h_{{{\mathbb{P}}^{t - 1}} \otimes {{\tilde{\mathcal{P}}}_t}}^*} \right) + {\mathcal R}_A \left({{\mathcal{P}_{(1:t)}},{{{\mathbb{P}}^{t - 1}} \otimes {{\tilde {\mathcal{P}}}_t}}} \right),
	\label{theorem2_equ1}
	\end{aligned}
	\end{equation}
	\noindent where ${\mathcal{P}_{(1:t)}}$ is the mixture distribution $\{{\mathcal{P}}_1 \otimes {\mathcal{P}}_2 ,\dots,\otimes {\mathcal{P}}_t\}$. As it can be seen in Eq.~\eqref{theorem2_equ1}, the performance on the target domain is largely depending on the discrepancy term even if ${\cal M}$ minimizes the source risk well. In the following, we provide an analytical bound that considers all previously learnt distributions. 
	\begin{equation}
	\begin{aligned}
	\frac{1}{t} {\sum\nolimits_{i = 1}^t {{{\mathcal{R}}_{{\mathcal{P}_i}}}} \left( {h,{f_{{\mathcal{P}_i}}}} \right)} &\le
	{{\mathcal{R}}_{{{\mathbb{P}}^{t - 1}} \otimes {{\tilde {\mathcal{P}}}_t}}}\left( {h,h_{{{\mathbb{P}}^{t - 1}} \otimes {{\tilde{\mathcal{P}}}_t}}^*} \right) + {\rm Err}^a + {\rm Err}^d,
	\label{theorem2_equ3}
	\end{aligned}
	\end{equation}
	where ${\rm Err}^d \ge 0$ evaluates the difference on the two risk terms, expressed by:
	\begin{equation}
	\begin{aligned}
	& \sum\nolimits_{k = 1}^{t - 1} \Big\{ {{{\mathcal{R}}_{{{\mathbb{P}}^{(t-k)}}}}\left( {h,h_{{{\mathbb{P}}^{(t-l)}}}^*} \right)} 
	- 
	{{\mathcal{R}}_{{{\mathbb{P}}^{(t-1-k)}} \otimes {{\tilde{\mathcal{P}}}_{(t-k)}}}}\left( {h,h_{{{\mathbb{P}}^{(t-1-k)}} \otimes {{\tilde{\mathcal{P}}}_{(t-k)}}}^*} \right) \Big\}  \,,
	\label{theorem1_equ5}
	\end{aligned}
	\end{equation}
	where  ${\mathcal R}_{{\mathbb P}^0 \otimes {\tilde{\mathcal{P}}}_1 } (h,h^*_{{\mathbb P}^0 \otimes {\tilde{\mathcal{P}}}_1}) =
	{{\mathcal{R}}_{{{\tilde{\mathcal{P}}}_1}}}( {h,h_{{{\tilde{\mathcal{P}}}_1}}^*})$. ${\rm Err}^a$ is the accumulated error term expressed by~:
	\begin{equation}
	\begin{aligned}
	& \sum\limits_{k = 1}^{t-2} \Big\{
	{\mathcal R}_A \left({{{\mathbb{P}}^{(t-1-k)}} \otimes {{\tilde{\mathcal{P}}}_{(t-k) }},{{\mathbb{P}}^{(t-k)}}} \right) \Big\} +  {\mathcal R}_A \left( {{{\cal P}_{(1:t)}},{{\mathbb{P}}^{(t - 1)}} \otimes {{\tilde {\cal P}}_t}}\right)+{\mathcal R}_A \left( 
	{{{\tilde{\mathcal{P}}}_1},{{\mathbb{P}}^1}}
	\right).
	\label{theorem1_equ6}
	\end{aligned}
	\end{equation}
	
	From Eq.~\eqref{theorem1_equ6}, we can observe that while learning more tasks ($t$ is increased) during the lifelong learning, the gap on the GB for ${\mathcal M}$ tends to be large since ${\rm Err}^a$ increases. This can also explain why GR fails when learning a long sequence of tasks \cite{GenerativeLifelong,LifelongVAEGAN}. Additionally, the term ${\mathcal R}_A ({{{\mathbb{P}}^{(t-1-k)}} \otimes {{\tilde{\mathcal{P}}}_{(t-k) }},{{\mathbb{P}}^{(t-k)}}} )$ and ${\rm Err}^d$ tend to be small when the discrepancy $disc^*_{\mathcal{L}}({\mathbb{P}^{i}},{\mathbb{P}^{(i-1)}} \otimes {\tilde{\mathcal{P}}_i})$ is equal to 0 in each task learning ($i=1,\dots,t$). This is achieved by the optimal generator distribution that approximates ${\mathbb{P}^{(i-1)}} \otimes {\tilde{\mathcal{P}}_i}$ exactly in each task learning (See also other explanation in Proposition 3 in Appendix~\ref{section_selfVAE}). 
	
	\noindent \textBF{Proof.}
	Firstly, we can derive the bound according to Theorem 1~:
	\begin{equation}
	\begin{aligned}
	{{{\mathcal{R}}_{{{\cal P}_{(i:t)}}}}} \left( {h,{f_{{{\cal P}_{(i:t)}}}}} \right) &\le {{\mathcal{R}}_{{{\mathbb{P}}^{(t - 1)}} \otimes {{\tilde {\cal P}}_t}}}\left( {h,h_{{{\mathbb{P}}^{(t - 1)}} \otimes {{\tilde {\cal P}}_t}}^*} \right) + disc_{\mathcal{L}}^{\star}\left( {{{\cal P}_{(1:t)}},{{\mathbb{P}}^{(t - 1)}} \otimes {{\tilde {\cal P}}_t}} \right) + \varepsilon \left( {{{\cal P}_{(1:t)}},{{\mathbb{P}}^{(t - 1)}} \otimes {{\tilde {\cal P}}_t}} \right)
	\label{Theorem2_obj_obj1}
	\end{aligned}
	\end{equation}
	where ${\mathcal{P}_{(1:t)}}$ represents the mixture distribution $\{{\mathcal{P}}_1 \otimes {\mathcal{P}}_2 ,\dots,\otimes {\mathcal{P}}_t\}$. Let $\rho_{(1:t)}({\bf x})$ represent the density function for ${\mathcal{P}_{(1:t)}}$ and $\rho_{(i)}({\bf x})$ the density function for $\mathcal{P}_i$. Since $\mathcal{P}_{(1:t)}$ is the mixture distribution and its density is expressed by $\rho_{(1:t)}({\bf x}) = \frac{1}{t} \sum\nolimits_{i = 1}^t \rho_{(i)}({\bf x})$. We know that ${{{\mathcal{R}}_{{{\cal P}_{(i:t)}}}}} ( {h,{f_{{{\cal P}_{(i:t)}}}}} )$ can be rewritten as the integral form $\int {{\rho_{(i:t)}}({\bf{x}}){\mathcal{L}}(h,{f_{{P_{(1:t)}}}})} \, \mathrm{d}{\bf{x}}$. We then take $\rho_{(1:t)}({\bf x}) = \frac{1}{t} \sum\nolimits_{i = 1}^t \rho_{(i)}({\bf x})$ in this integral form, resulting in~:
	\begin{equation}
	\begin{aligned}
	\frac{1}{t}\sum\nolimits_{i = 1}^t {\int {{\rho _{(i)}}({\bf{x}}){\mathcal{L}}(h,{f_{{P_{(1:t)}}}})} {\mkern 1mu} {\rm{d}}{\bf{x}}}
	\label{Theorem2_obj_integral}
	\end{aligned}
	\end{equation}
	
	We assume that ${\mathcal{P}}_i$ is independent from ${\mathcal{P}_j}$, where $i \ne j$, which is a reasonable assumption, since each task is associated with a different dataset. Therefore, the true labeling function $f_{\mathcal{P}_{(1:t)}}$ can be represented by $f_{{{\mathcal{P}}_{i}}}$ under the target distribution ${\mathcal{P}}_i$ of the $i$-th task. Then we rewrite Eq.~\eqref{Theorem2_obj_integral} as the expectation form $\frac{1}{t}\sum\nolimits_{i = 1}^t {{{\mathcal{R}}_{{{\mathcal{P}}_i}}}(h,{f_{{{\mathcal{P}}_i}}})}$.
	
	Based on the above results, Eq.~\eqref{Theorem2_obj_obj1} is rewritten as~:
	\begin{equation}
	\begin{aligned}
	\frac{1}{t}\sum\limits_{i = 1}^t {{{\mathcal{R}}_{{{\mathcal{P}}_i}}}(h,{f_{{{\mathcal{P}}_i}}})}  &\le {{\mathcal{R}}_{{{\mathbb{P}}^{(t - 1)}} \otimes {{\tilde {\cal P}}_t}}}\left( {h,h_{{{\mathbb{P}}^{(t - 1)}} \otimes {{\tilde {\cal P}}_t}}^*} \right) \\&+ disc_{\mathcal{L}}^{\star}\left( {{{\cal P}_{(1:t)}},{{\mathbb{P}}^{(t - 1)}} \otimes {{\tilde {\cal P}}_t}} \right) + \varepsilon \left( {{{\cal P}_{(1:t)}},{{\mathbb{P}}^{(t - 1)}} \otimes {{\tilde {\cal P}}_t}} \right)
	\label{Theorem2_obj_obj2}
	\end{aligned}
	\end{equation}
	
	In the following, we provide the derivations for ${\rm Err}^a$ and ${\rm Err}^d$ from Eq.~(\ref{theorem2_equ3}). We consider to take ${{{\mathbb{P}}^{t - 2}} \otimes {{\tilde{\mathcal{P}}}_{t - 1}}}$ and ${{{\mathbb{P}}^{t - 1}}}$ as the target and source domains, respectively. This is reasonable choice since we allow the ${{{\mathbb{P}}^{t - 1}}}$ (the generator distribution of the model) to approximate  ${{{\mathbb{P}}^{t - 2}} \otimes {{\tilde{\mathcal{P}}}_{t - 1}}}$. We derive the bound as~:
	\begin{equation}
	\begin{aligned}
	&{{\mathcal{R}}_{{{\mathbb{P}}^{t - 2}} \otimes {{\tilde{\mathcal{P}}}_{t - 1}}}}\left( {h,h_{{{\mathbb{P}}^{t - 2}} \otimes {{\tilde{\mathcal{P}}}_{t - 1}}}^*} \right) \le {{\mathcal{R}}_{{{\mathbb{P}}^{t - 1}}}}\left( {h,h_{{{\mathbb{P}}^{t - 1}}}^*} \right) +  disc_{\cal L}^\star \left( {{{\mathbb{P}}^{t - 2}} \otimes {{\tilde{\mathcal{P}}}_{t - 1}},{{\mathbb{P}}^{t - 1}}} \right) + \varepsilon \left( {{{\mathbb{P}}^{t - 2}} \otimes {{\tilde{\mathcal{P}}}_{t - 1}},{{\mathbb{P}}^{t - 1}}} \right)
	\end{aligned}
	\end{equation}
	
	We then consider to take ${{{\mathbb{P}}^{t - 3}} \otimes {{\tilde{\mathcal{P}}}_{t - 2}}}$ and ${{{\mathbb{P}}^{t - 2}}}$ as the target and source domains, respectively, and we derive the bound as~:
	\begin{equation}
	\begin{aligned}
	&{{\mathcal{R}}_{{{\mathbb{P}}^{t - 3}} \otimes {{\tilde{\mathcal{P}}}_{t - 2}}}}\left( {h,h_{{{\mathbb{P}}^{t - 3}} \otimes {{\tilde{\mathcal{P}}}_{t - 2}}}^*} \right) \le {{\mathcal{R}}_{{{\mathbb{P}}^{t - 2}}}}\left( {h,h_{{{\mathbb{P}}^{t - 2}}}^*} \right) +  disc_{\cal L}^\star \left( {{{\mathbb{P}}^{t - 3}} \otimes {{\tilde{\mathcal{P}}}_{t - 2}},{{\mathbb{P}}^{t - 2}}} \right) + \varepsilon \left( {{{\mathbb{P}}^{t - 3}} \otimes {{\tilde{\mathcal{P}}}_{t - 2}},{{\mathbb{P}}^{t - 2}}} \right)
	\end{aligned}
	\end{equation}
	
	According to the Inductive Inference, we have~:
	\begin{equation}
	\begin{aligned}
	\dots \\
	\dots \\
	&{{\mathcal{R}}_{{{\mathbb{P}}^1} \otimes {{\tilde{\mathcal{P}}}_2}}}\left( {h,h_{{{\mathbb{P}}^1}}^*} \right) \le {{\mathcal{R}}_{{{\mathbb{P}}^2}}}\left( {h,h_{{{\mathbb{P}}^2}}^*} \right) + disc_{\cal L}^\star \left( {{{\mathbb{P}}^1} \otimes {{\tilde {\mathcal{P}}}_2},{{\mathbb{P}}^2}} \right) + \varepsilon \left( {{{\mathbb{P}}^1} \otimes {{\tilde {\mathcal{P}}}_2},{{\mathbb{P}}^2}} \right)
	\end{aligned}
	\end{equation}
	
	\begin{equation}
	\begin{aligned}
	&{{\mathcal{R}}_{{{\tilde{\mathcal{P}}}_1}}}\left( {h,h_{{{\tilde {\mathcal{P}}}_1}}^*} \right) \le {{\mathcal{R}}_{{{\mathbb{P}}^1}}}\left( {h,h_{{{\mathbb{P}}^1}}^*} \right) + disc_{\cal L}^\star \left( {{{\tilde {\mathcal{P}}}_1},{{\mathbb{P}}^1}} \right) + \varepsilon \left( {{{\tilde{\mathcal{P}}}_1},{{\mathbb{P}}^1}} \right)
	\label{theorem2_proof_eq10}
	\end{aligned}
	\end{equation}
	
	We then sum up all the above derivations in the inequality from Eq.~\eqref{Theorem2_obj_obj2} to Eq.~\eqref{theorem2_proof_eq10}, resulting in~:
	
	\begin{equation}
	\begin{aligned}
	&\frac{1}{t} {\sum\limits_{i = 1}^t
		\Big\{
		{{{\mathcal{R}}_{{\mathcal{P}_i}}}} \left( {h,{f_{{\mathcal{P}_i}}}} \right)} \Big\} +{{\mathcal{R}}_{{{\tilde{\mathcal{P}} }_1}}}\left( {h,h_{{{\tilde {\mathcal{P}}}_1}}^*} \right) +   {\sum\limits_{k = 1}^{t - 2} \Big\{ {{{\mathcal{R}}_{{{\mathbb{P}}^{t - 1 - k}} \otimes {{\tilde {\mathcal{P}}}_{t - k}}}}\left( {h,h_{{{\mathbb{P}}^{t - 1 - k}} \otimes {{\tilde {\mathcal{P}}}_{t - k}}}^*} \right)} } \Big\}  \le   {\sum\limits_{k = 1}^{t - 2} \Big\{ {{{\mathcal{R}}_{{{\mathbb{P}}^{t - k}}}}\left( {h,h_{{{\mathbb{P}}^{t - k}}}^*} \right)} } \Big\} \\&+
	{\mathcal R}_{{\mathbb P}^1} \left(h,h^*_{{\mathbb P}^1}\right) + 
	{{\mathcal{R}}_{{{\mathbb{P}}^{t - 1}} \otimes {{\tilde {\mathcal{P}}}_t}}}\left( {h,h_{{{\mathbb{P}}^{t - 1}} \otimes {{\tilde{\mathcal{P}}}_t}}^*} \right) 
	+  \sum\limits_{k = 1}^{t - 2} \Big\{ disc_{\cal L}^\star \left( {{{\mathbb{P}}^{t - 1 - k}} \otimes {{\tilde{\mathcal{P}}}_{t - k}},{{\mathbb{P}}^{t - k}}} \right) 
	+
	\varepsilon \left( {{{\mathbb{P}}^{t - 1 - k}} \otimes {{\tilde{\mathcal{P}}}_{t - k}},{{\mathbb{P}}^{t - k}}} \right) \Big\}  \\&+ disc_{\mathcal L}^\star \left( {{{\cal P}_{(1:t)}},{{\mathbb{P}}^{(t - 1)}} \otimes {{\tilde {\cal P}}_t}} \right) + \varepsilon \left( {{{\cal P}_{(1:t)}},{{\mathbb{P}}^{(t - 1)}} \otimes {{\tilde {\cal P}}_t}} \right) + disc_{\cal L}^\star \left( {{{\tilde{\mathcal{P}}}_1},{{\mathbb{P}}^1}} \right) + \varepsilon \left( {{{\tilde{\mathcal{P}}}_1},{{\mathbb{P}}^1}} \right)
	\label{theorem2_equ2}
	\end{aligned}
	\end{equation}
	
	Then we move the second and third term in the left hand side to the right hand side in Eq.\eqref{theorem2_equ2}, resulting in~:
	
	\begin{equation}
	\begin{aligned}
	\frac{1}{t} {\sum\limits_{i = 1}^t \Big\{ {{{\mathcal{R}}_{{\mathcal{P}_i}}}} \left( {h,{f_{{\mathcal{P}_i}}}} \right)} \Big\} &\le
	{{\mathcal{R}}_{{{\mathbb{P}}^{t - 1}} \otimes {{\tilde {\mathcal{P}}}_t}}}\left( {h,h_{{{\mathbb{P}}^{t - 1}} \otimes {{\tilde{\mathcal{P}}}_t}}^*} \right) + {\mathcal R}_{{\mathbb P}^1} \left(h,h^*_{{\mathbb P}^1}\right) 
	- {{\mathcal{R}}_{{{\tilde{\mathcal{P}}}_1}}}\left( {h,h_{{{\tilde{\mathcal{P}}}_1}}^*} \right) \\&+  {\sum\limits_{k = 1}^{t - 2} \Big\{ {{{\mathcal{R}}_{{{\mathbb{P}}^{t - k}}}}\left( {h,h_{{{\mathbb{P}}^{t - k}}}^*} \right)}  - {{\mathcal{R}}_{{{\mathbb{P}}^{t - 1 - k}} \otimes {{\tilde{\mathcal{P}}}_{t - k}}}}\left( {h,h_{{{\mathbb{P}}^{t - 1 - k}} \otimes {{\tilde{\mathcal{P}}}_{t - k}}}^*} \right)} \Big\} \\&+  \sum\limits_{k = 1}^{t - 2} \left( disc_{\cal L}^\star \left( {{{\mathbb{P}}^{t - 1 - k}} \otimes {{\tilde{\mathcal{P}}}_{t - k}},{{\mathbb{P}}^{t - k}}} \right) 
	\right. \\
	& \left. 
	+ \varepsilon \left( {{{\mathbb{P}}^{t - 1 - k}} \otimes {{\tilde{\mathcal{P}}}_{t - k}},{{\mathbb{P}}^{t - k}}} \right)  \right) + disc_{\mathcal L}^\star \left( {{{\cal P}_{(1:t)}},{{\mathbb{P}}^{(t - 1)}} \otimes {{\tilde {\cal P}}_t}} \right) \\&+ \varepsilon \left( {{{\cal P}_{(1:t)}},{{\mathbb{P}}^{(t - 1)}} \otimes {{\tilde {\cal P}}_t}} \right) + disc_{\cal L}^\star \left( {{{\tilde{\mathcal{P}}}_1},{{\mathbb{P}}^1}} \right) + \varepsilon \left( {{{\tilde{\mathcal{P}}}_1},{{\mathbb{P}}^1}} \right)
	\label{theorem2_equ3_1}
	\end{aligned}
	\end{equation}
	
	Then we can rewrite Eq.~\eqref{theorem2_equ3_1} as~: 
	\begin{equation}
	\begin{aligned}
	\frac{1}{t} {\sum\limits_{i = 1}^t
		\Big\{
		{{{\mathcal{R}}_{{\mathcal{P}_i}}}} \left( {h,{f_{{\mathcal{P}_i}}}} \right)} \Big\} &\le
	{{\mathcal{R}}_{{{\mathbb{P}}^{t - 1}} \otimes {{\tilde {\mathcal{P}}}_t}}}\left( {h,h_{{{\mathbb{P}}^{t - 1}} \otimes {{\tilde{\mathcal{P}}}_t}}^*} \right) +
	{\mathcal R}_{{\mathbb P}^1} \left(h,h^*_{{\mathbb P}^1}\right) 
	- {{\mathcal{R}}_{{{\tilde{\mathcal{P}}}_1}}}\left( {h,h_{{{\tilde{\mathcal{P}}}_1}}^*} \right) \\&+ \underbrace{  {\sum\limits_{k = 1}^{t - 1} \Big\{ {{{\mathcal{R}}_{{{\mathbb{P}}^{t - k}}}}\left( {h,h_{{{\mathbb{P}}^{t - k}}}^*} \right)}  - {{\mathcal{R}}_{{{\mathbb{P}}^{t - 1 - k}} \otimes {{\tilde{\mathcal{P}}}_{t - k}}}}\left( {h,h_{{{\mathbb{P}}^{t - 1 - k}} \otimes {{\tilde{\mathcal{P}}}_{t - k}}}^*} \right)} \Big\} }_{\rm{Err}^a} \\&+ \underbrace{  \sum\limits_{k = 1}^{t - 2} \Big\{ {\mathcal R}_A \left({{{\mathbb{P}}^{t - 1 - k}} \otimes {{\tilde{\mathcal{P}}}_{t - k}},{{\mathbb{P}}^{t - k}}} \right) \Big\}   +
		{\mathcal R}_A \left( {{{\cal P}_{(1:t)}},{{\mathbb{P}}^{(t - 1)}} \otimes {{\tilde {\cal P}}_t}}\right) + {\mathcal R}_A \left( 
		{{{\tilde{\mathcal{P}}}_1},{{\mathbb{P}}^1}}
		\right)}_{\rm{Err}^d} 
	\end{aligned}
	\end{equation}
	where ${\mathcal R}_{{\mathbb P}^0 \otimes {\tilde{\mathcal{P}}}_1 } (h,h^*_{{\mathbb P}^0 \otimes {\tilde{\mathcal{P}}}_1}) =
	{{\mathcal{R}}_{{{\tilde{\mathcal{P}}}_1}}}( {h,h_{{{\tilde{\mathcal{P}}}_1}}^*})$ and this proves Theorem 2.
	
	\section{The proof of Lemma 1}
	\label{AppendixC}
	
	According to the bound on the KL divergence~:
	\begin{equation}
	\begin{aligned}
	\frac{1}{t}\sum\limits_{i = 1}^t {{\mathbb{E}_{{{\mathcal{P}}_i}}}KL(q_{\omega^t}({\bf{z}} \mid {\bf{x}}_i^T) \mid\mid p({\bf{z}}))}  &\le {{\mathbb E}_{{{\mathbb P}^{t-1} \otimes {\tilde{\mathcal{P}}}_t }}}KL(q_{\omega^t}({\bf{z}} \mid {\bf{\tilde x}}^t) \mid\mid p({\bf{z}})) \\&+ \Big| { {{\mathbb E}_{{{\mathbb P}^{t-1} \otimes {\tilde{\mathcal{P}}}_t }}}KL(q_{\omega^t}({\bf{z}} \mid {\bf{\tilde x}}^t) \mid\mid p({\bf{z}})) - \frac{1}{t}\sum\limits_{i = 1}^t {{\mathbb{E}_{{{\mathcal{P}}_i}}}KL(q_{\omega^t}({\bf{z}} \mid {\bf{x}}_i^T) \mid\mid p({\bf{z}}))} } \Big|
	\label{theorem2_equ4}
	\end{aligned}
	\end{equation}
	
	We also know that $\mathcal{L}_{ELBO}\left({\bf x};\{\theta,\omega \} \right)$ is expressed as~:
	
	\begin{equation}
	\begin{aligned}
	\mathcal{L}_{ELBO}\left({\bf x};\{\theta,\omega \} \right) := {\mathbb{E}_{{q_\omega }\left( {{\bf z}\mid{\bf x}} \right)}}\left[ {\log {p_\theta }\left( {{\bf x} \mid {\bf z}} \right)} \right]  - KL\left[ {{q_\omega }\left( {{\bf z}\,|\,{\bf x}} \right) \mid\mid p\left( {\bf z} \right)} \right]\,,
	\end{aligned}
	\end{equation}
	
	When the decoder models a Gaussian distribution, ${\log {p_\theta }\left( {{\bf x}\,|\,{\bf z}} \right)}$ can be represented as~:
	\begin{equation}
	\begin{aligned}
	{\log {p_\theta }\left( {{\bf x}\,|\,{\bf z}} \right)} =  - \frac{1}{{2\sigma _\theta ^2\left( {\bf{z}} \right)}}{\left\| {{\bf{x}} - {\mu _\theta }\left( {\bf{z}} \right)} \right\|^2} - \frac{1}{2}\log 2\pi \sigma _\theta ^2\left( {\bf{z}} \right)
	\label{logP_1}
	\end{aligned}
	\end{equation}
	where ${\sigma _\theta \left( {\bf{z}} \right)}$ and ${{\mu _\theta }\left( {\bf{z}} \right)}$ are the variance and mean of Gaussian distribution, obtained by the decoder. ${\left\| {\cdot } \right\|^2}$ represents the reconstruction error (square loss). We implement the decoder by a Gaussian distribution with the fixed variance ${\mathcal{N}}({\mu _\theta }\left( {\bf{z}} \right),\sigma {\bf{I}})$ where ${{\mu _\theta }\left( {\bf{z}} \right)}$ is a deep convolutional neural network and ${\bf{I}}$ is the identity matrix. Therefore, Eq.~\eqref{logP_1} is represented by the fixed variance $\sigma$~:
	\begin{equation}
	\begin{aligned}
	{\log {p_\theta }\left( {{\bf x}\,|\,{\bf z}} \right)} =  - \frac{1}{{2\sigma  ^2}}{\left\| {{\bf{x}} - {\mu _\theta }\left( {\bf{z}} \right)} \right\|^2} - \frac{1}{2}\log 2\pi \sigma  ^2
	\label{logP_2}
	\end{aligned}
	\end{equation}
	
	Since $h$ is the hypothesis of the model ($\mathcal{M}^t$), implemented as a encoding-decoding process, we have $${\mathcal{L}_{ELBO}}({\bf{x}}_i^T ;h ) =  - \frac{1}{{2\sigma  ^2}}  {\mathcal{L}}(h({\bf{x}}_i^T), {f_{{{\mathcal{P}}_i}}}({\bf{x}}_i^T)) 
	-  \frac{1}{2}\log 2\pi \sigma  ^2
	- {{KL}}(q_{\omega^t}({\bf{z}} \mid {\bf{x}}_i^T) \mid\mid p({\bf{z}}))$$. 
	
	Then we focuse on the negative ELBO 
	$$-{\mathcal{L}_{ELBO}}({\bf{x}}_i^T ;h ) =  \frac{1}{{2\sigma  ^2}}  {\mathcal{L}}(h({\bf{x}}_i^T), {f_{{{\mathcal{P}}_i}}}({\bf{x}}_i^T))
	+  \frac{1}{2}\log 2\pi \sigma  ^2
	+ {{KL}}(q_{\omega^t}({\bf{z}} \mid {\bf{x}}_i^T) \mid\mid p({\bf{z}}))$$. 
	
	And we know that ${\mathcal R}_{\mathcal{P}_i} (h,f_{\mathcal{P}_i}) = {\mathbb E}_{{\bf x} \sim \mathcal{P}_i} {\mathcal{L}} (h({\bf x}^T_i),f_{\mathcal{P}_i}({\bf x}^T_i))$ and we have~:
	\begin{equation}
	\begin{aligned}
	{\mathbb E}_{{\bf x} \sim \mathcal{P}_i} \left[ -\mathcal{L}_{ELBO}({\bf x}^T_i;h) \right] = {\mathbb E}_{{\bf x}^T_i \sim \mathcal{P}_i} \left\{ \frac{1}{{2\sigma  ^2}}  {\mathcal {L}}(h({\bf x}^T_i),f_{\mathcal{P}_i}({\bf x}^T_i))+ KL(q_{\omega^t}({\bf z} \mid {\bf x}^T_i) \mid\mid p({\bf z})) \right\} +  \frac{1}{2}\log 2\pi \sigma^2
	\label{logP_3}
	\end{aligned}
	\end{equation}
	
	We observe that $\frac{1}{2}\log 2\pi \sigma  ^2$ and $\frac{1}{{2\sigma  ^2}}$ are constants. In order to simplify the notations, we set $\sigma =  \frac{1}{{\sqrt 2 }}$. Therefore, Eq.~\eqref{logP_3} is rewritten as~:
	\begin{equation}
	\begin{aligned}
	{\mathbb E}_{{\bf x}^T_i \sim \mathcal{P}_i} \left[ -\mathcal{L}_{ELBO}({\bf x}^T_i;h) \right] = {\mathbb E}_{{\bf x}^T_i \sim \mathcal{P}_i}\{ {\mathcal{L}}(h({\bf x}^T_i),f_{\mathcal{P}_i}({\bf x}^T_i))+ KL(q_{\omega^t}({\bf z} \mid {\bf x}^T_i) \mid\mid p({\bf z})) \}+\frac{1}{2}\log \pi 
	\label{lemma1_proof_eu1}
	\end{aligned}
	\end{equation}
	
	Eq.~\eqref{lemma1_proof_eu1} can be seen as the average ELBO for all samples. We then take Eq.~\eqref{theorem2_equ4} in Eq.(7) in the paper, we have~:
	\begin{equation}
	\begin{aligned}
	\frac{1}{t}\sum\limits_{i = 1}^t \Big\{ {\mathbb E}_{{\bf x}^T_i \sim \mathcal{P}_i} \{ {\mathcal{L}}(h({\bf x}^T_i),f_{\mathcal{P}_i}({\bf x}^T_i)) + {{KL}}(q_{\omega^t}({\bf{z}}|{\bf{x}}_i^T) \mid\mid p({\bf{z}}))\} \Big\} &\le 
	{{\mathbb{E}}_{{\bf x}^t \sim {{\mathbb{P}}^{t - 1}} \otimes {{\tilde {\mathcal{P}}}_t}}} \Big\{{\mathcal{L}}\left( {h({\tilde{\bf x}}^t),h_{{{\mathbb{P}}^{t - 1}} \otimes {{\tilde{\mathcal{P}}}_t}}^*({\tilde{\bf x}}^t)} \right) \\&+
	{{{KL}}(q_{\omega^t}({\bf{z}} \mid {\bf{\tilde x}}^t) \mid\mid p({\bf{z}}))} \Big\} \\&+ \left| KL_1 - KL_2 \right|\  + {\mathcal R}_A \left({{\mathcal{P}_{(1:t)}},{{{\mathbb{P}}^{t - 1}} \otimes {{\tilde {\mathcal{P}}}_t}}} \right)
	\label{theorem2_equ3_2}
	\end{aligned}
	\end{equation}
	
	It notes that we can add the constant $\frac{1}{2}\log \pi$ in both sides of Eq.~\eqref{theorem2_equ3_2}. According to Eq.~\eqref{lemma1_proof_eu1}, we can rewrite Eq.~\eqref{theorem2_equ3_2} as~:
	\begin{equation}
	\begin{aligned}
	\frac{1}{t}{\sum\limits_{i = 1}^t {\mathbb{E}}_{{\bf x}^T_i \sim {\mathcal{P}_i}}  [ -{\mathcal{L}_{ELBO}\left({\bf x}^T_i ; h \right) }} ] &\le  
	{\mathbb{E}}_{{\bf x}^t \sim {\mathbb P}^{t-1}\otimes{\tilde {\mathcal{P}}}_t} \left[-{\mathcal{L}_{ELBO}\left({\bf \tilde x}^t ;h \right) }\right] + 
	\left| KL_1 - KL_2 \right|
	+ {\mathcal R}_A \left({{\mathcal{P}_{(1:t)}},{{{\mathbb{P}}^{t - 1}} \otimes {{\tilde {\mathcal{P}}}_t}}} \right)\,.
	\label{lemma1_proof_eq1}
	\end{aligned}
	\end{equation}
	This proves Lemma 1. We can further replace the latest term in the right hand side of Eq.~\eqref{lemma1_proof_eq1} by ${\rm Err}^a + {\rm Err}^d $ (See details in the proof of Theorem 2), resulting in~:
	\begin{equation}
	\begin{aligned}
	\frac{1}{t}{\sum\limits_{i = 1}^t {\mathbb{E}}_{{\bf x}^T_i \sim {\mathcal{P}_i}}  [ -{\mathcal{L}_{ELBO}\left({\bf x}^T_i ; h \right) }} ] &\le  
	{\mathbb{E}}_{{\bf x}^t \sim {\mathbb P}^{t-1}\otimes{\tilde {\mathcal{P}}}_t} \left[-{\mathcal{L}_{ELBO}\left({\bf \tilde x}^t;h \right) }\right] + 
	\left| KL_1 - KL_2 \right|
	+{\rm Err}^a + {\rm Err}^d.
	\end{aligned}
	\end{equation}
	
	\section{The proof of Theorem 3}
	
	Firstly, ${\mathcal R}_C$ can be easily proved since the task was trained only once and we simply derive the bound between the training sets and testing sets. 
	
	For the components that are trained more than once, we firstly consider the $c'_i$-th component and $a(i,j)$-th task. This can be easily generalized to other components and other tasks. We firstly consider to take ${\mathcal{P}}_{a(i,j)}$ and ${\mathbb P}^0_{a(i,j)}$ as the target and source distribution and we have a bound~:
	\begin{equation}
	\begin{aligned}
	&{{\mathcal{R}}_{{\mathcal{P}}_{a(i,j)}}}\left( {h_{c'_i},{f_{{\mathcal{P}}_{a(i,j)}}}} \right) \le {{\mathcal{R}}_{{\mathbb{P}}_{a(i,j)}^0}}\left( {h_{c'_i},{h^*_{{\mathbb{P}}_{a(i,j)}^0}}} \right) +  disc_{\mathcal L}^\star \left( {{\mathcal{P}}_{a(i,j)},{\mathbb{P}}_{a(i,j)}^0} \right) + \varepsilon \left( {{\mathcal{P}}_{a(i,j)},{\mathbb{P}}_{a(i,j)}^0} \right)
	\end{aligned}
	\end{equation}
	
	We can observe that ${\mathbb P}^0_{a(i,1)}$ represent the training set ${\tilde{\mathcal{P}}}_{a(i,j)}$. Then we consider to take ${{\mathbb{P}}_{a(i,j)}^0}$ and ${{\mathbb{P}}_{a(i,j)}^1}$ as the target and source domains, respectively. Then we have the bound as~:
	\begin{equation}
	\begin{aligned}
	&{{\mathcal{R}}_{{\mathbb{P}}_{a(i,j)}^0}}\left( {h_{c'_i},{f_{{\mathbb{P}}_{a(i,j)}^0}}} \right) \le {{\mathcal{R}}_{{\mathbb{P}}_{a(i,j)}^1}}\left( {h_{c'_i},{h^*_{{\mathbb{P}}_{a(i,j)}^1}}} \right) +  disc_{\mathcal L}^\star\left( {{\mathbb{P}}_{a(i,j)}^0,{\mathbb{P}}_{a(i,j)}^1} \right) + \varepsilon \left( {{\mathbb{P}}_{a(i,j)}^0,{\mathbb{P}}_{a(i,j)}^1} \right)
	\label{theorem3_equ5}
	\end{aligned}
	\end{equation}
	
	Similarly, we have the following bounds~:
	\begin{equation}
	\begin{aligned}
	&{{\mathcal{R}}_{{\mathbb{P}}_{a(i,j)}^1}}\left( {h_{c'_i},{h^*_{{\mathbb{P}}_{a(i,j)}^1}}} \right) \le {{\mathcal{R}}_{{\mathbb{P}}_{a(i,j)}^2}}\left( {h_{c'_i},{h^*_{{\mathbb{P}}_{a(i,j)}^2}}} \right) + disc_{\mathcal L}^\star\left( {{\mathbb{P}}_{a(i,j)}^1,{\mathbb{P}}_{a(i,j)}^2} \right) + \varepsilon \left( {{\mathbb{P}}_{a(i,j)}^1,{\mathbb{P}}_{a(i,j)}^2} \right)
	\end{aligned}
	\end{equation}
	
	\begin{equation}
	\begin{aligned}
	\dots \\
	&{{\mathcal{R}}_{{\mathbb{P}}_{a(i,j)}^{c(i,j) - 2}}}\left( {h_{c'_i},{f_{{\mathbb{P}}_{a(i,j)}^{c(i,j) - 2}}}} \right) \le  {{\mathcal{R}}_{{\mathbb{P}}_{a(i,j)}^{c(i,j) - 1}}}\left( {h_{c'_i},h_{{\mathbb{P}}_{a(i,j)}^{c(i,j) - 1}}^*} \right) + disc_{\mathcal L}^\star\left( {{\mathbb{P}}_{a(i,j)}^{c(i,j) - 2},{\mathbb{P}}_{a(i,j)}^{c(i,j) - 1}} \right) + \varepsilon \left( {{\mathbb{P}}_{a(i,j)}^{c(i,j) - 2},{\mathbb{P}}_{a(i,j)}^{c(i,j) - 1}} \right)
	\end{aligned}
	\end{equation}
	
	\begin{equation}
	\begin{aligned}
	&{{\mathcal{R}}_{{\mathbb{P}}_{a(i,j)}^{c(i,j) - 1}}}\left( {h_{c'_i},{f_{{\mathbb{P}}_{a(i,j)}^{c(i,j) - 1}}}} \right) \le {{\mathcal{R}}_{{\mathbb{P}}_{a(i,j)}^{c(i,j)}}}\left( {h_{c'_i},h_{{\mathbb{P}}_{a(i,j)}^{c(i,j)}}^*} \right) + disc_{\mathcal L}^\star\left( {{\mathbb{P}}_{a(i,j)}^{c(i,j) - 1},{\mathbb{P}}_{a(i,j)}^{c(i,j)}} \right) + \varepsilon \left( {{\mathbb{P}}_{a(i,j)}^{c(i,j) - 1},{\mathbb{P}}_{a(i,j)}^{c(i,j)}} \right)
	\end{aligned}
	\end{equation}
	
	Then we sum up all the above relationships, resulting in~:
	\begin{equation}
	\begin{aligned}
	{{\mathcal{R}}_{{\mathcal{P}_{a(i,j)}}}} \left( {h_{c'_i},{f_{{\mathcal{P}_{a(i,j)}}}}} \right) &\le {{\mathcal{R}}_{{\mathbb{P}}_{a(i,j)}^{c(i,j)}}}\left( {h_{c'_i},h_{{\mathbb{P}}_{a(i,j)}^{c(i,j)}}^*} \right)  \\&  + \sum\limits_{k =  -1}^{c(i,j) - 1} { \Big\{ {disc_{\mathcal{L}}^\star \left( {{\mathbb{P}}_{a(i,j)}^k,{\mathbb{P}}_{a(i,j)}^{k + 1}} \right) + \varepsilon \left( {{\mathbb{P}}_{a(i,j)}^k,{\mathbb{P}}_{a(i,j)}^{k + 1}} \right)} \Big\}8i } 
	\label{theorem3_equ7}
	\end{aligned}
	\end{equation}
	
	\noindent where we also use ${\mathbb P}^{-1}_{a(i,j)}$ represent ${\mathcal{P}}_{a(i,j)}$. RHS of Eq.~\eqref{theorem3_equ7} is an upper bound to the target risk for a single task ${\mathcal{P}}_{a(i,j)}$ modelled by using the $c'_i$-th component. In the following, we consider all components $C'$ that are trained more than once~:
	\begin{equation}
	\begin{aligned}
	\sum\limits_{i = 1}^{|C'|} {\sum\limits_{j = 1}^{{{\tilde a}_i}}   {{\mathcal{R}}_{{\mathcal{P}_{a(i,j)}}}} \left( {h_{c'_i},{f_{{\mathcal{P}_{a(i,j)}}}}} \right)  } 
	&\le \sum\limits_{i = 1}^{|C'|} \sum\limits_{j = 1}^{{{\tilde a}_i}} \left\{ {{\mathcal{R}}_{{\mathbb{P}}_{a(i,j)}^{c(i,j)}}}\left( {h_{c'_i},h_{{\mathbb{P}}_{a(i,j)}^{c(i,j)}}^*} \right)  
	\right. \\
	& \left. 
	{+
		\sum\limits_{k =  -1}^{c(i,j) - 1} {\left( {disc_{\mathcal{L}}^\star \left( {{\mathbb{P}}_{a(i,j)}^k,{\mathbb{P}}_{a(i,j)}^{k + 1}} \right) + \varepsilon \left( {{\mathbb{P}}_{a(i,j)}^k,{\mathbb{P}}_{a(i,j)}^{k + 1}} \right)} \right)} } \right\} 
	\label{theorem3_equ8}
	\end{aligned}
	\end{equation}
	
	RHS of Eq.~\eqref{theorem3_equ8} is still an upper bound to the target risk of tasks modelled by the components that trained more than once. Therefore, ${\mathcal R}_{A'}$ in the paper, can be expressed by RHS of Eq.~\eqref{theorem3_equ8}, which proves Theorem 3.
	
	Based on the results from Theorem 3, we provide the additional analysis of the results of Theorem 3 in the following. Firstly, we rewrite Eq.(11) of the paper by~:
	\begin{equation}
	\begin{aligned}
	&\frac{1}{t}\sum\nolimits_{i = 1}^{ |C|}  \Big\{ {\mathcal{R}}_{{{\mathcal P}_{a_i}}} \left( h_{c_i},f_{{\mathcal P}_{a_i}} \right)   \Big\}
	+
	\frac{1}{t} \sum\nolimits_{i = 1}^{ |C'|} \sum\nolimits_{j = 1}^{{{\tilde a}_i}}  \Big\{ \mathcal{R}_{{\mathcal P}_{a(i,j)}}\left( {h_{c'_i},{f_{{{\mathcal P}_{a(i,j)}}}}} \right)  \Big\}
	\le  \sum\nolimits_{i = 1}^{|C|}
	\Big\{
	\mathcal{R}_{{\tilde{\mathcal{P}}_{a_i}}}\left( h_{c_i},h^{*}_{\tilde{\mathcal{P}}_{a_i}} \right) + {\mathcal R}_A \left({{\mathcal{P}_{a_i}},{\tilde{\mathcal{P}}_{a_i}}} \right)  \Big\} \\&+
	\sum\limits_{i = 1}^{|C'|} \sum\limits_{j = 1}^{{{\tilde a}_i}} \left\{ {{\mathcal{R}}_{\mathbb{P}_{a(i,j)}^{c(i,j)}}}\left( {h_{c'_i},h_{{\mathbb{P}}_{a(i,j)}^{c(i,j)}}^{*}} \right)
	{+
		\sum\limits_{k =  -1}^{c(i,j) - 1} {\left( {\mathcal R}_{A}\left( {{\mathbb{P}}_{a(i,j)}^k,{\mathbb{P}}_{a(i,j)}^{k + 1}} \right)
			\right)} } \right\}\,.
	\label{theorem3_analyse1}
	\end{aligned}
	\end{equation}
	
	We consider an extreme case where ${\bf M}$ only has a single component after LLL, ($|C'| = 1$ and $|C| = 0$). Then the first term in RHS of Eq.~\eqref{theorem3_analyse1} disappears and resulting in~:
	\begin{equation}
	\begin{aligned}
	&
	\frac{1}{t} \sum\nolimits_{i = 1}^{ |C'|} \sum\nolimits_{j = 1}^{{{\tilde a}_i}}  \mathcal{R}_{{\mathcal P}_{a(i,j)}}\left( {h_{c'_i},{f_{{{\mathcal P}_{a(i,j)}}}}} \right) 
	\le  
	\sum\limits_{i = 1}^{|C'|} \sum\limits_{j = 1}^{{{\tilde a}_i}} \left\{ {{\mathcal{R}}_{\mathbb{P}_{a(i,j)}^{c(i,j)}}}\left( {h_{c'_i},h_{{\mathbb{P}}_{a(i,j)}^{c(i,j)}}^{*}} \right)
	{+
		\sum\limits_{k =  -1}^{c(i,j) - 1} { \Big\{ {\mathcal R}_{A}\left( {{\mathbb{P}}_{a(i,j)}^k,{\mathbb{P}}_{a(i,j)}^{k + 1}} \right) \Big\}
	} } \right\}\,.
	\label{theorem3_analyse2}
	\end{aligned}
	\end{equation}
	
	In this case, learning early tasks ($a(i,j)$ is small) tends to increase the number of error terms more than when learning more recent tasks ($a(i,j)$ is large), This is caused by the number of accumulated error terms ${\mathcal{R}}_{A}(\cdot)$ controlled by the times of GR processes $c(i,j) = t-a(i,j)$. In the opposite case where the number of components ($K$) is equal to the number of tasks ($t$), $|C'| = 0$ and the mixture model has not accumulated errors. The GB for this case is~:
	\begin{equation}
	\begin{aligned}
	\frac{1}{t}\sum\nolimits_{i = 1}^{ |C|}   {\mathcal{R}}_{{{\mathcal P}_{a_i}}} \left( h_{c_i},f_{{\mathcal P}_{a_i}} \right)  
	\le  \sum\nolimits_{i = 1}^{|C|}
	\Big\{
	\mathcal{R}_{{\tilde{\mathcal{P}}_{a_i}}}\left( h_{c_i},h^{*}_{\tilde{\mathcal{P}}_{a_i}} \right) + {\mathcal R}_A \left({{\mathcal{P}_{a_i}},{\tilde{\mathcal{P}}_{a_i}}} \right) \Big\} \,.
	\end{aligned}
	\end{equation}
	
	\noindent where $|C| = K = t$. Then the lifelong learning problem is transformed to be the generalization problem under the generative modelling. This motivates us to propose a novel dynamic mixture model which would not accumulate errors during LLL.
	
	\section{The proof of Lemma 2}
	Similarly to the proof for Theorem 3, we firstly consider the components that are trained only once~:
	\begin{equation}
	\begin{aligned}
	{\mathcal{R}}_{C} &= \sum\limits_{i = 1}^{{|C|}} \Big\{
	{{\mathcal{R}}_{{\tilde{\mathcal{P}}_{a_i}}}}\left( {h_{c_i},h^*_{\tilde{\mathcal{P}}_{a_i}}} \right) + {\mathcal R}_A \left({{\mathcal{P}_{a_i}},{\tilde{\mathcal{P}}_{a_i}}} \right) \Big\},
	\label{Lemma2_eq1_1}
	\end{aligned}
	\end{equation}
	Then we add the KL divergence term and $D_{diff}$ term in Eq.~\eqref{Lemma2_eq1_1}, resulting in~:
	\begin{equation}
	\begin{aligned}
	{\mathcal{R}}_{C} &= \sum\limits_{i = 1}^{|C|} \Big\{
	{{\mathcal{R}}_{{\tilde{\mathcal{P}}_{a_i}}}}\left( {h_{c_i},h^*_{\tilde{\mathcal{P}}_{a_i}}} \right) + {\mathbb E}_{{\tilde{\mathcal{P}}}_{a_i}} KL\left( p_{c_i}({\bf z} \mid {\bf x}^S_{a_i}) \mid\mid p({\bf z}) \right) + D_{diff}\left({\bf x}^T_{a_i},{\bf x}^S_{a_i} \right) +  {\mathcal R}_A \left({{\mathcal{P}_{a_i}},{\tilde{\mathcal{P}}_{a_i}}} \right) \Big\},
	\label{Lemma2_eq1}
	\end{aligned}
	\end{equation}
	where $p_{c_i}({\bf z} \mid \cdot)$ represents the variational distribution modelled by the inference model of the $c_i$-th component. $D_{diff}(\cdot,\cdot)$ is defined as~:
	\begin{equation}
	\begin{aligned}
	D_{diff}({\bf x}^T_{a_i},{\bf x}^S_{a_i}) =  \Big| { {{\mathbb E}_{{{\mathcal P}_{a_i}  }}}KL(q_{\omega_{c_i}}({\bf{z}} \mid {\bf{ x}}_i^T) \mid\mid p({\bf{z}})) - 
		{{\mathbb E}_{{\tilde {\mathcal P}_{a_i}  }}}KL(q_{\omega_{c_i}}({\bf{z}} \mid {\bf{ x}}_i^S) \mid\mid p({\bf{z}}))
	} \Big|
	\end{aligned}
	\end{equation}
	
	We can rewrite Eq.~\eqref{Lemma2_eq1} as the negative ELBO form~:
	\begin{equation}
	\begin{aligned}
	{\mathcal{R}}_{C} &= \sum\limits_{i = 1}^{|C|} \Big\{ {\mathbb E}_{{ \tilde{\mathcal{P}}}_i} \{
	-\mathcal{L}_{ELBO}\left({\bf x}^S_{a_i} ;h_{c_i} \right) \} + D_{diff}\left({\bf x}^T_{a_i},{\bf x}^S_{a_i} \right) +  {\mathcal R}_A \left({{\mathcal{P}_{a_i}},{\tilde{\mathcal{P}}_{a_i}}} \right) \Big\} ,
	\label{Lemma2_eq2}
	\end{aligned}
	\end{equation}
	where ${\mathcal{L}}_{ELBO}(\cdot;h_{c_i})$ represents the ELBO estimated by the $c_i$-th component. Secondly, we consider the components that are trained more than once~:
	\begin{equation}
	\begin{aligned}
	{{\mathcal{R}}_{A'}} &= \sum\limits_{i = 1}^{|C'|} \sum\limits_{j = 1}^{{\tilde a}_i} \left\{ {{\mathcal{R}}_{{\mathbb{P}}_{a(i,j)}^{c(i,j)}}}\left( {h_{c'_i},{f_{{\mathbb{P}}_{a(i,j)}^{c(i,j)}}}} \right) 
	+ 
	{\mathcal R}_A\left({{{\mathcal{P}}_{a(i,j)}},{\mathbb{P}}_{a(i,j)}^{c(i,j)}}  \right)
	\right\},    
	\label{Lemma3_RA_eq1}
	\end{aligned}
	\end{equation}
	We then add the KL divergence term and $D_{diff}$ term in eq.(\ref{Lemma3_RA_eq1}), resulting in~:
	\begin{equation}
	\begin{aligned}
	{{\mathcal{R}}_{A'}} &= \sum\limits_{i = 1}^{|C'|} \sum\limits_{j = 1}^{{\tilde a}_i} \left\{ {{\mathcal{R}}_{{\mathbb{P}}_{a(i,j)}^{c(i,j)}}}\left( {h_{c'_i},{f_{{\mathbb{P}}_{a(i,j)}^{c(i,j)}}}} \right)  + {\mathbb E}_{{{\mathbb{P}}_{a(i,j)}^{c(i,j)}}}  KL\left( p_{c'_i}({\bf z} \mid {\bf x}^t_{a(i,j)}) \mid\mid p({\bf z}) \right)
	\right. \\
	& \left. 
	+ D_{diff}\left({\bf x}^T_{a(i,j)},{\bf x}^t_{a(i,j)} \right)
	+ 
	{\mathcal R}_A\left({{{\mathcal{P}}_{a(i,j)}},{\mathbb{P}}_{a(i,j)}^{c(i,j)}}  \right)
	\right\},    
	\label{Lemma3_RA_eq2}
	\end{aligned}
	\end{equation}
	\noindent where the conditional distribution $p_{c'_i}({\bf z}|{\bf x}^t_{a(i,j)})$ is modelled by the inference model of the $c'_i$-th component. Then we rewrite Eq.~\eqref{Lemma3_RA_eq2} as the negative ELBO form~:
	\begin{equation}
	\begin{aligned}
	{{\mathcal{R}}_{A'}} &= \sum\limits_{i = 1}^{|C'|} \sum\limits_{j = 1}^{{\tilde a}_i} {\mathbb E}_{{\mathbb P}^{c(i,j)}_{a(i,j)}} \left\{ 
	-\mathcal{L}_{ELBO}\left({\bf x}^t_{a(i,j)} ; h_{c'_i} \right) +
	D_{diff}\left({\bf x}^T_{a(i,j)},{\bf x}^t_{a(i,j)} \right)
	+ 
	{\mathcal R}_A\left({{{\mathcal{P}}_{a(i,j)}},{\mathbb{P}}_{a(i,j)}^{c(i,j)}}  \right)
	\right\},    
	\label{Lemma3_RA_eq3}
	\end{aligned}
	\end{equation}
	
	We summarize all KL divergence and $D_{diff}$ terms~:
	\begin{equation}
	\begin{aligned}
	&\sum\limits_{i = 1}^{|C|} \Big\{
	{\mathbb E}_{{\tilde{\mathcal{P}}}_{a_i}}
	KL\left( p_{c_i}({\bf z} \mid {\bf x}^S_{a_i}) \mid\mid p({\bf z}) \right)  + D_{diff}\left({\bf x}^T_{a_i},{\bf x}^S_{a_i} \right) \Big\} \\&+  \sum\limits_{i = 1}^{|C'|} \sum\limits_{j = 1}^{{\tilde a}_i} \left\{ {\mathbb E}_{{\mathbb P}^{c(i,j)}_{a(i,j)}}  KL\left( p_{c'_i}({\bf z} \mid {\bf x}^t_{a(i,j)}) \mid\mid p({\bf z}) \right) + D_{diff}\left({\bf x}^T_{a(i,j)},{\bf x}^t_{a(i,j)} \right)
	\right\},    
	\end{aligned}
	\end{equation}
	
	We also know that~:
	\begin{equation}
	\begin{aligned}
	\frac{1}{t} \sum\limits_{i = 1}^t 
	{\mathbb E}_{{{\mathcal{P}}}_{i}}
	KL\left( p({\bf z} \mid {\bf x}^T_{i}) \mid\mid p({\bf z}) \right) 
	& \le 
	\frac{1}{t}
	\sum\limits_{i = 1}^{|C|} \Big\{ {\mathbb E}_{{\tilde{\mathcal{P}}}_{a_i}}
	KL\left( p({\bf z} \mid {\bf x}^S_{a_i}) \mid\mid p({\bf z}) \right) + D_{diff}\left({\bf x}^T_{a_i},{\bf x}^S_{a_i} \right) \Big\} \\&+   
	\frac{1}{t}
	\sum\limits_{i = 1}^{|C'|} \sum\limits_{j = 1}^{{\tilde a}_i} \left\{ 
	{\mathbb E}_{{\mathbb P}^{c(i,j)}_{a(i,j)}}
	KL\left( p({\bf z} \mid {\bf x}^t_{a(i,j)}) \mid\mid p({\bf z}) \right) + D_{diff}\left({\bf x}^T_{a(i,j)},{\bf x}^t_{a(i,j)} \right)
	\right\}
	\label{lemma2_kl_eq1}
	\end{aligned}
	\end{equation}
	where we omit the subscript for $p{({\bf z} \mid \cdot)}$ for simplicity. We then consider the inequality from Eq.~\eqref{lemma2_kl_eq1} into Eq.~(11) of the paper and we have~:
	
	\begin{equation}
	\begin{aligned}
	&\frac{1}{t}{\sum\limits_{i = 1}^{|C|} {\mathbb E}_{{\mathcal{P}}_{a_i}} \Big\{ -{\mathcal{L}_{ELBO}\left({\bf x}^T_{a_i} ; h_{c_i} \right) }} \Big\}
	+
	\frac{1}{t}\sum\limits_{i = 1}^{|C'|} {\sum\limits_{j = 1}^{{{\tilde a}_i}}
		{\mathbb E}_{{\mathcal{P}}_{a(i,j)}}
		\Big\{ -{\mathcal{L}_{ELBO}\left({\bf x}^T_{a(i,j)};h_{c'_i} \right) }} \Big\}
	\le
	\frac{1}{t} \sum\limits_{i = 1}^{|C|} \Big\{ {\mathbb E}_{{\tilde{\mathcal{P}}}_{a_i}} \left\{  - {{{\cal L}_{ELBO}}\left( {{\bf{x}}_{{a_i}}^S};h_{c_i} \right)}
	\right\} \Big\}  \\&+ 
	\frac{1}{t} \sum\limits_{i = 1}^{|C'|} \sum\limits_{j = 1}^{|{A'_{c'_i}}|} \Big\{  {\mathbb E}_{{\mathbb P}^{c(i,j)}_{a(i,j)}} \left\{ { - {{\cal L}_{ELBO}}\left( {{\bf{x}}_{{a(i,j)}}^t};h_{c'_i} \right)
	}
	\right\} \Big\} + \frac{1}{t}\{ {\mathcal R}^{II}_{A'}+{{\mathcal{R}}_{C}^{II}} + D_{diff}*\}
	\end{aligned}
	\end{equation}
	This proves Lemma 2. We should also observe that $D_{diff}*$ is expressed by~:
	\begin{equation}
	\begin{aligned}
	D_{diff}* = \sum\limits_{i = 1}^{|C|} \Big\{
	D_{diff}\left({\bf x}^T_{a_i},{\bf x}^S_{a_i} \right) \Big\} +
	\sum\limits_{i = 1}^{|C'|} \sum\limits_{j = 1}^{{\tilde a}_i} \left\{  D_{diff}\left({\bf x}^T_{a(i,j)},{\bf x}^t_{a(i,j)} \right)
	\right\}, 
	\end{aligned}
	\end{equation}
	
	Based on the above results, in the following, we derive the risk bound of the mixture model to NLL~:
	
	\begin{equation}
	\begin{aligned}
	&\frac{1}{t}{\sum\limits_{i = 1}^{|C|} {\mathbb E}_{{\mathcal{P}}_{a_i}} \Big\{ -\log {p_{c_i}\left({\bf x}^T_{a_i} \right) }} \Big\}
	+
	\frac{1}{t}\sum\limits_{i = 1}^{|C'|} {\sum\limits_{j = 1}^{{{\tilde a}_i}} \Big\{ -{\log p_{c'_i}\left({\bf x}^T_{a(i,j)} \right) }} \Big\}
	\le
	\frac{1}{t} \sum\limits_{i = 1}^{|C|} \Big\{ {\mathbb E}_{{\tilde{\mathcal{P}}}_{a_i}} \left\{  - {{{\cal L}_{ELBO}}\left( {{\bf{x}}_{{a_i}}^S};h_{c_i} \right)}
	\right\} \Big\} \\&+  
	\frac{1}{t} \sum\limits_{i = 1}^{|C'|} \sum\limits_{j = 1}^{|{A'_{c'_i}}|} \Big\{ {\mathbb E}_{{\mathbb P}^{c(i,j)}_{a(i,j)}} \left\{ { - {{\cal L}_{ELBO}}\left( {{\bf{x}}_{{a(i,j)}}^t};h_{c'_i} \right)
	}
	\right\} \Big\}  + \frac{1}{t}\Big\{ {\mathcal R}^{II}_{A'}+{{\mathcal{R}}_{C}^{II}} + D_{diff}* \Big\}
	\end{aligned}
	\end{equation}
	
	\noindent where $\log p_{c_i}(\cdot)$ represents the sample log-likelihood (model likelihood) under the $c_i$-th component.
	
	\section{Generalize existing ELBOs to LELBO}
	
	In this section, we generalize previous well know ELBOs to LELBO.
	
	\subsection{Importance sampling}
	\label{importantSamplingSection}
	The main idea of Importance Weighted Autoencoder (IWELBO) \cite{IWVAE} is to allow the recognition network to generate multiple samples during the optimization leading to a better modelling of the posterior probabilities. The corresponding ELBO for sampling $K'$ samples is defined as~: 
	\begin{equation}
	\begin{aligned}
	{{\mathcal{L}}_{{ELBO}_{ K'}}}\left({\bf{x}};{\mathcal{M}} \right) = {{\mathbb E}_{{{\bf{z}}_1},..,{{\bf{z}}_{K'}} \sim q\left( {{\bf{z}}|{\bf{x}}} \right)}}\left[ {\log \frac{1}{K'}\sum\limits_{i = 1}^{K'} {\frac{{p\left( {{\bf{x}},{{\bf{z}}_i}} \right)}}{{q\left( {{{\bf{z}}_i}|{\bf{x}}} \right)}}} } \right]
	\label{IWVAE_sample_eq1}
	\end{aligned}
	\end{equation}
	where $K'$ is the number of weighted samples and $K'=1$ is equivalent to the standard ELBO. In order to calculate $w_i = p({\bf x},{\bf z}_i)/q({\bf z}_i| {\bf x})$ in practice, we rewrite $w_i$ as $\exp( {\log w_i })$. By calculating the right hand side of Eq.~\eqref{IWVAE_sample_eq1} requires to estimate each individual ${\mathbb E}_{z_i} \log w_i$ which is a standard ELBO.
	
	In the following, we extend this IWELBO to the LLL setting. From Lemma 1, we know that~:
	\begin{equation}
	\begin{aligned}
	\frac{1}{t} {\sum\nolimits_{i = 1}^t {\mathbb{E}}_{{\bf x}^T_i \sim {\mathcal{P}_i}}  \Big[-{\mathcal{L}_{ELBO}\left( {\bf x}^T_i ; h \right) }\Big] } &\le  
	{\mathbb{E}}_{{\bf x}^t \sim {\mathbb P}^{t-1}\otimes{\tilde {\mathcal{P}}}_t} \Big[-{\mathcal{L}_{ELBO}\left( {\bf \tilde x}^t ;h \right) }\Big] \\&+  \left| KL_1 - KL_2 \right| +  {\rm Err}^a + {\rm Err}^d.
	\label{importantce_lemma1_equ1}
	\end{aligned}
	\end{equation}
	where ${\mathcal{L}_{ELBO}(\cdot)}$ has the form according to \cite{VAE}. We can rewrite the above equation, by considering importance sampling, as~:
	\begin{equation}
	\begin{aligned}
	\frac{1}{t} {\sum\nolimits_{i = 1}^t {\mathbb{E}}_{{\bf x}^T_i \sim {\mathcal{P}_i}} \Big[ -{ \log p\left({\bf x}^T_i \right) }\Big] } &\le  
	{\mathbb{E}}_{{\bf x}^t \sim {\mathbb P}^{t-1}\otimes{\tilde {\mathcal{P}}}_t} \Big[-
	{{\mathbb E}_{{\bf{z}} \sim q\left( {{\bf{z}}\mid{\bf{x}}} \right)}}\left[ {\log \frac{{p\left( {{{\tilde{\bf{x}}}^t},{\bf{z}}} \right)}}{{q\left( {{\bf{z}}\mid{\bf{x}}} \right)}}} \right]
	\Big] \\&+  \left| KL_1 - KL_2 \right| +  {\rm Err}^a + {\rm Err}^d.
	\label{importantce_equ1}
	\end{aligned}
	\end{equation}
	
	According to $ - \log p({\bf{x}}) \le - {{\mathcal{L}}_{{ELBO}_{K' + 1}}}({\bf{x}};{\mathcal{M}} ) \le - {{\mathcal{L}}_{{ELBO}_{K'}}}({\bf{x}};{\mathcal{M}} )  $ \cite{IWVAE}, we have $  - {{\mathcal{L}}_{{ELBO}_{K' + 1}}}({\tilde {\bf{x}}^t};{\mathcal{M}} ) \le   - {{\mathcal{L}}_{{ELBO}_{K'}}}({ {\tilde {\bf{x}}}^t} ;{\mathcal{M}} ) $, based on the assumation that ${\mathbb P}^{t-1}$ is fixed. We note that $\mathcal{M}$ represents the model and $h$ is the hypothesis of $\mathcal{M}$. We assume that when the $h^\star$ is an optimal solution for $-\log p( {\tilde {\bf{x}}}^t)$ and we have $- \log p({\tilde {\bf{x}}}^t) \le  -{\mathcal{L}}_{ELBO}( {\tilde {\bf{x}}}^t; h^\star ) \le - {\mathcal{L}}_{{ELBO}_{K'}}({\tilde {\bf{x}}}^t;h)$. In the following, we rewrite Eq.~\eqref{importantce_lemma1_equ1} by using $h^\star$, resulting in~:
	\begin{equation}
	\begin{aligned}
	\frac{1}{t} {\sum\nolimits_{i = 1}^t {\mathbb{E}}_{{\bf x}^T_i \sim {\mathcal{P}_i}}  \Big[-{\mathcal{L}_{ELBO}\left( {\bf x}^T_i ; h^\star \right) }\Big] } &\le  
	{\mathbb{E}}_{{\bf x}^t \sim {\mathbb P}^{t-1}\otimes{\tilde {\mathcal{P}}}_t} \Big[-{\mathcal{L}_{ELBO}\left( {\bf \tilde x}^t ;h^\star \right) }\Big] \\&+  \left| KL_1 - KL_2 \right| +  {\rm Err}^a + {\rm Err}^d.
	\label{importantce_lemma1_equ1_optimal}
	\end{aligned}
	\end{equation}
	
	We observe that $|KL_1 - KL_2|$ of Eq.~\eqref{importantce_lemma1_equ1_optimal} is estimated by $h^\star$ ($\{ \theta_\star, \omega_\star \}$ are the corresponding model parameters). We can replace the first term in RHS of Eq.~\eqref{importantce_lemma1_equ1_optimal} by using ${\mathcal{L}}_{{ELBO}_{K'}}({\tilde {\bf{x}}}^t;h)$, resulting in~:
	\begin{equation}
	\begin{aligned}
	\frac{1}{t} {\sum\nolimits_{i = 1}^t {\mathbb{E}}_{{\bf x}^T_i \sim {\mathcal{P}_i}}  \Big[-{\mathcal{L}_{ELBO}\left( {\bf x}^T_i ; h^\star \right) }\Big] } &\le  
	{\mathbb{E}}_{{\bf x}^t \sim {\mathbb P}^{t-1}\otimes{\tilde {\mathcal{P}}}_t} \Big[-{\mathcal{L}}_{{ELBO}_{K'}}({\tilde {\bf{x}}}^t;h)\Big] \\&+  \left| KL_1 - KL_2 \right| +  {\rm Err}^a + {\rm Err}^d.
	\label{importantce_lemma1_equ1_optima2}
	\end{aligned}
	\end{equation}
	
	LHS of Eq.~\eqref{importantce_lemma1_equ1_optima2} is an upper bound to $\frac{1}{t} {\sum\nolimits_{i = 1}^t} {\mathbb{E}}_{{\bf x}^T_i \sim {\mathcal{P}_i}} [ -\log p_{\theta^\star}({\bf x}_i^T) ]$ and we rewrite Eq.~\eqref{importantce_lemma1_equ1_optima2} as~:
	\begin{equation}
	\begin{aligned}
	\frac{1}{t} {\sum\nolimits_{i = 1}^t} {\mathbb{E}}_{{\bf x}^T_i \sim {\mathcal{P}_i}} \Big[ -\log p_{\theta^\star}({\bf x}_i^T) \Big] &\le  
	{\mathbb{E}}_{{\bf x}^t \sim {\mathbb P}^{t-1}\otimes{\tilde {\mathcal{P}}}_t} \Big[-{\mathcal{L}}_{{ELBO}_{K'}}({\tilde {\bf{x}}}^t;h)\Big] \\&+  \left| KL_1 - KL_2 \right| +  {\rm Err}^a + {\rm Err}^d.
	\label{importantce_lemma1_equ1_optima3}
	\end{aligned}
	\end{equation}
	
	We then decompose the first term in RHS of Eq.~\eqref{importantce_lemma1_equ1_optima3}, and we have~:
	\begin{equation}
	\begin{aligned}
	\frac{1}{t} {\sum\nolimits_{i = 1}^t {\mathbb{E}}_{{\bf x}^T_i \sim {\mathcal{P}_i}} \Big[ -{ \log p_{\theta^\star} \left({\bf x}^T_i \right) }\Big] } &\le  
	{\mathbb{E}}_{{\bf x}^t \sim {\mathbb P}^{t-1}\otimes{\tilde {\mathcal{P}}}_t} \left[ - {\mathbb{E}_{{{\bf{z}}_1},\dots,{{\bf{z}}_{K'}}\sim q\left( {{\bf{z}}\mid {\bf{x}}} \right)}}\left[ {\log \frac{1}{K'}\sum\limits_{i = 1}^{K'} {\frac{{p_\theta \left( {{{\tilde {\bf{x}}}^t},{{\bf{z}}_i}} \right)}}{{q_\omega \left( {{{\bf{z}}_i}\mid {\bf{x}}} \right)}}} } \right]
	\right] \\&+  \left| KL_1 - KL_2 \right| +  {\rm Err}^a + {\rm Err}^d.
	\label{importantce_equ4}
	\end{aligned}
	\end{equation}
	
	If we do not consider ${\rm Err}^a$ and ${\rm Err}^d$ as in Lemma 1 (See details in Appendix~\ref{AppendixC}), Eq.~\eqref{importantce_equ4} can be~:
	\begin{equation}
	\begin{aligned}
	\frac{{\rm{1}}}{t}{\sum\limits_{i = 1}^t {\mathbb{E}}_{{\bf x}^T_i \sim {\mathcal{P}_i}} \left[ -{ \log p\left({\bf x}^T_i \right) }\right] } &\le
	{\mathbb{E}}_{{\tilde{\bf x}}^t \sim {\mathbb P}^{t-1}\otimes{\tilde {\mathcal{P}}}_t} \left[  - {\mathbb{E}_{{{\bf{z}}_{1}}\dots,{\bf{z}}_{K'} \sim q\left( {{\bf{z}}\mid {\bf{x}}} \right)}}\left[ {\log \frac{1}{K’}\sum\limits_{i = 1}^{K’} {\frac{{p\left( {{{\tilde {\bf{x}}}^t},{{\bf{z}}_i}} \right)}}{{q\left( {{{\bf{z}}_i}\mid {\bf{x}}} \right)}}} } \right]
	\right] \\
	& +  \left| KL_1 - KL_2 \right| + {\mathcal R}_A \left({{\mathcal{P}_{(1:t)}},{{{\mathbb{P}}^{t - 1}} \otimes {{\tilde {\mathcal{P}}}_t}}} \right)\,.
	\label{importantce_equ4_1}
	\end{aligned}
	\end{equation}
	where we omit the subscript (the model's parameters) for Eq.~\eqref{importantce_equ4_1} for the sake of simplification. It notes that $h \in \mathcal{H}$ is the model and its parameters are $\{\theta,\omega \}$ optimized by Eq.\eqref{importantce_equ4}. We call RHS of Eq.~\eqref{importantce_equ4} as $\mathcal{L}_{{LELBO}_{K'}}$ and when $K'=1$, $\mathcal{L}_{{LELBO}_{K'}}$ is equal to $\mathcal{L}_{LELBO}$ (RHS of Eq.~\eqref{importantce_equ1}). Based on the assumption that ${\mathbb P}^{t-1}\otimes{\tilde {\mathcal{P}}}_t$ is fixed, we have $\mathcal{L}_{{LELBO}_{K'+1}} \le \mathcal{L}_{{LELBO}_{K'}} $. We can observe that the tightness of ELBO on the marginal log-likelihood of the source distribution ${\mathbb P}^{t-1}\otimes{\tilde {\mathcal{P}}}_t$ can not guarantee a tight GB on the marginal log-likelihood of the target distribution. However, the tightness of GB is largely depending on the discrepancy distance between the evolved source and target distribution.
	
	\subsection{Hierarchical Variational Inference}
	
	In this section, we review the Hierarchical latent variable model and extend it to IELBO.
	
	\noindent \textBF{Auxiliary Deep Generative Models (ADGM) \cite{Aux_DGM}} ADGM is a classical hierarchical latent variable model. ADGM introduces an auxiliary variable $\bf a$ into the variational distribution $q({\bf a},{\bf z} \,|\, {\bf x}) = q({\bf z} \,|\, {\bf a},{\bf x})q({\bf a} \,|\, {\bf x})$ and its ELBO is expressed as~:
	\begin{equation}
	\begin{aligned}
	\log p\left( {\bf{x}} \right) = \log \iint {p\left( {{\bf{x}},{\bf{a}},{\bf{z}}} \right)d{\bf{a}}d{\bf{z}}}  \ge {{\mathbb E}_{q\left( {{\bf{a}},{\bf{z}}\mid {\bf{x}}} \right)}}\left[ {\log \frac{{p\left( {{\bf{a}}\mid{\bf{z}},{\bf{x}}} \right)p\left( {{\bf{x}}\mid{\bf{z}}} \right)p\left( {\bf{z}} \right)}}{{q\left( {{\bf{a}}\mid{\bf{x}}} \right)q\left( {{\bf{z}}\mid{\bf{a}},{\bf{x}}} \right)}}} \right] = {\mathcal{L}}_{ADGM}\left({\mathcal{M}}, {\bf x} \right)\,.
	\label{ADGM_ELBO}
	\end{aligned}
	\end{equation}
	
	We decompose $\mathcal{L}_{ADGM}({\bf x} ;{\mathcal{M}} )$ as~:
	\begin{equation}
	\begin{aligned}
	{{\cal L}_{ADGM}}\left({\mathcal{M}},{\bf{x}} \right) = {{\mathbb E}_{q\left( {{\bf{a}},{\bf{z}}\mid {\bf{x}}} \right)}}\left[ {\log p\left( {{\bf{x}}\mid {\bf{z}}} \right)} \right] + {{\mathbb E}_{q\left( {{\bf{a}},{\bf{z}}\mid {\bf{x}}} \right)}}\left[ {\log \frac{{p\left( {{\bf{a}}\mid {\bf{z}},{\bf{x}}} \right)p\left( {\bf{z}} \right)}}{{q\left( {{\bf{a}}\mid {\bf{x}}} \right)q\left( {{\bf{z}}\mid {\bf{a}},{\bf{x}}} \right)}}} \right]
	\label{ADGM_ELBO2}
	\end{aligned}
	\end{equation}
	
	We consider the following inequality~:
	\begin{equation}
	\begin{aligned}
	&\frac{1}{t}\sum\limits_{i = 1}^t
	{\mathbb{E}}_{{\bf x}^T_i \sim \mathcal{P}_i}
	{{{\mathbb E}_{q\left( {{\bf{a}},{\bf{z}}\mid {\bf{x}}_i^T} \right)}}\left[ {\log \frac{{p\left( {{\bf{a}}\mid {\bf{z}},{\bf{x}}_i^T} \right)p\left( {\bf{z}} \right)}}{{q\left( {{\bf{a}}\mid {\bf{x}}_i^T} \right)q\left( {{\bf{z}}\mid {\bf{a}},{\bf{x}}_i^T} \right)}}} \right]}  \le {\mathbb E}_{ \tilde{{\bf x}}^t \sim {\mathbb P}^{t-1} \otimes {\tilde{\mathcal{P}}}_i } {{\mathbb E}_{q\left( {{\bf{a}},{\bf{z}}\mid {{\bf{x}}^t}} \right)}}\left[ {\log \frac{{p\left( {{\bf{a}}\mid {\bf{z}},{ \tilde{\bf{x}}^t}} \right)p\left( {\bf{z}} \right)}}{{q\left( {{\bf{a}}\mid { \tilde{\bf{x}}^t}} \right)q\left( {{\bf{z}}\mid {\bf{a}},{ \tilde{\bf{x}}^t}} \right)}}} \right] \\&+ \left| {\mathbb E}_{ {\tilde{\bf x}}^t \sim {\mathbb P}^{t-1} \otimes {\tilde{\mathcal{P}}}_i } {{{\mathbb E}_{q\left( {{\bf{a}},{\bf{z}}\mid {\tilde{\bf{x}}^t}} \right)}}\left[ {\log \frac{{p\left( {{\bf{a}}\mid {\bf{z}},{\tilde{\bf{x}}^t}} \right)p\left( {\bf{z}} \right)}}{{q\left( {{\bf{a}}\mid {\tilde{\bf{x}}^t}} \right)q\left( {{\bf{z}}\mid {\bf{a}},{\tilde{\bf{x}}^t}} \right)}}} \right] - \frac{1}{t}\sum\limits_{i = 1}^t {\mathbb{E}}_{{\bf x}^T_i \sim \mathcal{P}_i} {{{\mathbb E}_{q\left( {{\bf{a}},{\bf{z}}\mid {\bf{x}}_i^T} \right)}}\left[ {\log \frac{{p\left( {{\bf{a}}\mid {\bf{z}},{\bf{x}}_i^T} \right)p\left( {\bf{z}} \right)}}{{q\left( {{\bf{a}}\mid {\bf{x}}_i^T} \right)q\left( {{\bf{z}}\mid {\bf{a}},{\bf{x}}_i^T} \right)}}} \right]} } \right|\,.
	\label{ADGM_ELBO2_KL}
	\end{aligned}
	\end{equation}
	
	We name the second term of RHS of Eq.~\eqref{ADGM_ELBO2_KL} as $|{\bf E}^*_1 - {\bf E}^*_2 |$. By considering Lemma 1, we have~:
	\begin{equation}
	\begin{aligned}
	\left(1/t \right){\sum\nolimits_{i = 1}^t {\mathbb{E}}_{{\bf x}^T_i \sim {\mathcal{P}_i}}  \Big[-{\mathcal{L}_{ADGM}\left({\bf x}^T_i \right) }\Big] } &\le  
	{\mathbb{E}}_{\tilde{\bf x}^t \sim {\mathbb P}^{t-1}\otimes{\tilde {\mathcal{P}}}_t} \Big[-{{\mathbb E}_{q\left( {{\bf{a}},{\bf{z}}\mid  {\tilde{\bf{x}}^t}} \right)}}\left[ {\log \frac{{p\left( {{\bf{a}}\mid{\bf{z}},{\tilde{\bf{x}}}^t} \right)p\left( {{\tilde{\bf{x}}^t}\mid{\bf{z}}} \right)p\left( {\bf{z}} \right)}}{{q\left( {{\bf{a}}\mid{\tilde{\bf{x}}^t}} \right)q\left( {{\bf{z}}\mid{\bf{a}},{\tilde{\bf{x}}^t}} \right)}}} \right]\Big] \\&+  \left|{\bf E}^*_1 - {\bf E}^*_2 \right| +  {\rm Err}^a + {\rm Err}^d.
	\end{aligned}
	\end{equation}
	
	\noindent \textBF{Importance Weighted Hierarchical Variational
		Inference (IWHVI) \cite{ImportanceHVAE}.} IWHVI introduces a multisample generalization for the hierarchical variable model by using two auxiliary distributions, $q({\bf{\varsigma }} \,|\, {\bf{z}})$, $\tau ({\bf{\psi }} \,|\, {\bf{z}},{\bf{x}})$ and its ELBO is expressed as~:
	\begin{equation}
	\begin{aligned}
	\log p\left( {\bf{x}} \right) \ge {{\mathbb E}_{q\left( {{\bf{z}}\mid{\bf{x}}} \right)}}\log \frac{{p\left( {{\bf{x}},{\bf{z}}} \right)}}{{q\left( {{\bf{z}} \mid {\bf{x}}} \right)}} \ge {{\mathbb E}_{q\left( {{\bf{z}},{{\bf{\psi }}_0} \mid {\bf{x}}} \right)}}{{\mathbb E}_{\tau \left( {{{\bf{\psi }}_{1:K}} \mid {\bf{z}},{\bf{x}}} \right)}}{{\mathbb E}_{q\left( {{{\bf{\zeta }}_{1:L}} \mid {\bf{z}}} \right)}}\log \frac{{p\left( {{\bf{x}} \mid {\bf{z}}} \right)\frac{1}{L}\sum\limits_{k = 1}^L {\frac{{p\left( {{\bf{z}},{{\bf{\zeta }}_k}} \right)}}{{p\left( {{{\bf{\zeta }}_k},{\bf{z}}} \right)}}} }}{{\frac{1}{{K + 1}}\sum\limits_{k = 0}^K {\frac{{q\left( {{\bf{z}},{{\bf{\psi }}_k}\mid {\bf{x}}} \right)}}{{\tau \left( {{{\bf{\psi }}_k} \mid {\bf{z}},{\bf{x}}} \right)}}} }} = {\mathcal{L}}_{IWHVI}\left({\mathcal{M}},{\bf x} \right) \,.
	\label{IWHVI_equ1}
	\end{aligned}
	\end{equation}
	
	From Eq.~\eqref{IWHVI_equ1}, we have $ -\log p(\bf x) \le -\mathcal{L}_{ELBO}({\bf x};{\mathcal{M}} ) \le - {\mathcal{L}}_{IWHVI}({\bf x};{\mathcal{M}} )$. Therefore, according to Lemma 1, we generalize Eq.~\eqref{IWHVI_equ1} to LLL setting~:
	\begin{equation}
	\begin{aligned}
	\frac{1}{t} {\sum\nolimits_{i = 1}^t {\mathbb{E}}_{{\bf x}^T_i \sim {\mathcal{P}_i}} \{ -{ \log p\left({\bf x}^T_i \right) }\} } &\le  
	{\mathbb{E}}_{\tilde{\bf x}^t \sim {\mathbb P}^{t-1}\otimes{\tilde {\mathcal{P}}}_t} \{ -{\mathcal{L}}_{IWHVI}\left(\tilde{\bf x}^t ; h \right)
	\} \\&+  \left| KL_1 - KL_2 \right| +  {\rm Err}^a + {\rm Err}^d \,.
	\label{IWHVI_IELBO}
	\end{aligned}
	\end{equation}
	
	Additionally, Eq~\eqref{IWHVI_IELBO} can be extended to the IWELBO bound~:
	
	\begin{equation}
	\begin{aligned}
	\left(1/t \right){\sum\nolimits_{i = 1}^t {\mathbb{E}}_{{\bf x}^T_i \sim {\mathcal{P}_i}} \{ -{ \log p\left({\bf x}^T_i \right) }\} } &\le  
	{\mathbb{E}}_{{\bf x}^t \sim {\mathbb P}^{t-1}\otimes{\tilde {\mathcal{P}}}_t}  \left\{  -
	{\mathbb E}\log \frac{1}{{K'}}\sum\limits_{i = 1}^{K'} {\frac{{p\left( {{\bf{x}}|{{\bf{z}}_i}} \right)\frac{1}{L}\sum\limits_{j = 1}^L {\frac{{p\left( {{{\bf{z}}_i},{{\bf{\zeta }}_{\left( {i.j} \right)}}} \right)}}{{q\left( {{{\bf{\zeta }}_{\left( {i.j} \right)}}\mid {{\bf{z}}_i}} \right)}}} }}{{\frac{1}{{K + 1}}\sum\limits_{k = 0}^K {\frac{{q\left( {{{\bf{z}}_i},{{\bf{\psi }}_{\left( {i,k} \right)}}\mid {\bf{x}}} \right)}}{{\tau \left( {{{\bf{\psi }}_{\left( {i,k} \right)}}\mid {\bf{x}},{{\bf{z}}_i}} \right)}}} }}} 
	\right\} \\&+  \left| KL_1 - KL_2 \right| +  {\rm Err}^a + {\rm Err}^d \,.
	\label{IWHVI_IELBO_importance}
	\end{aligned}
	\end{equation}
	
	We call the first term of RHS of Eq.~\eqref{IWHVI_IELBO_importance} as ${\mathcal{L}}_{{LIWHVI}_{K'}}$. Each ${{{\bf{\psi }}_{\left( {i,0} \right)}}}$ is sampled from $q({{{\bf{\psi }}}}\mid {\bf x})$ and each ${{{\bf{\psi }}_{\left( {i,k} \right)}}}$ is sampled from ${\tau \left( {{{\bf{\psi }}}\mid {\bf{z}}_i,{\bf{x}}} \right)}$. ${{{\bf{\zeta }}_{\left( {i.j} \right)}}}$ is sampled from $q({{{\bf{\zeta }}}}\mid {\bf z}_i)$ and ${\bf z}_i$ is sampled from $q({\bf z} \mid {\bf x},{{{\bf{\psi }}_{\left( {i,0} \right)}}})$. Based on the assumption that ${\mathbb P}^{t-1}$ is fixed and $| KL_1 - KL_2 | = 0$, we have ${\mathcal{L}}_{{LIWHVI}_{K'}} \ge {\mathcal{L}}_{{LIWHVI}_{K'+1}}$.
	
	\noindent \textBF{Ladder Variational Autoencoders (LVA) \cite{Ladder_VAE}. } LVA introduces a new inference model into the VAE framework, which allows to generate the latent variable ${\bf z}_i$ in each $i$-th layer and its approximate posterior is expressed as~:
	\begin{equation}
	\begin{aligned}
	q\left( {{\bf{z}}\mid {\bf{x}}} \right) = q\left( {{{\bf{z}}_1}\mid {\bf{x}}} \right)\prod\limits_{i = 2}^L {q\left( {{{\bf{z}}_i}\mid {{\bf{z}}_{i - 1}}} \right)}\,. 
	\label{lva_posterior}
	\end{aligned}
	\end{equation}
	
	We can also extend ELBO frm LVAE to the LLL setting, resulting in~:
	\begin{equation}
	\begin{aligned}
	\frac{1}{t} {\sum\nolimits_{i = 1}^t {\mathbb{E}}_{{\bf x}^T_i \sim {\mathcal{P}_i}}  \{-{\mathcal{L}_{ELBO}\left( {\bf x}^T_i ; h \right) }\} } &\le  
	{\mathbb{E}}_{{\bf x}^t \sim {\mathbb P}^{t-1}\otimes{\tilde {\mathcal{P}}}_t} \{-{\mathcal{L}_{LVA}\left( {\bf \tilde x}^t ; h \right) }\} \\&+  \left| KL_1 - KL_2 \right| +  {\rm Err}^a + {\rm Err}^d.
	\end{aligned}
	\end{equation}
	
	We use the approximate posterior (Eq.~\eqref{lva_posterior}) in $|KL_1 - KL_2  |$.
	
	\section{Forgetting behaviour of other generative models}
	
	In this section, we extend the proposed GB to other types of generative models including GANs and Energy-based GANs.
	
	\subsection{Generative Adversarial Nets (GANs)}
	
	The discrepancy distance has been used in GANs \cite{DiscrepancyGAN} but its usage has not been explored within the LLL setting. Following from \cite{DiscrepancyGAN}, we define ${{\mathcal{L}}_{\mathcal H}}{\rm{ = \{ }}{\mathcal{L}}(h({\bf{x}}),h'({\bf{x}})):h,h' \in {\mathcal{H}}{\rm{\} }}$ as the family of discriminators which is used in the discrepancy distance (See Definition 2 in the paper). The errors of the model on the target distribution are bounded, as in \cite{DiscrepancyGAN}~:
	\begin{equation}
	\begin{aligned}
	{{\mathcal{R}}_{{\mathcal{P}_i}}} \left( h ,{f_{\mathcal{P}_i}} \right) &\le  {{\mathcal{R}}_{{\mathbb P}^1}}\left( {h,{f_{\mathcal{P}_i}} } \right) + dis{c^{\star}_{\mathcal{L}}}\left( {{\mathcal{P}_1},{{\mathbb P}^i}} \right) \,,
	\label{gan_eq1}
	\end{aligned}
	\end{equation}
	where we assume that ${f_{\mathcal{P}_i}}$ is the true labeling function for the $i$-th task. The detailed proof can be found in \cite{DiscrepancyGAN}. ${\mathbb P}^1$ represents the generator distribution of a GAN model trained on the $i$-th task learning. However, Eq.~\eqref{gan_eq1} is only applied for a single domain. In the following, we generalize Eq.~\eqref{gan_eq1} to the LLL setting, inspired by the proposed GB.
	
	In order to overcome the forgetting, GANs are trained in a Self-Supervised (GR process) fashion in which the generator and discriminator are retrained on its generations. In this case, the previously generated samples and samples from the current task are treated as real images while the generated images during the training are treated as fake images. In the following, we derive a GB between the target distribution and the evolved source distribution.
	
	\begin{proposition} Let a GAN model be trained on a sequence of $t$ tasks by using the GR process, then we derive a GB for this GAN between the target and the evolved source distribution during the $t$-th task learning~:
		\begin{equation}
		\begin{aligned}
		{{\mathcal{R}}_{{\mathcal{P}_{\left(1:t \right)}}}} \left( h ,{f_{\mathcal{P}_{\left (1:t \right)}}} \right) &\le  {{\mathcal{R}}_{{{\mathbb P}^{t}}}}\left( {h,{f_{\mathcal{P}_{\left (1:t \right)}}} } \right) + dis{c^{\star}_{\mathcal{L}}}\left( {{\mathcal{P}_{\left(1:t \right)}},{{\mathbb P}^{t}} } \right) \,,
		\label{GAN_bound1}
		\end{aligned}
		\end{equation}
	\end{proposition}
	where ${\mathcal{P}}_{(1:t)}$ represents the mixture distribution formed by samples draw from $\{ {\mathcal{P}}_1,\dots,{\mathcal{P}}_t \}$. ${\mathbb P}^t$ represents the generator distribution of a GAN model trained on the $t$-th task learning. In the following, we derive a GB to exhibit how a GAN model lose previously learnt knowledge for each task learning.
	
	\begin{proposition}
		For a given a sequence of $t$ tasks, a GAN model is trained with the GR process. We derive a GB on the risk of all tasks between the target distribution and the evolved distribution at the $t$-th task learning~:
		\begin{equation}
		\begin{aligned}
		\sum\limits_{k = 1}^t {{{\cal R}_{{{\cal P}_k}}}\left( {h,{f_{{{\cal P}_k}}}} \right)} {\rm{ }} \le \sum\limits_{k = 1}^t {\left\{ {{{\cal R}_{{\mathbb P}_k^t}}\left( {h,{f_{{\mathcal P}_k}}} \right) + \sum\limits_{j = k}^t {\{ disc_{\cal L}^ \star \left( {{\mathbb P}_k^{j - 1},{\mathbb P}_k^j} \right)\} }  + disc_{\cal L}^ \star \left( {{\mathbb P}_k^{k - 1},{{\cal P}_i}} \right){\mkern 1mu} } \right\}} ,
		\label{GAN_bound6}
		\end{aligned}
		\end{equation}
	\end{proposition}
	
	Eq.~\eqref{GAN_bound6} explicitly measures the degenerated performance for each task learning, caused by the discrepancy distance between two distributions ${\mathbb P}^{j-1}_i$ and ${\mathbb P}^{j}_i$. From Eq.~\eqref{GAN_bound6}, we find that as learning more tasks ($t$ is increased), The generalization performance of the GAN model would be gradually degenerated since the discrepancy distance terms are increased (The second term in the right hand side of Eq.~\eqref{GAN_bound6}.

	\noindent \textBF{Proof.} 
	In order to measure the forgetting behaviour of a GAN for each task learning, we need to define the individual approximation distribution related to each task. Let us define the approximation distribution ${\mathbb P}^j_i$ formed by the sampling process ${\bf x} \sim {\mathbb P}^j$ if $I_{\mathcal{T}}({\bf x}) = i$. ${\mathbb P}^j_i$ is the probabilistic representation of the generated data related to the $i$-th task where $j$ represents that the GAN model has been trained on $j$ number of tasks. we use ${\mathbb P}^{(i-1)}_i$ represent ${\tilde{\mathcal{P}}}_i$ for simplicity. Therefore, for the $i$-th task, we can have the following bound~:
	\begin{equation}
	\begin{aligned}
	{{\mathcal{R}}_{{\mathbb P}^{(i-1)}_i}} \left( {h,f_{{\mathcal P}_i }} \right) &\le  {{\mathcal{R}}_{{\mathbb P}^{i}_i}}\left( {h,f_{{\mathcal P}_i }} \right) + dis{c^{\star}_{\mathcal{L}}}\left( {{\mathbb P}^{i}_i,{\mathbb P}^{(i-1)}_i} \right) \,,
	\end{aligned}
	\end{equation}
	and
	\begin{equation}
	\begin{aligned}
	{{\mathcal{R}}_{{\mathbb P}^i_i}} \left( {h,f_{{\mathcal P}_i }} \right) &\le  {{\mathcal{R}}_{{\mathbb P}^{(i+1)}_i}}\left( {h,f_{{\mathcal P}_i }} \right) + dis{c^{\star}_{\mathcal{L}}}\left( {{\mathbb P}^{(i+1)}_i,{\mathbb P}^i_i} \right) \,,
	\label{GAN_bound2}
	\end{aligned}
	\end{equation}
	
	In the following, we treat ${\mathbb P}^{(i+1)}_i$ as the target distribution and ${\mathbb P}^{(i+2)}_i$ as the source distribution, we have~:
	
	\begin{equation}
	\begin{aligned}
	{{\mathcal{R}}_{{\mathbb P}^{(i+1)}_i}} \left( {h,f_{{\mathcal P}_i }} \right) &\le  {{\mathcal{R}}_{{\mathbb P}^{(i+2)}_i}}\left( {h,f_{{\mathcal P}_i }} \right) + dis{c^{\star}_{\mathcal{L}}}\left( {{\mathbb P}^{(i+2)}_i,{\mathbb P}^{(i+1)}_i} \right) \,,
	\end{aligned}
	\end{equation}
	
	We repeat this process, resulting in~:
	\begin{equation}
	\begin{aligned}
	{{\mathcal{R}}_{{\mathbb P}^{((i+2)}_i}} \left( {h,f_{{\mathcal P}_i }} \right) &\le  {{\mathcal{R}}_{{\mathbb P}^{((i+3)}_i}}\left( {h,f_{{\mathcal P}_i }} \right) + dis{c^{\star}_{\mathcal{L}}}\left( {{\mathbb P}^{(i+3)}_i,{\mathbb P}^{(i+2)}_i} \right) 
	\\
	\dots
	\\
	\dots
	\\
	{{\mathcal{R}}_{{\mathbb P}^{t-1}_i}} \left( {h,f_{{\mathcal P}_i }} \right) &\le  {{\mathcal{R}}_{{\mathbb P}^t_i}}\left( {h,f_{{\mathcal P}_i }} \right) + dis{c^{\star}_{\mathcal{L}}}\left( {{\mathbb P}^t_i,{\mathbb P}^{t-1}_i} \right) 
	\end{aligned}
	\end{equation}
	
	We then sum up all inequalities, resulting in~:
	\begin{equation}
	\begin{aligned}
	{{\mathcal{R}}_{{\mathbb P}^{(i-1)}_i}} \left( {h,f_{{\mathcal P}_{i} }}\right) &\le  {{\mathcal{R}}_{{\mathbb P}^t_i}}\left( {h,{f_{{\mathcal P}_i }} } \right) + \sum\limits_{j = i}^{t} {disc_{\mathcal L}^{\star}\left( {{\mathbb P}_i^{j - 1},{\mathbb P}_i^j} \right)}  \,,
	\label{GAN_bound3}
	\end{aligned}
	\end{equation}
	
	We can observe that the left hand side of Eq.~\eqref{GAN_bound3} is also an upper bound to the target risk of the model at the $i$-th task~:
	
	\begin{equation}
	\begin{aligned}
	{{\mathcal{R}}_{{\mathcal P}_i}} \left( h ,{f_{\mathcal{P}_i}} \right) &\le  {{\mathcal{R}}_{{\mathbb P}^{(i-1)}_i}}\left( {h,{f_{{\mathcal P}_i }} } \right)  + dis{c^{\star}_{\mathcal{L}}}\left( {{\mathbb P}^{(i-1)}_i,{\mathcal P}_i} \right)  \,,
	\label{GAN_bound4}
	\end{aligned}
	\end{equation}
	
	By comparing Eq.~\eqref{GAN_bound4} and Eq.~\eqref{GAN_bound3}, we have a GB for the $i$-th task~:
	
	\begin{equation}
	\begin{aligned}
	{{\mathcal{R}}_{{\mathcal P}_i}} \left( h ,{f_{\mathcal{P}_i}} \right) &\le  {{\mathcal{R}}_{{\mathbb P}^t_i}}\left( {h,{f_{{\mathcal P}_i }} } \right) + \sum\limits_{j = i}^t \{ {disc_{\mathcal L}^{\star}\left( {{\mathbb P}_i^{j - 1},{\mathbb P}_i^j} \right)}\} + dis{c^{\star}_{\mathcal{L}}}\left( {{\mathbb P}^{i-1}_i,{\mathcal P}_i} \right)  \,,
	\label{GAN_bound5}
	\end{aligned}
	\end{equation}
	
	Then we can easily obtain a GB for all tasks based on Eq.~\eqref{GAN_bound5}.
	
	\begin{equation}
	\begin{aligned}
	\sum\limits_{k = 1}^t {{{\cal R}_{{{\cal P}_k}}}\left( {h,{f_{{{\cal P}_k}}}} \right)} {\rm{ }} \le \sum\limits_{k = 1}^t {\left\{ {{{\cal R}_{{\mathbb P}_k^t}}\left( {h,{f_{{\mathcal P}_k}}} \right) + \sum\limits_{j = k}^t {\{ disc_{\cal L}^ \star \left( {{\mathbb P}_k^{j - 1},{\mathbb P}_k^j} \right)\} }  + disc_{\cal L}^ \star \left( {{\mathbb P}_k^{k - 1},{{\cal P}_i}} \right){\mkern 1mu} } \right\}} ,
	\end{aligned}
	\end{equation}
	
	This proves Proposition 2.
	
	\subsection{Energy-based GANs}
	
	Energy function was firstly used in GANs \cite{EBGAN} in 2016, called EBGAN. Different from GANs, EBGAN introduces using a discriminator consisting of an auto-encoder to calculate the energy value which is the reconstruction loss. It shows that EBGAN assigns higher energy to fake images and low energy to real images. We implement $h \in {\mathcal{H}}$ as the discriminator. EBGAN in lifelong learning can be seen as an advanced Self-Supervised VAEs without the KL divergence term since its generator can produce sharper images than VAEs. Then we can derive a GB for EBGAN, similar to that defined by Theorem 2.
	\begin{equation}
	\begin{aligned}
	\frac{1}{t} \sum\nolimits_{i = 1}^t {{{\mathcal{R}}_{{\mathcal{P}_i}}}} \left( {h,{f_{{\mathcal{P}_i}}}} \right)&\le {{\mathcal{R}}_{{{\mathbb{P}}^{t - 1}} \otimes {{\tilde {\mathcal{P}}}_t}}}\left( {h,h_{{{\mathbb{P}}^{t - 1}} \otimes {{\tilde{\mathcal{P}}}_t}}^*} \right) + {\mathcal R}_A \left({{\mathcal{P}_{(1:t)}},{{{\mathbb{P}}^{t - 1}} \otimes {{\tilde {\mathcal{P}}}_t}}} \right),
	\end{aligned}
	\end{equation}
	
	Different from Theorem 2, ${{{\mathbb{P}}^{t - 1}}}$ is approximated by the generator of EBGAN trained on the $(t-1)$ number of tasks. Since EBGAN does not have the KL regularization and therefore the generator tends to generate more realistic samples when comparing with the generator of VAEs. In addition to the EBGAN, the proposed theory framework can be used to analyze the forgetting behaviour for the existing generative models \cite{Infogan,GAN_Maximum,InfoVAEGAN_conference,JontLatentVAEs,DeepMixtureVAE,MixtureOfVAEs} and lifelong learning approaches \cite{Lifelong_GAN,LifelongMixuteOfVAEs,LifelongInfinite,Lifelong_expandable,LifelongTwin,LifelongTeacherStudent,LifelongVAEGAN,Lifelonginterpretable}. We will investigate these methods in the future work.
	
	\section{Risk bound estimation from finite samples}
	
	In this section, we introduce how to estimate the risk bound from finite samples.
	
	\begin{definition}
		\textBF{(Rademacher complexity).} Let $\mathcal{H}$ represent a hypothesis class, For a given
		unlabeled sample $U = \{ {{\bf{x}}_i}\} _{i = 1}^m$, the Rademacher complexity of $\mathcal{H}$ with respect to the sample $U$ is defined as follows~:
		\begin{equation}
		\begin{aligned}
		{{\mathop{\rm Re}\nolimits} _U}\left( \mathcal{H} \right) = \mathop \mathbb{E} \limits_{\mathcal{K}} \left[ {\mathop {\sup }\limits_{h \in \mathcal{H}} \frac{2}{m}\sum\limits_{i = 1}^m {{{\mathcal{K}}_i}h\left( {{{\bf{x}}_i}} \right)} } \right]
		\end{aligned}
		\end{equation}
		where $\mathcal{K}_i$ is an independent uniform random variable within $\{-1,+1 \}$. The Rademacher complexity for the whole hypothesis class is defined as~:
		\begin{equation}
		\begin{aligned}
		{{\mathop{\rm Re}\nolimits} _n}\left( \mathcal{H} \right) = {\mathbb{E}_{U \sim \left( {{D}} \right)^n}}{{\mathop{\rm Re}\nolimits} _U}\left(\mathcal{H} \right)
		\end{aligned}
		\end{equation}
	\end{definition}

	\textBF{Lemma 3}
	\label{lemma5}
	Let $\mathcal{H}$ represent a hypothesis class and let ${\mathcal{L}_{\mathcal{H}}} = \{ \mathcal{X} \to \mathcal{L}(h'({\bf{x}}),h({\bf{x}})),(h',h) \in \mathcal{H}\} $. Let $\mathcal{L} : \mathcal{X} \times \mathcal{X} \to {\mathbb{R}_ + }$ be a loss function, satisfying $\forall ({\bf{x}},{\bf{x'}}) \in \mathcal{X},\mathcal{L}({\bf{x}},{\bf{x'}}) > M$ and $M$ is the positive value. Let $\mathcal{P}$ be a distribution over $\mathcal{X}$ and let $\tilde{\mathcal{P}}$ be the corresponding empirical distribution formed by a sample $U=\{{{\bf x}_1,\dots,{\bf x}_m}\}$. Then for any $\delta  \in (0,1)$, with the probability $1-\delta $, we have~:
	\begin{equation}
	\begin{aligned}
	disc_\mathcal{L}\left( {\mathcal{P},{\tilde{\mathcal{P}}}} \right) \le {{\mathop{\rm Re}\nolimits} _U}\left( {{\mathcal{L}_{\mathcal{H}}}} \right) + 3M\sqrt {\frac{{\log \frac{2}{\delta }}}{{2m}}} 
	\end{aligned}
	\end{equation}
	From Lemma 3, we can generalize the bound in case of more general loss function $\forall ({\bf{x'}},{\bf{x}}) \in {\mathcal{X}^2},{\mathcal{L}_q}({\bf{x'}},{\bf{x}}) = |{\bf{x'}} - {\bf{x}}{|^q}$ for some $q$. In the following, we show how to estimate the discrepancy distance by using the finite number of samples according to \cite{domainTheory}.
	
	\begin{corollary}
		\label{corollary1}
		Let $\mathcal{P}$ and $\mathbb{P}$ represent two domains over $\mathcal{X}$, respectively. Let $U_{\mathcal{P}}$ and $U_{\mathbb{P}}$ represent samples of size $m_{\mathcal{P}}$ and $m_{\mathbb{P}}$, drawn independently from $U_{\mathcal{P}}$ and $U_{\mathbb{P}}$. Let $\hat{\mathcal{P}}$ and $\hat{\mathbb{P}}$ represent the empirical distributions for $U_{\mathcal{P}}$ and $U_{\mathbb{P}}$. let ${\mathcal{L}}({\bf{x'}},{\bf{x}}) = |{\bf{x'}} - {\bf{x}}{|^2}$ be a loss function, satisfying $\forall ({\bf{x}},{\bf{x'}}) \in \mathcal{X},\mathcal{L}({\bf{x}},{\bf{x'}}) > M$ and $M$ is the positive value. Then with probability $1-\delta$, we have~:
		\begin{equation}
		\begin{aligned}
		disc_\mathcal{L}\left( {\mathcal{P},\mathbb{P}} \right) &\le disc_{\mathcal{L}}\left( {\hat{\mathcal{P}},\hat{\mathbb{P}}} \right) + 4q\left( {{{\mathop{\rm Re}\nolimits} }_{{U_{\mathcal P}}}}\left(\mathcal{H} \right) 
		+
		{{{\mathop{\rm Re}\nolimits} }_{{U_{\mathbb{P}}}}}\left(\mathcal{H} \right) \right) \\&+ 3M\left( {\sqrt {\frac{{\log \left( {\frac{4}{\delta }} \right)}}{{2{m_{\mathcal{P}}}}}}  + \sqrt {\frac{{\log \left( {\frac{4}{\delta }} \right)}}{{2{m_{\mathbb P}}}}} } \right)
		\label{colorllary_eq1}
		\end{aligned}
		\end{equation}
		
		We use $disc_\mathcal{L}^\star( {\mathcal{P},\mathbb{P}} )$ to represent the right hand side of eq.(\ref{colorllary_eq1}).
		
		Corollary~\ref{corollary1} can allow us to estimate the risk bound by using finite samples. In the following, we derive the risk bound for the model based on the empirical data distribution.
	\end{corollary}
	
	\section{Self-Supervised VAEs}
	\label{section_selfVAE}
	
	Self-Supervised technologies have been used in the semi-supervised learning \cite{SelfSupervisedSemi}, multi-task learning \cite{SelfMultiTask} and generative models \cite{SelfSupervisedGAN,SelfSupervisedGAN2,AutoGAM}. In the context of lifelong learning, self-supervised learning is used to overcome catastrophic forgetting \cite{Lifelong_GAN} that this paper focuses on.

	The learning process of the Self-Supervised VAEs is training a model $\mathcal M$ on the samples generated by its decoder $g_\theta(.)$ and the samples from the current task. For instance, at the $i$-th task learning, we get samples from the mixture distribution $\mathbb{P}^{(i-1)} \otimes {\tilde{\mathcal{P}}}_{i}$ where $\mathbb{P}^{(i-1)}$ is the distribution for ${\mathcal  M}^{(i-1)}$, which was trained with the $(i-1)$-th task. We assume that at the $i$-th task learning, we can generate a joint sample $({\bf x}'_{(i-1)},{\bf x}^S_i)$ from $\mathbb{P}^{(i-1)}$ and ${\tilde{\mathcal{P}}}_{i}$, respectively. Then the model can be represented by ${p_\theta }({{\bf{x}}'_{(i - 1)}},{\bf{x}}_i^S,{{\bf{z}}'_{(i - 1)}},{\bf{z}}_i^S) = {p_\theta }({{\bf{x}}'_{(i - 1)}},{\bf{x}}_i^S|{{\bf{z}}'_{(i - 1)}},{\bf{z}}_i^S)p({{\bf{z}}'_{(i - 1)}},{\bf{z}}_i^S)$ where ${{\bf{z}}'_{(i - 1)}}$ and ${\bf{z}}_i^S$ are two latent variables corresponding to different observed variables. We aim to maximize the sample log-likelihood, expressed as~:
	\begin{equation}
	\begin{aligned}
	\log {p_\theta }({{{\bf{x}}}'_{(i - 1)}},{\bf{x}}_i^S) &= \log \int {\int {{p_\theta }({{{\bf{x}}}'_{(i - 1)}},{\bf{x}}_i^S|{{{\bf{z}}}'_{(i - 1)}},{\bf{z}}_i^S)p({{{\bf{z}}}'_{(i - 1)}},{\bf{z}}_i^S)d{{{\bf{z'}}}_{(i - 1)}}d{\bf{z}}_i^S} } \\&
	\ge \int {\int {\log {p_\theta }({{{\bf{x}}}'_{(i - 1)}},{\bf{x}}_i^S|{{{\bf{z}}}'_{(i - 1)}},{\bf{z}}_i^S)p({{{\bf{z}}}'_{(i - 1)}},{\bf{z}}_i^S)d{{{\bf{z}}}'_{(i - 1)}}d{\bf{z}}_i^S} } 
	\label{VAE_eq1}
	\end{aligned}
	\end{equation}
	
	Since ${\bf x}'_{(i-1)}$ and ${\bf x}^S_i$ are independent, we can rewrite Eq.(\ref{VAE_eq1}) as~:
	\begin{equation}
	\begin{aligned}
	\log {p_\theta }({{{\bf{x}}}'_{(i - 1)}}) + \log {p_\theta }({\bf{x}}_i^S) \ge \int {\log {p_\theta }({{{\bf{x}}}'_{(i - 1)}}|{{{\bf{z}}}'_{(i - 1)}})p({{{\bf{z}}}'_{(i - 1)}})d{{{\bf{z}}}'_{(i - 1)}} + } \int {\log {p_\theta }({\bf{x}}_i^S|{\bf{z}}_i^S)p({\bf{z}}_i^S)d{\bf{z}}_i^S} 
	\label{VAE_eq2}
	\end{aligned}
	\end{equation}
	By considering to use the variational distributions \cite{VAE}, we rewrite eq.(\ref{VAE_eq2}) as~:
	\begin{equation}
	\begin{aligned}
	\log {p_\theta }({{{\bf{x}}}'_{(i - 1)}}) + \log {p_\theta }({\bf{x}}_i^S) &\ge {\mathbb{E}_{{q_{{\omega ^i}}}({{{\bf{z}}}'_{(i - 1)}}|{{{\bf{x}}}'_{(i - 1)}})}}[\log {p_\theta }({{{\bf{x}}}'_{(i - 1)}}|{{{\bf{z}}}'_{(i - 1)}})] + KL[{q_{{\omega ^i}}}({{{\bf{z}}}'_{(i - 1)}}|{{{\bf{x}}}'_{(i - 1)}})||p({{{\bf{z}}}'_{(i - 1)}})] \\&+ {\mathbb{E}_{{q_{{\omega ^i}}}({\bf{z}}_i^S|{\bf{x}}_i^S)}}[\log {p_\theta }({\bf{x}}_i^S|{\bf{z}}_i^S)] + KL[{q_{{\omega ^i}}}({\bf{z}}_i^S|{\bf{x}}_i^S)||p({\bf{z}}_i^S)]
	\label{VAE_eq3}
	\end{aligned}
	\end{equation}
	
	\noindent where $p({{{\bf{z}}}'_{(i - 1)}})$ and $p({\bf{z}}_i^S)$ are normal distributions. Eq.\eqref{VAE_eq3} is used for training the VAEs model in a Self-Supervised fashion at the $i$-th task learning. In order to simplify the loss function used for VAEs under LLL, we uniformly generate samples from a mixture distribution ${\mathbb P}^i \otimes {\tilde{\mathcal{P}}}_i$. Then the loss function can be defined as~:
	\begin{equation}
	\begin{aligned}
	\log {p_\theta }({\tilde{\bf{x}}^i}) \ge {{\mathbb E}_{{q_{{\omega ^i}}}({\bf{z}}|{{\tilde{ \bf{x}}}^i})}}[\log {p_\theta }({{\tilde{\bf{x}}}^i}|{\bf{z}})] + KL[{q_{{\omega ^i}}}({\bf{z}}|{{\tilde{ \bf{x}}}^i})||p({\bf{z}})]
	\end{aligned}
	\end{equation}
	where ${\tilde{\bf x}}^i$ is sampled from ${\mathbb P}^i \otimes {\tilde{\mathcal{P}}}_i$. Based on this objective function, we analyze how the Self-Supervised VAEs can achieve an optimal solution in lifelong generative modelling.
	
	\begin{proposition}
		\label{proposition1}
		Let $ {\tilde{\mathcal{P}}}_{(1:t)}=\{{\tilde{\mathcal{P}}}_1 \otimes {\tilde{\mathcal{P}}}_2,\dots,\otimes {\tilde{\mathcal{P}}}_t \}$ be a mixture distribution. The Self-Supervised VAEs approximates ${\tilde{\mathcal{P}}}_{(1:t)}$ exactly when VAEs approximates each target distribution exactly in each task learning.
	\end{proposition}
	
	When satisfying Proposition~\ref{proposition1}, the GB from Theorem 2 in the paper is redefined as~:
	\begin{equation}
	\begin{aligned}
	\frac{1}{t} \sum\limits_{i = 1}^t {{{\mathcal{R}}_{{\mathcal{P}_i}}}} \left( {h,{f_{{\mathcal{P}_i}}}} \right)&\le \frac{1}{t} \sum\limits_{i = 1}^t \{ {{{\mathcal{R}}_{{\tilde{ \mathcal{P}}_i}}}}\left(h,h^*_{\tilde{\mathcal{P}}_i}  \right) \}
	+ {\mathcal R}_A \left({{\mathcal{P}_{(1:t)}}},{{\tilde{\mathcal{P}}_{(1:t)}}} \right).
	\label{proposition1_equ1}
	\end{aligned}
	\end{equation}
	
	From eq.(\ref{proposition1_equ1}), the lifelong learning problem is transformed into the multiple source-target generalization problem. In the following, we assume that the random variables $\{ {\bf x}^T_1,\dots,{\bf x}^T_t \}$ are sampled from $\{ {\mathcal{P}}_1,\dots,{\mathcal{P}}_t \}$ and $\{ {\bf x}^S_1,\dots,{\bf x}^S_t \}$ are sampled from $\{ {\tilde{\mathcal{P}}}_1,\dots,{\tilde{\mathcal{P}}}_t \}$. We can define the KL divergence as follows~:
	\begin{equation}
	\begin{aligned}
	\frac{1}{t}\sum\limits_{i = 1}^t \Big\{ {{{\mathbb E}_{{{\mathcal{P}}_i}}}KL\left( {p\left( {{\bf{z}}|{\bf{x}}_i^T} \right)||p\left( {\bf{z}} \right)} \right)} \Big\} &\le \frac{1}{t}\sum\limits_{i = 1}^t \Big\{ {{{\mathbb E}_{{{\tilde{\mathcal{P}}}_i}}}KL\left( {p\left( {{\bf{z}}|{\bf{x}}_i^S} \right)||p\left( {\bf{z}} \right)} \right)} \Big\} \\&+ \left| {\frac{1}{t}\sum\limits_{i = 1}^t \Big\{ {{{\mathbb E}_{{{\mathcal{P}}_i}}}KL\left( {p\left( {{\bf{z}}|{\bf{x}}_i^T} \right)||p\left( {\bf{z}} \right)} \right)} \Big\} - \frac{1}{t}\sum\limits_{i = 1}^t \Big\{ {{{\mathbb E}_{{{\tilde{\mathcal{P}}}_i}}}KL\left( {p\left( {{\bf{z}}|{\bf{x}}_i^S} \right)||p\left( {\bf{z}} \right)} \right)} \Big\} } \right|
	\label{proposition1_KLEq1}
	\end{aligned}
	\end{equation}
	where the latest term in RHS of Eq.~\eqref{proposition1_KLEq1} is defined as $D_{diff}({\bf x}^T_i,{\bf x}^S_i)$. According to the definition of NLL, we have~:
	
	\begin{equation}
	\begin{aligned}
	\frac{1}{t} \sum\limits_{i = 1}^t {\mathbb E}_{{\mathcal{P}}_i} \{-\log p\left( {\bf x}^T_i \right) \}
	&\le \frac{1}{t} \sum\limits_{i = 1}^t {\mathbb E}_{{\mathcal{P}}_i} \{ -{\mathcal{L}_{ELBO}\left({\bf x}^T_i ; h \right)}  \}.
	\label{proposition1_equ2}
	\end{aligned}
	\end{equation}
	
	In the following, we consider to sum up Eq.~\eqref{proposition1_KLEq1} and Eq.~\eqref{proposition1_equ1}, resulting in~:
	\begin{equation}
	\begin{aligned}
	\frac{1}{t} \sum\limits_{i = 1}^t 
	{\mathbb E}_{{\mathcal{P}}_i} \{ -{\mathcal{L}_{ELBO}\left({\bf x}^T_i ; h \right)}\} &\le \frac{1}{t} \sum\limits_{i = 1}^t \Big\{  {\mathbb E}_{{\tilde{\mathcal{P}}}_i} \{ -{\mathcal{L}_{ELBO}\left( {\bf x}^S_i ; h \right)} + D_{diff}\left({\bf x}^T_i,{\bf x}^S_i\right) \} \Big\}
	+ {\mathcal R}_A \left({{\mathcal{P}_{(1:t)}}},{{\tilde{\mathcal{P}}_{(1:t)}}} \right).
	\label{proposition1_equ3}
	\end{aligned}
	\end{equation}
	
	Eq.~\eqref{proposition1_equ3} is derived based on the formulation of ELBO (See details in Lemma 1 of the paper). It can be observed that LHS of Eq.~\eqref{proposition1_equ3} is RHS of Eq.~\eqref{proposition1_equ2} and we have~:
	\begin{equation}
	\begin{aligned}
	\frac{1}{t} \sum\limits_{i = 1}^t 
	{\mathbb E}_{{\mathcal{P}}_i} \{ -{\log p\left({\bf x}^T_i \right)}\} &\le \frac{1}{t} \sum\limits_{i = 1}^t \Big\{ -{\mathcal{L}_{ELBO}\left({\bf x}^S_i ; h \right)} + D_{diff}\left({\bf x}^T_i,{\bf x}^S_i\right)  \Big\}
	+ {\mathcal R}_A \left({{\mathcal{P}_{(1:t)}}},{{\tilde{\mathcal{P}}_{(1:t)}}} \right).
	\label{proposition1_equ4}
	\end{aligned}
	\end{equation}
	
	Eq.~\eqref{proposition1_equ4} explicitly defines a GB for the VAE mode that is trained on all training sets or the generator distribution that approximates the distribution of all training samples, exactly.
	
	\noindent \textBF{Proof.} We begin with the first task in which $\mathcal M$ is trained on $\tilde{\mathcal{P}}_1$ and the optimal generator distribution $\mathbb{P}^1 = \tilde{\mathcal{P}}_1$. In the second task learning, ${\mathcal M}^{1}$ is trained on the mixture distribution $\mathbb{P}^1 \otimes {\tilde{\mathcal{P}}}_2$. Since $\mathbb{P}^1 = \tilde{\mathcal{P}}_1$, we replace $\mathbb{P}^1 \otimes {\tilde{\mathcal{P}}}_2$ by ${\tilde{\mathcal{P}}}_1 \otimes {\tilde{\mathcal{P}}}_2$ and the optimal generator distribution $\mathbb{P}^2 = {\tilde{\mathcal{P}}}_1 \otimes {\tilde{\mathcal{P}}}_2$. By using the mathematical recursion, we have $\mathbb{P}^t = {\tilde{\mathcal{P}}}_{(1:t)}$ if VAEs approximates the target distribution exactly in each task learning.
	
	\begin{proposition}
		\label{proposition2}
		For a given sequence of tasks $\{ {\mathcal{T}}_1,\dots,{\mathcal{T}}_t\}$, the sample log-likelihood of a single model $\bf{M}$ evaluated on the target distribution $\{ {\mathcal{P}}_1,\dots,{\mathcal{P}}_t \}$ during the $t$-th task learning is bounded by the accumulated errors and the generalization errors.
		
		\begin{equation}
		\begin{aligned}
		\frac{1}{t}\sum\limits_{i = 1}^t {\mathbb E}_{{\mathcal{P}}_i} {\left\{ { - \log p\left( {{\bf{x}}_t^T} \right)} \right\}}
		&\le  {\mathbb E}_{{{\mathbb{P}}^{t - 1}} \otimes {{\tilde {\mathcal{P}}}_t}} \{ { - {\mathcal{L}_{ELBO}}\left( {{\bf{x}}^t} ;h \right)} \} +
		\underbrace{
			\frac{1}{t}\sum\limits_{i = 1}^t \{ {{D_{diff}}\left( {{\bf{x}}_i^S,{\bf{x}}^t} \right)} \}
			+ {\mathcal R}_A \left({{ \tilde{\mathcal{P}}_{(1:t)}},{{{\mathbb{P}}^{t - 1}} \otimes {{\tilde {\mathcal{P}}}_t}}} \right)}_{\text{Accumulated error}} 
		\\&+ \underbrace{ \frac{1}{t}\sum\limits_{i = 1}^t \{ {{D_{diff}}\left( {{\bf{x}}_i^S,{\bf{x}}_i^T} \right)} \}
			+ {\mathcal R}_A \left({{ \tilde{\mathcal{P}}_{(1:t)}},{\mathcal{P}}_{(1:t)} } \right)}_{\text{Generalization error}}
		\label{prop2_eq1}
		\end{aligned}
		\end{equation}
		
		We have the accumulated error ${\rm Err}^c$ and the generalization error term ${\rm Err}^h$~:
		\begin{equation}
		\begin{aligned}
		{\rm Err}^c = \frac{1}{t}\sum\limits_{i = 1}^t \{ {{D_{diff}}\left( {{\bf{x}}_i^S,{\bf{x}}^t} \right)} \} + {\mathcal R}_A \left({{ \tilde{\mathcal{P}}_{(1:t)}},{{{\mathbb{P}}^{t - 1}} \otimes {{\tilde {\mathcal{P}}}_t}}} \right) 
		\end{aligned}
		\end{equation}
		
		\begin{equation}
		\begin{aligned}
		{\rm Err}^h = \frac{1}{t}\sum\limits_{i = 1}^t \{ {{D_{diff}}\left( {{\bf{x}}_i^S,{\bf{x}}_i^T} \right)} \}
		+ {\mathcal R}_A \left({{ \tilde{\mathcal{P}}_{(1:t)}},{\mathcal{P}}_{(1:t)} } \right)
		\end{aligned}
		\end{equation}
		
		Then, Eq.~(\ref{prop2_eq1}) is rewritten as~:
		\begin{equation}
		\begin{aligned}
		\frac{1}{t}\sum\limits_{i = 1}^t {\mathbb E}_{{\mathcal{P}}_i} {\left\{ { - \log p\left( {{\bf{x}}_t^T} \right)} \right\}}
		&\le  {\mathbb E}_{{{\mathbb{P}}^{t - 1}} \otimes {{\tilde {\mathcal{P}}}_t}} \{ { - {\mathcal{L}_{ELBO}}\left( {{\bf{x}}^t} ; h \right)} \} +{\rm Err}^c + {\rm Err}^h
		\label{prop2_eq4}
		\end{aligned}
		\end{equation}
	\end{proposition}
	
	Eq.~(\ref{prop2_eq4}) explicitly defines a GB for the negative sample log-likelihood for $\{ {\mathcal{P}}_1,\dots,{\mathcal{P}}_t \}$ where ${\rm Err}^c$ and ${\rm Err}^h$ are two error terms. We show that the gap on this GB depends not only on ${\rm Err}^c$ but also on ${\rm Err}^h$.
	
	\noindent \textBF{Proof.} Since the model $\mathcal M$ is trained on a sequence of distributions $\{ {\tilde{\mathcal{P}}},\dots,{\tilde{\mathcal{P}}}_t\}$, we would like to derive the GB between the model distribution and $\{ {\tilde{\mathcal{P}}},\dots,{\tilde{\mathcal{P}}}_t\}$ as~:
	\begin{equation}
	\begin{aligned}
	\frac{1}{t} \sum\limits_{i = 1}^t {{{\mathcal{R}}_{{\tilde {\mathcal{P}}_i}}}} \left( {h,{f_{{\tilde{ \mathcal{P}}_i}}}} \right)&\le {{\mathcal{R}}_{{{\mathbb{P}}^{t - 1}} \otimes {{\tilde {\mathcal{P}}}_t}}}\left( {h,h_{{{\mathbb{P}}^{t - 1}} \otimes {{\tilde{\mathcal{P}}}_t}}^*} \right) + {\mathcal R}_A \left({{ \tilde{\mathcal{P}}_{(1:t)}},{{{\mathbb{P}}^{t - 1}} \otimes {{\tilde {\mathcal{P}}}_t}}} \right),
	\label{prop2_proof_equ1}
	\end{aligned}
	\end{equation}
	\noindent where ${\tilde{\mathcal{P}}}_{(1:t)}$ represents the mixture distribution $\{ {\tilde{\mathcal{P}}}_1 \otimes \dots \otimes{\tilde{\mathcal{P}}}_t \}$. As similar to the proof from Lemma 1 of the paper, we add the KL divergence term in both sides in eq.(\ref{prop2_proof_equ1}), resulting in~:
	\begin{equation}
	\begin{aligned}
	\frac{1}{t} \sum\limits_{i = 1}^t \{
	{{{\mathcal{R}}_{{\tilde {\mathcal{P}}_i}}}} \left( {h,{f_{{\tilde{ \mathcal{P}}_i}}}} \right) + 
	KL\left( {q\left( {{\bf{z}}|{\bf{x}}_i^S} \right)||p\left( {\bf{z}} \right)} \right) \}
	&\le {{\mathcal{R}}_{{{\mathbb{P}}^{t - 1}} \otimes {{\tilde {\mathcal{P}}}_t}}}\left( {h,h_{{{\mathbb{P}}^{t - 1}} \otimes {{\tilde{\mathcal{P}}}_t}}^*} \right) + {{\mathbb E}_{{{\mathbb P}^{t-1} \otimes {\tilde{\mathcal{P}}}_t }}}KL(p({\bf{z}}|{\bf{\tilde x}}^t)||p({\bf{z}})) \\&+
	\frac{1}{t}\sum\limits_{i = 1}^t \{  + {D_{diff}}\left( {{\bf{x}}_i^S,{\bf{x}}^t} \right) \}
	+ {\mathcal R}_A \left({{ \tilde{\mathcal{P}}_{(1:t)}},{{{\mathbb{P}}^{t - 1}} \otimes {{\tilde {\mathcal{P}}}_t}}} \right),
	\label{prop2_proof_equ2}
	\end{aligned}
	\end{equation}
	
	We can rewrite Eq.\eqref{prop2_proof_equ2} According to ELBO, resulting in~:
	\begin{equation}
	\begin{aligned}
	\frac{1}{t} \sum\limits_{i = 1}^t {\mathbb E}_{{\tilde{\mathcal{P}}}_i} \{
	- {\mathcal{L}_{ELBO}}\left( {{\bf{x}}_i^S} ; h \right) \}
	&\le  {\mathbb E}_{{{\mathbb{P}}^{t - 1}} \otimes {{\tilde {\mathcal{P}}}_t}} \{ { - {\mathcal{L}_{ELBO}}\left( {{\bf{x}}^t} ; h \right)} \} + \frac{1}{t}\sum\limits_{i = 1}^t \{ {{D_{diff}}\left( {{\bf{x}}_i^S,{\bf{x}}^t} \right)} \}
	+ {\mathcal R}_A \left({{ \tilde{\mathcal{P}}_{(1:t)}},{{{\mathbb{P}}^{t - 1}} \otimes {{\tilde {\mathcal{P}}}_t}}} \right),
	\label{prop2_proof_equ3}
	\end{aligned}
	\end{equation}
	
	Secondly, we consider an ideal situation in which the model $\mathcal M$ approximates the learning target distribution exactly in each task learning. Then, we can derive the GB for the risk between $\{ {\mathcal{P}}_1,\dots,{\mathcal{P}}_t\}$ and $\{ {\tilde{\mathcal{P}}}_1,\dots,{\tilde{\mathcal{P}}}_t\}$~:
	\begin{equation}
	\begin{aligned}
	\frac{1}{t} \sum\limits_{i = 1}^t {\mathbb E}_{{\mathcal{P}}_i} \{
	- {\mathcal{L}_{ELBO}}\left( {{\bf{x}}_i^T} ; h \right) \}
	&\le \frac{1}{t}\sum\limits_{i = 1}^t {\mathbb E}_{{\tilde{\mathcal{P}}}_i} \{ { - {\mathcal{L}_{ELBO}}\left( {{\bf{x}}_i^S};h \right)} \} + \frac{1}{t}\sum\limits_{i = 1}^t \{ {{D_{diff}}\left( {{\bf{x}}_i^S,{\bf{x}}_i^T} \right)} \}
	+ {\mathcal R}_A \left({{ \tilde{\mathcal{P}}_{(1:t)}},{\mathcal{P}}_{(1:t)} } \right),
	\label{prop2_proof_equ4}
	\end{aligned}
	\end{equation}
	
	We also know that the left hand side of Eq.~(\ref{prop2_proof_equ4}) is a bound for $\frac{1}{t}\sum\limits_{i = 1}^t {\mathbb E}_{{\mathcal{P}}_i} {\left\{ { - \log p\left( {{\bf{x}}_t^T} \right)} \right\}} $, expressed as~:
	\begin{equation}
	\begin{aligned}
	\frac{1}{t}\sum\limits_{i = 1}^t {\mathbb E}_{{\mathcal{P}}_i} {\left\{ { - \log p\left( {{\bf{x}}_t^T} \right)} \right\}} \le \frac{1}{t} \sum\limits_{i = 1}^t {\mathbb E}_{\mathcal{P}_i} \{
	- {\mathcal{L}_{ELBO}}\left( {{\bf{x}}_i^T};h \right) \} 
	\label{prop2_proof_equ5}
	\end{aligned}
	\end{equation}
	
	By comparing eq.(\ref{prop2_proof_equ5}) and eq.(\ref{prop2_proof_equ4}), we have~:
	\begin{equation}
	\begin{aligned}
	\frac{1}{t}\sum\limits_{i = 1}^t {\mathbb E}_{{\mathcal{P}}_i} {\left\{ { - \log p\left( {{\bf{x}}_t^T} \right)} \right\}} \le \frac{1}{t}\sum\limits_{i = 1}^t \{ {\mathbb E}_{{\tilde{\mathcal{P}}}_i} \{ { - {\mathcal{L}_{ELBO}}\left( {{\bf{x}}_i^S};h \right)} \} \} + \frac{1}{t}\sum\limits_{i = 1}^t \{ {{D_{diff}}\left( {{\bf{x}}_i^S,{\bf{x}}_i^T} \right)} \}
	+ {\mathcal R}_A \left({{ \tilde{\mathcal{P}}_{(1:t)}},{\mathcal{P}}_{(1:t)} } \right)
	\label{prop2_proof_equ6}
	\end{aligned}
	\end{equation}
	
	Then we add the last two terms in the right hand side of eq.(\ref{prop2_proof_equ4}) to both sides of eq.(\ref{prop2_proof_equ3}), resulting in~:
	\begin{equation}
	\begin{aligned}&
	\frac{1}{t} \sum\limits_{i = 1}^t \{ {\mathbb E}_{{\tilde{\mathcal{P}}}_i} \{
	- {\mathcal{L}_{ELBO}}\left( {{\bf{x}}_i^S};h \right) \} \} + \frac{1}{t}\sum\limits_{i = 1}^t \{ {{D_{diff}}\left( {{\bf{x}}_i^S,{\bf{x}}_i^T} \right)} \}
	+ {\mathcal R}_A \left({{ \tilde{\mathcal{P}}_{(1:t)}},{\mathcal{P}}_{(1:t)} } \right)
	\le 
	{\mathbb E}_{{{\mathbb{P}}^{t - 1}} \otimes {{\tilde {\mathcal{P}}}_t}} \{ { - {\mathcal{L}_{ELBO}}\left( {{\bf{x}}^t};h \right)} \}
	\\&+ \frac{1}{t}\sum\limits_{i = 1}^t \{ {{D_{diff}}\left( {{\bf{x}}_i^S,{\bf{x}}^t} \right)} \}
	+ {\mathcal R}_A \left({{ \tilde{\mathcal{P}}_{(1:t)}},{{{\mathbb{P}}^{t - 1}} \otimes {{\tilde {\mathcal{P}}}_t}}} \right) 
	+ \frac{1}{t}\sum\limits_{i = 1}^t \{ {{D_{diff}}\left( {{\bf{x}}_i^S,{\bf{x}}_i^T} \right)} \}
	+ {\mathcal R}_A \left({{ \tilde{\mathcal{P}}_{(1:t)}},{\mathcal{P}}_{(1:t)} } \right)
	\label{prop2_proof_equ7}
	\end{aligned}
	\end{equation}
	
	It can be observed that the right hand side of eq.(\ref{prop2_proof_equ6}) is equal to the left hand side of eq.(\ref{prop2_proof_equ7}). By comparing eq.(\ref{prop2_proof_equ6}) and eq.(\ref{prop2_proof_equ7}), we have~:
	\begin{equation}
	\begin{aligned}
	\frac{1}{t}\sum\limits_{i = 1}^t {\mathbb E}_{{\mathcal{P}}_i} {\left\{ { - \log p\left( {{\bf{x}}_i^T} \right)} \right\}}
	&\le  {\mathbb E}_{{{\mathbb{P}}^{t - 1}} \otimes {{\tilde {\mathcal{P}}}_t}} \{ { - {\mathcal{L}_{ELBO}}\left( {{\bf{x}}^t};h \right)} \} + \frac{1}{t}\sum\limits_{i = 1}^t \{ {{D_{diff}}\left( {{\bf{x}}_i^S,{\bf{x}}^t} \right)} \}
	\\&+ {\mathcal R}_A \left({{ \tilde{\mathcal{P}}_{(1:t)}},{{{\mathbb{P}}^{t - 1}} \otimes {{\tilde {\mathcal{P}}}_t}}} \right) 
	+ \frac{1}{t}\sum\limits_{i = 1}^t \{ {{D_{diff}}\left( {{\bf{x}}_i^S,{\bf{x}}_i^T} \right)} \}
	+ {\mathcal R}_A \left({{ \tilde{\mathcal{P}}_{(1:t)}},{\mathcal{P}}_{(1:t)} } \right) 
	\label{prop2_proof_equ8}
	\end{aligned}
	\end{equation}
	
	This proves Proposition 4.
	
	\begin{proposition}
		\label{proposition3}
		For a given mixture model $\bf{M}$, The sample log-likelihood of $\bf{M}$ evaluated on the target distribution $\{ {\mathcal{P}}_1,\dots,{\mathcal{P}}_t \}$ during the $t$-th task learning is bounded by~:
		\begin{equation}
		\begin{aligned}
		\frac{1}{t} \sum\limits_{i = 1}^t 
		{\mathbb E}_{{\mathcal{P}}_i} \{ -{\mathcal{L}_{ELBO}\left( {\bf x}^T_i ;h \right)} &\le 
		\frac{1}{t}
		\sum\limits_{i = 1}^{|C'|} \sum\limits_{j = 1}^{{\tilde a}_i} 
		{\mathbb E}_{{\mathbb P}^{c(i,j)}_{a(i,j)}}
		\left
		\{
		- {\mathcal{L}_{ELBO}}\left( {{\bf{x}}_{{a(i,j)}}^t} ; h \right) +
		{\mathcal R}_A\left({{{\tilde{\mathcal{P}}}_{a(i,j)}},{\mathbb{P}}_{a(i,j)}^{c(i,j)}}  \right) 
		\right\} \\&+\frac{1}{t}\sum\limits_{i = 1}^{|C|} 
		{\mathbb E}_{{\tilde{\mathcal{P}}_{a_i}}}
		{\{ 
			- {\mathcal{L}_{ELBO}}\left( {{\bf{x}}_{{a_i}}^S} ; h \right)
			\}  }  +\frac{1}{t}\sum\limits_{i = 1}^{|C'|} {\sum\limits_{j = 1}^{{{\tilde a}_i}} {\{ {D_{diff}}\left( {{\bf{x}}_{a(i,j)}^S,{\bf{x}}_{a(i,j)}^t} \right)\} } } \\&+
		\frac{1}{t}\sum\limits_{i = 1}^t {{D_{diff}}\left( {{\bf{x}}_i^T,{\bf{x}}_i^S} \right)}  + {{\mathcal{R}}_A}\left( {{{\cal P}_{(1:t)}},{{\widetilde {\cal P}}_{(1:t)}}} \right)
		\label{prop3_equ1}
		\end{aligned}
		\end{equation}
		
	\end{proposition}
	
	From eq.(\ref{prop3_equ1}), it can be observed that the generalization error term $\frac{1}{t}\sum\limits_{i = 1}^t {{D_{diff}}\left( {{\bf{x}}_i^T,{\bf{x}}_i^S} \right)}  + {{\mathcal{R}}_A}\left( {{{\cal P}_{(1:t)}},{{\widetilde {\cal P}}_{(1:t)}}} \right)$ is the same for both the single model $\mathcal M$ and the mixture model $\bf{M}$. This means that the mixture model can obtain a tight GB by reducing the accumulated error term during lifelong learning.
	
	\noindent \textBF{Proof.} Firstly, we consider to take $\{{\tilde{\mathcal{P}}_1},\dots,{\tilde{\mathcal{P}}_t} \}$ as the target distribution and we can derive the GB for $\bf{M}$ as~: 
	\begin{equation}
	\begin{aligned}
	\frac{1}{t}\sum\limits_{i = 1}^{|C|} {\{ {{\mathcal{R}}_{{{\tilde{\mathcal P}}_{{a_i}}}}}\left( {h,{f_{{{\tilde{\mathcal P}}_{{a_i}}}}}} \right)\} }  + \frac{1}{t}\sum\limits_{i = 1}^{|C'|} {\sum\limits_{j = 1}^{{{\tilde a}_i}} {\{ {{\mathcal{R}}_{\tilde{\mathcal{P}}_{a(i,j)}}}\left( {h,{f_{\tilde{\mathcal{P}}_{a(i,j)}}}} \right)\} } }  &\le 
	\frac{1}{t}
	\sum\limits_{i = 1}^{|C'|} \sum\limits_{j = 1}^{{\tilde a}_i} \left\{ {{\mathcal{R}}_{{\mathbb{P}}_{a(i,j)}^{c(i,j)}}}\left( {h,{f_{{\mathbb{P}}_{a(i,j)}^{c(i,j)}}}} \right) 
	\right. \\
	& \left. 
	+ 
	{\mathcal R}_A\left({{{\tilde{\mathcal{P}}}_{a(i,j)}},{\mathbb{P}}_{a(i,j)}^{c(i,j)}}  \right)
	\right\} \\&+\frac{1}{t}\sum\limits_{i = 1}^{|C|} {\{ {{\mathcal{R}}_{{{\tilde{\mathcal P}}_{{a_i}}}}}\left( {h,{f_{{{\tilde{\mathcal P}}_{{a_i}}}}}} \right)\} } 
	\label{prop3_proof_eq1}
	\end{aligned}
	\end{equation}
	We then add the KL divergence in both sides of eq.(\ref{prop3_proof_eq1}), resulting in~:
	\begin{equation}
	\begin{aligned}
	&\frac{1}{t}\sum\limits_{i = 1}^{|C|} {\{ {{\mathcal{R}}_{{{\tilde{\mathcal P}}_{{a_i}}}}}\left( {h,{f_{{{\tilde{\mathcal P}}_{{a_i}}}}}} \right) + KL\left( {q\left( {{\bf{z}}|{\bf{x}}_{{a_i}}^S} \right)||p\left( {\bf{z}} \right)} \right) \} }  + \frac{1}{t}\sum\limits_{i = 1}^{|C'|} {\sum\limits_{j = 1}^{{{\tilde a}_i}} {\{ {{\mathcal{R}}_{\tilde{\mathcal{P}}_{a(i,j)}}}\left( {h,{f_{\tilde{\mathcal{P}}_{a(i,j)}}}} \right) +KL\left( {q\left( {{\bf{z}}|{\bf{x}}_{a(i,j)}^S} \right)||p\left( {\bf{z}} \right)} \right) \} } }  \le 
	\\&
	\frac{1}{t}
	\sum\limits_{i = 1}^{|C'|} \sum\limits_{j = 1}^{{\tilde a}_i} \left\{ {{\mathcal{R}}_{{\mathbb{P}}_{a(i,j)}^{c(i,j)}}}\left( {h,{f_{{\mathbb{P}}_{a(i,j)}^{c(i,j)}}}} \right) 
	+ 
	{\mathcal R}_A\left({{{\tilde{\mathcal{P}}}_{a(i,j)}},{\mathbb{P}}_{a(i,j)}^{c(i,j)}}  \right) +KL\left( {q\left( {{\bf{z}}|{\bf{x}}_{a(i,j)}^t} \right)||p\left( {\bf{z}} \right)} \right)
	\right\} \\&+\frac{1}{t}\sum\limits_{i = 1}^{|C|} {\{ {{\mathcal{R}}_{{{\tilde{\mathcal P}}_{{a_i}}}}}\left( {h,{f_{{{\tilde{\mathcal P}}_{{a_i}}}}}} \right)\} +KL\left( {q\left( {{\bf{z}}|{\bf{x}}_{a_i}^S} \right)||p\left( {\bf{z}} \right)} \right) }  +\frac{1}{t}\sum\limits_{i = 1}^{|C'|} {\sum\limits_{j = 1}^{{{\tilde a}_i}} {\{ {D_{diff}}\left( {{\bf{x}}_{a(i,j)}^S,{\bf{x}}_{a(i,j)}^t} \right)\} } } 
	\label{prop3_proof_eq2}
	\end{aligned}
	\end{equation}
	
	We rewrite eq.(\ref{prop3_proof_eq2}) as the ELBO from as~:
	\begin{equation}
	\begin{aligned}
	&\frac{1}{t}\sum\limits_{i = 1}^{|C|} {\mathbb E}_{{\tilde{\mathcal{P}}_{a_i}}} {\{  - {\mathcal{L}_{ELBO}}\left( {{\bf{x}}_{{a_i}}^S} ; h \right) \} }  + \frac{1}{t}\sum\limits_{i = 1}^{|C'|} {\sum\limits_{j = 1}^{{{\tilde a}_i}} {\mathbb E}_{{\tilde{\mathcal{P}}_{a(i,j)}}}{\{ - {\mathcal{L}_{ELBO}}\left( {{\bf{x}}_{{a(i,j)}}^S} ; h \right) \} } }  \le 
	\\&
	\frac{1}{t}
	\sum\limits_{i = 1}^{|C'|} \sum\limits_{j = 1}^{{\tilde a}_i} {\mathbb E}_{{\mathbb P}^{c(i,j)}_{a(i,j)}}
	\left
	\{
	- {\mathcal{L}_{ELBO}}\left( {{\bf{x}}_{{a(i,j)}}^t};h \right) +
	{\mathcal R}_A\left({{{\tilde{\mathcal{P}}}_{a(i,j)}},{\mathbb{P}}_{a(i,j)}^{c(i,j)}}  \right) 
	\right\} \\&+\frac{1}{t}\sum\limits_{i = 1}^{|C|} 
	{\mathbb E}_{{\tilde{\mathcal{P}}_{a_i}}}
	{\{ 
		- {\mathcal{L}_{ELBO}}\left( {{\bf{x}}_{{a_i}}^S};h \right)
		\}  }  +\frac{1}{t}\sum\limits_{i = 1}^{|C'|} {\sum\limits_{j = 1}^{{{\tilde a}_i}} {\{ {D_{diff}}\left( {{\bf{x}}_{a(i,j)}^S,{\bf{x}}_{a(i,j)}^t} \right)\} } } 
	\label{prop3_proof_eq3}
	\end{aligned}
	\end{equation}
	
	In the following, we consider to take $\{ {\mathcal{P}}_1,\dots,{\mathcal{P}}_t \}$ as the target distribution and $\{ {\tilde{\mathcal{P}}}_1,\dots,{\tilde{\mathcal{P}}}_t \}$ as the source distribution. Then we have~:
	\begin{equation}
	\begin{aligned}
	\frac{1}{t} \sum\limits_{i = 1}^t 
	{\mathbb E}_{{\mathcal{P}}_i} \{ -{\mathcal{L}_{ELBO}\left( {\bf x}^T_i ; h \right)} &\le \frac{1}{t} \sum\limits_{i = 1}^t {\mathbb E}_{{\tilde{\mathcal{P}}}_i} \{ -{\mathcal{L}_{ELBO}\left( {\bf x}^S_i ; h \right)} + D_{diff}\left({\bf x}^T_i,{\bf x}^S_i\right) \}
	+ {\mathcal R}_A \left({{\mathcal{P}_{(1:t)}}},{{\tilde{\mathcal{P}}_{(1:t)}}} \right).
	\label{prop3_proof_eq4}
	\end{aligned}
	\end{equation}
	
	We then add the latest two term of RHS of Eq.~\eqref{prop3_proof_eq4} into both sides of Eq.~\eqref{prop3_proof_eq3}, resulting in~:
	\begin{equation}
	\begin{aligned}
	&
	\frac{1}{t}\sum\limits_{i = 1}^t {{D_{diff}}\left( {{\bf{x}}_i^T,{\bf{x}}_i^S} \right)}  + {{\mathcal{R}}_A}\left( {{{\cal P}_{(1:t)}},{{\widetilde {\cal P}}_{(1:t)}}} \right)
	\frac{1}{t}\sum\limits_{i = 1}^{|C|} {\mathbb E}_{{\tilde{\mathcal{P}}_{a_i}}} {\{  - {\mathcal{L}_{ELBO}}\left( {{\bf{x}}_{{a_i}}^S} ; h \right) \} }  + \frac{1}{t}\sum\limits_{i = 1}^{|C'|} {\sum\limits_{j = 1}^{{{\tilde a}_i}} {\mathbb E}_{{\tilde{\mathcal{P}}_{a(i,j)}}}{\{ - {\mathcal{L}_{ELBO}}\left( {{\bf{x}}_{{a(i,j)}}^S} ; h \right) \} } }  \le 
	\\&
	\frac{1}{t}
	\sum\limits_{i = 1}^{|C'|} \sum\limits_{j = 1}^{{\tilde a}_i} {\mathbb E}_{{\mathbb P}^{c(i,j)}_{a(i,j)}}
	\left
	\{
	- {\mathcal{L}_{ELBO}}\left( {{\bf{x}}_{{a(i,j)}}^t} ; h \right) +
	{\mathcal R}_A\left({{{\tilde{\mathcal{P}}}_{a(i,j)}},{\mathbb{P}}_{a(i,j)}^{c(i,j)}}  \right) 
	\right\} \\&+\frac{1}{t}\sum\limits_{i = 1}^{|C|} 
	{\mathbb E}_{{\tilde{\mathcal{P}}_{a_i}}}
	{\{ 
		- {\mathcal{L}_{ELBO}}\left( {{\bf{x}}_{{a_i}}^S};h \right)
		\}  }  +\frac{1}{t}\sum\limits_{i = 1}^{|C'|} {\sum\limits_{j = 1}^{{{\tilde a}_i}} {\{ {D_{diff}}\left( {{\bf{x}}_{a(i,j)}^S,{\bf{x}}_{a(i,j)}^t} \right)\} } } \\&+
	\frac{1}{t}\sum\limits_{i = 1}^t {{D_{diff}}\left( {{\bf{x}}_i^T,{\bf{x}}_i^S} \right)}  + {{\mathcal{R}}_A}\left( {{{\cal P}_{(1:t)}},{{\widetilde {\cal P}}_{(1:t)}}} \right)
	\label{prop3_proof_eq5}
	\end{aligned}
	\end{equation}
	
	It can be noted that LHS of Eq.~\eqref{prop3_proof_eq5} is equal to RHS of Eq.~\eqref{prop3_proof_eq4}, therefore, RHS of Eq.~\eqref{prop3_proof_eq5} is also an upper bound to LHS of Eq.~\eqref{prop3_proof_eq4}.
	
	\begin{equation}
	\begin{aligned}
	\frac{1}{t} \sum\limits_{i = 1}^t 
	{\mathbb E}_{{\mathcal{P}}_i} \{ -{\mathcal{L}_{ELBO}\left( {\bf x}^T_i ; h \right)} &\le 
	\frac{1}{t}
	\sum\limits_{i = 1}^{|C'|} \sum\limits_{j = 1}^{{\tilde a}_i} {\mathbb E}_{{\mathbb P}^{c(i,j)}_{a(i,j)}}
	\left
	\{
	- {\mathcal{L}_{ELBO}}\left( {{\bf{x}}_{{a(i,j)}}^t};h \right) +
	{\mathcal R}_A\left({{{\tilde{\mathcal{P}}}_{a(i,j)}},{\mathbb{P}}_{a(i,j)}^{c(i,j)}}  \right) 
	\right\} \\&+\frac{1}{t}\sum\limits_{i = 1}^{|C|} 
	{\mathbb E}_{{\tilde{\mathcal{P}}_{a_i}}}
	{\{ 
		- {\mathcal{L}_{ELBO}}\left( {{\bf{x}}_{{a_i}}^S};h \right)
		\}  }  +\frac{1}{t}\sum\limits_{i = 1}^{|C'|} {\sum\limits_{j = 1}^{{{\tilde a}_i}} {\{ {D_{diff}}\left( {{\bf{x}}_{a(i,j)}^S,{\bf{x}}_{a(i,j)}^t} \right)\} } } \\&+
	\frac{1}{t}\sum\limits_{i = 1}^t {{D_{diff}}\left( {{\bf{x}}_i^T,{\bf{x}}_i^S} \right)}  + {{\mathcal{R}}_A}\left( {{{\cal P}_{(1:t)}},{{\widetilde {\cal P}}_{(1:t)}}} \right)
	\label{prop3_proof_eq6}
	\end{aligned}
	\end{equation}
	
	Eventually, we know that LHS of Eq.~\eqref{prop3_proof_eq6} is an upper bound to NLL, then RHS of Eq.~\eqref{prop3_proof_eq6} is also an upper bound to NLL.
	
	\begin{equation}
	\begin{aligned}
	\frac{1}{t} \sum\limits_{i = 1}^t 
	{\mathbb E}_{{\mathcal{P}}_i} \{ -{\log p\left({\bf x}^T_i \right)} &\le 
	\frac{1}{t}
	\sum\limits_{i = 1}^{|C'|} \sum\limits_{j = 1}^{{\tilde a}_i} {\mathbb E}_{{\mathbb P}^{c(i,j)}_{a(i,j)}}
	\left
	\{
	- {\mathcal{L}_{ELBO}}\left( {{\bf{x}}_{{a(i,j)}}^t};h \right) +
	{\mathcal R}_A\left({{{\tilde{\mathcal{P}}}_{a(i,j)}},{\mathbb{P}}_{a(i,j)}^{c(i,j)}}  \right) 
	\right\} \\&+\frac{1}{t}\sum\limits_{i = 1}^{|C|} 
	{\mathbb E}_{{\tilde{\mathcal{P}}_{a_i}}}
	{\{ 
		- {\mathcal{L}_{ELBO}}\left( {{\bf{x}}_{{a_i}}^S};h \right)
		\}  }  +\frac{1}{t}\sum\limits_{i = 1}^{|C'|} {\sum\limits_{j = 1}^{{{\tilde a}_i}} {\{ {D_{diff}}\left( {{\bf{x}}_{a(i,j)}^S,{\bf{x}}_{a(i,j)}^t} \right)\} } } \\&+
	\frac{1}{t}\sum\limits_{i = 1}^t {{D_{diff}}\left( {{\bf{x}}_i^T,{\bf{x}}_i^S} \right)}  + {{\mathcal{R}}_A}\left( {{{\cal P}_{(1:t)}},{{\widetilde {\cal P}}_{(1:t)}}} \right)
	\label{prop3_proof_eq7}
	\end{aligned}
	\end{equation}
	
	This proves Proposition~\ref{proposition3}.
	
	\begin{proposition}
		Let $\mathcal{P}_i$ be a target distribution and let ${\mathbb P}^i$ be an approximation distribution that partly overlaps with $\mathcal{P}_i$. Let $\{\theta^{*}, \omega^{*} \}$ be the optimal parameters for a VAE model trained on samples ${\bf x}'$ drawn from ${\mathbb P}^i$. Then ${\mathcal{L}}_{ELBO}(\{\theta^{*}, \omega^{*} \}, {\bf x})$ is not a tight ELBO for $\log p_\theta({\bf x})$ by using the model's parameters $\{\theta^{*}, \omega^{*} \}$.
	\end{proposition}
	
	\noindent \textBF{Proof} Let $\{\theta', \omega' \}$ be an optimal parameter set for a VAE model trained on real samples ${\bf x}$ drawn from ${\mathcal{P}}_i$ and we have ${\mathcal{L}}_{ELBO}(\{\theta', \omega' \} ,{\bf x}) > {\mathcal{L}}_{ELBO}(\{{\tilde \theta}, {\tilde \omega} \}, {\bf x}), {\tilde \theta} \ne \theta', {\tilde \omega} \ne \omega'$. Since the model's parameters $\{ \theta^*, \omega^* \}$ are optimized by maximizing ELBO on samples draw from ${\mathbb P}_i$ and are not equal to $\{ \theta', \omega'\}$. Therefore we have ${\mathcal{L}}_{ELBO}(\{\theta', \omega' \}, {\bf x}) > {\mathcal{L}}_{ELBO}(\{ { \theta^*}, {\omega^*} \}, {\bf x})$ and ${\mathcal{L}}_{ELBO}(\{{ \theta^*}, {\omega^*} \}, {\bf x})$ is not a tight ELBO to the real sample log-likelihood $\log p_\theta({\bf x})$.

	\begin{figure}[h]
		\centering
		\includegraphics[scale=0.5]{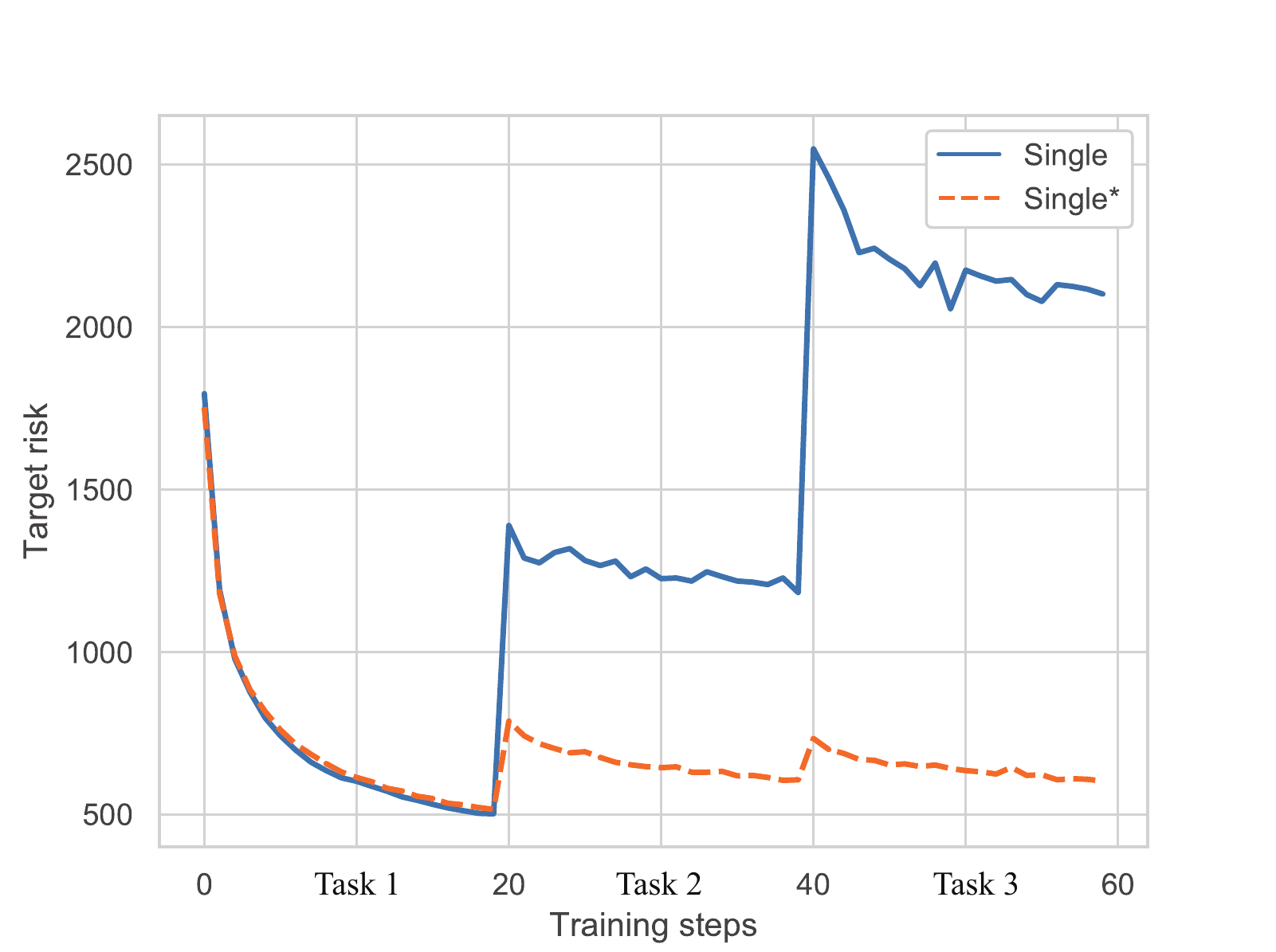}
		\centering
		\caption{The target risk for a single model under MNIST, Fashion, IFashion lifelong learning. }
		\label{proposition6_results}
	\end{figure}
	
	In order to investigate the results for Proposition 6, we train a single model with GR under MNIST, Fashion, IFashion (MIF) lifelong learning (The detailed setting is provided in Appendix~\ref{Appendix_Results_GB}). We also train a single model that can access the real training samples from previous tasks during LLL and this model can be seen to be optimal for the real sample log-likelihood. We evaluate the average target risk (estimated by the negative ELBO on all testing samples) for each training epoch and the results are reported in Fig.~\ref{proposition6_results} where "Single" and "Single*" represent the single model with GR and the single model that can access the real training samples from previous tasks, respectively. It notes that "Single" is trained on the approximation distribution instead of the distributon of real training sets. It shows that, as learning more tasks, a VAE model with GR process gradually loses performance on the real data distribution. This demonstrates that the optimal parameters $\{\theta^*,\omega^* \}$ achieved from the approximation distribution are not optimal for the real sample log-likelihood, which empirically proves Proposition 6.

	\newpage
	\section{The main objective function, algorithm and pipeline for DEGM
	}
	
	In this section, we introduce additional information for the proposed DEGM.
	
	\subsection{Derivation of ${\mathcal{L}_{MELBO}}$}
	
	As illustrated in Fig.2 from the paper, where we show the graph structure implementing DEGM, we have $K$ number of basic nodes in DEGM. When building a new specific node for learning a new task, we derive the main objective function showing as follows.
	
	\noindent \textBF{Theorem 4.} When a new specific node ($(t+1)$-th node) is built for learning the $(t+1)$-th task. This specific node connected with all basic nodes can be seen as a sub-graph model which can be trained by a valid lower bound (ELBO).
	
	\noindent \textBF{Proof.} We start by considering the KL divergence \cite{Tutoria_VAEs}~:
	\begin{equation}
	\begin{aligned}
	{{KL}}\left[ {Q\left( {\bf{z}} \right)|| p\left( {{\bf{z}}|{\bf{x}}} \right)} \right] = {\mathbb{E}_{z \sim Q\left( {\bf{z}} \right)}}\left[ {\log Q\left( {\bf{z}} \right) - \log p\left( {{\bf{z}}|{\bf{x}}} \right)} \right]
	\end{aligned}
	\end{equation}
	\noindent where $Q({\bf z})$ is the variational distribution. We can rewrite the above equation as~:
	\begin{equation}
	\begin{aligned}
	{{KL}}\left[ {Q\left( {\bf{z}} \right)||p\left( {{\bf{z}}|{\bf{x}}} \right)} \right] &=  {\mathbb{E}_{z \sim Q\left( {\bf{z}} \right)}}\left[ {\log Q\left( {\bf{z}} \right) - \log p\left( {{\bf{x}}|{\bf{z}}} \right) - \log p\left( z \right)} \right] + \log p\left( {\bf{x}} \right)
	\end{aligned}
	\end{equation}
	And we have~:
	\begin{equation}
	\begin{aligned}
	\log p\left( {\bf{x}} \right) - {{KL}}\left[ {Q\left( {\bf{z}} \right)||p\left( {{\bf{z}}|{\bf{x}}} \right)} \right] &= {\mathbb{E}_{z \sim Q\left( {\bf{z}} \right)}}\left[ {\log p\left( {{\bf{x}}|{\bf{z}}} \right)} \right] - {{KL}}\left[ {Q\left( {\bf{z}} \right)||p\left( {\bf{z}} \right)} \right]
	\label{ELBO_eq2}
	\end{aligned}
	\end{equation}
	\noindent where the right hand side is also called evidence lower bound (ELBO). We particularly focus on the KL term ${{KL}}\left[ {Q\left( {\bf{z}} \right)||p\left( {\bf{z}} \right)} \right]$ which has the following from~:
	\begin{equation}
	\begin{aligned}
	{{KL}}\left[ {Q\left( {\bf{z}} \right)||p\left( {\bf{z}} \right)} \right] = \int {q\left( {\bf{z}} \right)} \log \frac{{q\left( {\bf{z}} \right)}}{{p\left( {\bf{z}} \right)}}d{\bf{z}}
	\label{KL_equ1}
	\end{aligned}
	\end{equation}
	\noindent where $q({\bf z})$ is the density of $Q({\bf x})$. Since we have $K$ components and we consider the $q({\bf z})$ to be mixture density function $q\left( {\bf{z}} \right) = \sum\nolimits_{i = 1}^w {{\pi _i}{q_{{{\tilde \omega }_{\mathcal{SG}(i)}}}} \circ {q_{{{\omega '}_{(t + 1)}}}}\left( {{\bf{z}}|{\bf{x}}} \right)}$ where $\pi _i$ is the weight. We then rewrite Eq.~(\ref{KL_equ1}) as~:
	\begin{equation}
	\begin{aligned}
	{{KL}}\left[ {Q\left( {\bf{z}} \right)||p\left( {\bf{z}} \right)} \right]
	&=  \int {\left( {\sum\nolimits_{i = 1}^K {{\pi _i}{q_{{{\tilde \omega }_{\mathcal{GI}(i)}}}} \circ {q_{{{\omega '}_{(t + 1)}}}}\left( {{\bf{z}}|{\bf{x}}} \right)} } \right)} \log \frac{{q\left( {\bf{z}} \right)}}{{p\left( {\bf{z}} \right)}}d{\bf{z}}
	\\&= \sum\nolimits_{i = 1}^K {{\pi _i}} \int {{q_{{{\tilde \omega }_{\mathcal{GI}(i)}}}} \circ {q_{{{\omega '}_{(t + 1)}}}}\left( {{\bf{z}}|{\bf{x}}} \right)\log \frac{{q\left( {\bf{z}} \right)}}{{p\left( {\bf{z}} \right)}}d{\bf{z}}} 
	\label{KL_equ2}
	\end{aligned}
	\end{equation}
	We then add the term ${q_{{{\tilde \omega }_{\mathcal{SG}(i)}}}} \circ {q_{{{\omega '}_{(t + 1)}}}}\left( {{\bf{z}}|{\bf{x}}} \right)/{q_{{{\tilde \omega }_{\mathcal{SG}(i)}}}} \circ {q_{{{\omega '}_{(t + 1)}}}}\left( {{\bf{z}}|{\bf{x}}} \right)$ to Eq.~(\ref{KL_equ2}), resulting in~:
	\begin{equation}
	\begin{aligned}
	{{KL}}\left[ {Q\left( {\bf{z}} \right)||p\left( {\bf{z}} \right)} \right] &=  \sum\nolimits_{i = 1}^K {{\pi _i}} \int {q_{{{\tilde \omega }_{\mathcal{GI}(i)}}}} \circ {q_{{{\omega '}_{(t + 1)}}}}\left( {{\bf{z}}|{\bf{x}}} \right)\log \{ \frac{{q\left( {\bf{z}} \right)}}{{p\left( {\bf{z}} \right)}}
	\times 
	\frac{{{q_{{{\tilde \omega }_{\mathcal{GI}(i)}}}} \circ {q_{{{\omega '}_{(t + 1)}}}}\left( {{\bf{z}}|{\bf{x}}} \right)}}{{{q_{{{\tilde \omega }_{\mathcal{GI}(i)}}}} \circ {q_{{{\omega '}_{(t + 1)}}}}\left( {{\bf{z}}|{\bf{x}}} \right)}} \} d{\bf{z}} 
	\end{aligned}
	\end{equation}
	We rewrite the above equation as~:
	\begin{equation}
	\begin{aligned}
	&\sum\nolimits_{i = 1}^K {{\pi _i}} \int {q_{{{\tilde \omega }_{\mathcal{GI}(i)}}}} \circ {q_{{{\omega '}_{(t + 1)}}}}\left( {{\bf{z}}|{\bf{x}}} \right) \times   \log \{ \frac{{{q_{{{\tilde \omega }_{\mathcal{GI}(i)}}}} \circ {q_{{{\omega '}_{(t + 1)}}}}\left( {{\bf{z}}|{\bf{x}}} \right)}}{{p\left( {\bf{z}} \right)}}  \times
	\frac{{q\left( {\bf{z}} \right)}}{{{q_{{{\tilde \omega }_{\mathcal{GI}(i)}}}} \circ {q_{{{\omega '}_{(t + 1)}}}}\left( {{\bf{z}}|{\bf{x}}} \right)}} \} d{\bf{z}} = \\&
	\sum\nolimits_{i = 1}^K {{\pi _i}} \int {q_{{{\tilde \omega }_{\mathcal{GI}(i)}}}} \circ {q_{{{\omega '}_{(t + 1)}}}}\left( {{\bf{z}}|{\bf{x}}} \right) \times  \left( {\log \frac{{{q_{{{\tilde \omega }_{\mathcal{GI}(i)}}}} \circ {q_{{{\omega '}_{(t + 1)}}}}\left( {{\bf{z}}|{\bf{x}}} \right)}} {{p\left( {\bf{z}} \right)}} + 
		\log \frac{{q\left( {\bf{z}} \right)}}{{{q_{{{\tilde \omega }_{\mathcal{GI}(i)}}}} \circ {q_{{{\omega '}_{(t + 1)}}}}\left( {{\bf{z}}|{\bf{x}}} \right)}}} \right)d{\bf{z}} 
	\end{aligned}
	\end{equation}
	Then we can rewrite the above equation as KL terms~:
	\begin{equation}
	\begin{aligned}
	{{KL}}\left[ {Q\left( {\bf{z}} \right)||p\left( {{\bf{z}}|{\bf{x}}} \right)} \right] &=  \sum\nolimits_{i = 1}^K \left( {{\pi _i}{{KL}}\left[ {Q_{{{{\tilde \omega }_{\mathcal{GI}(i)}}} \, \circ \, {\omega '}_{(t + 1)} } \left( {{\bf{z}}|{\bf{x}}} \right)||p\left( {\bf{z}} \right)} \right]} \right)  \\&-  \sum\nolimits_{i = 1}^K {\left( {{\pi _i}{{KL}}\left[ {Q_{{{{\tilde \omega }_{\mathcal{GI}(i)}}} \, \circ \, {\omega '}_{(t + 1)} } \left( {{\bf{z}}|{\bf{x}}} \right)||p\left( {\bf{z}} \right)} \right]} \right)}
	\label{KL_term4}
	\end{aligned}
	\end{equation}
	\noindent where ${q_{{{\tilde \omega }_{\mathcal{GI}(i)}}}} \circ {q_{{{\omega '}_{(t + 1)}}}}\left( {{\bf{z}}|{\bf{x}}} \right)$ is the density form of ${Q_{{{{\tilde \omega }_{\mathcal{GI}(i)}}} \, \circ \, {\omega '}_{(t + 1)} } \left( {{\bf{z}}|{\bf{x}}} \right)}$. We take the expression of $KL\left[ {Q\left( {\bf{z}} \right)||p\left( {{\bf{z}}|{\bf{x}}} \right)} \right]$ to Eq.~(\ref{ELBO_eq2}), resulting in~:
	\begin{equation}
	\begin{aligned}
	\log p\left( {\bf x} \right) - KL\left[ {Q\left( {\bf{z}} \right)||p\left( {{\bf{z}}|{\bf{x}}} \right)} \right] &= {\mathbb{E}_{z \sim Q\left( {\bf{z}} \right)}}\left[ {\log p\left( {{\bf{x}}|{\bf{z}}} \right)} \right] -  \sum\nolimits_{i = 1}^K {\left( {{\pi _i}{{KL}}\left[ {Q_{{{{\tilde \omega }_{\mathcal{GI}(i)}}} \, \circ \, {\omega '}_{(t + 1)} } \left( {{\bf{z}}|{\bf{x}}} \right)} || p({\bf z}) \right]} \right)} \\&+  \sum\nolimits_{i = 1}^K {\left( {{\pi _i}{{KL}}\left[ {
				{Q_{{{{\tilde \omega }_{\mathcal{GI}(i)}}} \, \circ \, {\omega '}_{(t + 1)} } \left( {{\bf{z}}|{\bf{x}}} \right)}
				||Q\left( {\bf{z}} \right)} \right]} \right)} 
	\end{aligned}
	\end{equation}
	where we move the last term of the right hand side to the left hand side, resulting in~:
	\begin{equation}
	\begin{aligned}
	&\log p\left( {\bf{x}} \right) - {{KL}}\left[ {Q\left( {\bf{z}} \right)||p\left( {{\bf{z}}|{\bf{x}}} \right)} \right] - \sum\nolimits_{i = 1}^K {\left( {{\pi _i}{{KL}}\left[ {
				{Q_{{{{\tilde \omega }_{\mathcal{GI}(i)}}} \, \circ \, {\omega '}_{(t + 1)} } \left( {{\bf{z}}|{\bf{x}}} \right)}
				||Q\left( {\bf{z}} \right)} \right]} \right)}  =\\& {\mathbb{E}_{z \sim Q\left( {\bf{z}} \right)}}\left[ {\log p\left( {{\bf{x}}|{\bf{z}}} \right)} \right] - \sum\nolimits_{i = 1}^K {\left( {{\pi _i}{{KL}}\left[ {
				{Q_{{{{\tilde \omega }_{\mathcal{GI}(i)}}} \, \circ \, {\omega '}_{(t + 1)} } \left( {{\bf{z}}|{\bf{x}}} \right)}
				||p\left( {\bf{z}} \right)} \right]} \right)} 
	\label{final_eq}
	\end{aligned}
	\end{equation}
	We know that KL terms are always larger or equal to zero and the right hand side of Eq.~(\ref{final_eq}) is a lower bound to the sample log-likelihood. Finally, the objective function for training DEGM is to maximize this lower bound~:
	\begin{equation}
	\begin{aligned}
	&{\mathbb{E}_{z \sim Q\left( {\bf{z}} \right)}}\left[ {\log p\left( {{\bf{x}}|{\bf{z}}} \right)} \right] -  \sum\nolimits_{i = 1}^K {\left( {{\pi _i}{{KL}}\left[ {
				{Q_{{{{\tilde \omega }_{\mathcal{GI}(i)}}} \, \circ \, {\omega '}_{(t + 1)} } \left( {{\bf{z}}|{\bf{x}}} \right)}
				||p\left( {\bf{z}} \right)} \right]} \right)}
	\label{obj_fun1}
	\end{aligned}
	\end{equation}
	where in the first term, ${\bf z}$ is sampled from the mixture distribution $Q\left( {\bf{z}} \right) = \sum\nolimits_{i = 1}^K {{\pi _i}{Q_{{{{\tilde \omega }_{\mathcal{GI}(i)}}} \, \circ \, {\omega '}_{(t + 1)} } \left( {{\bf{z}}|{\bf{x}}} \right)}}$. In order to reuse the parameters and transferable information from all basic nodes, we consider the following implementations for the decoder.
	When calculating the first term, we build an input layer on the top of the decoder. This input layer is used as the identity function and is connected with each sub-decoder $g_{{\tilde \theta}_{\mathcal{GI}(i)}}(\bf z)$ of each basic node, represented as one layer or module in a single decoder. Then we obtain the intermediate data representation $\tilde{\bf x} = \sum\nolimits_{i=1}^K \pi _i g_{{{\tilde \theta }_{{\mathcal{GI}}(i) }}}({\bf z})$ which is used as the input for the newly created sub-decoder $g_{\theta'_{(t+1)}}(\tilde{\bf x})$. Therefore, we treat the intermediate data representation ${\tilde {\bf x}}$ as the information between two layers in a single decoder and do not consider ${\tilde {\bf x}}$ to be the random variable. In this case we rewrite the decoding distribution $p({\bf x}|{\bf z})$ as~:
	$$ p({\bf x} | {\bf z}) =
	{p_{{\theta'_{(t + 1)}} \circ \{ {\tilde \theta}_{\mathcal{GI}(1)},\dots,{\tilde \theta}_{\mathcal{GI}(K)}\}
	}}\left( {\bf x} \,|\, {\bf z} \right)
	$$. 
	Therefore, we rewrite Eq.~(\ref{obj_fun1}) as~:
	\begin{equation}
	\begin{aligned}
	\mathcal{L}_{MELBO} \left({\bf x};\mathcal{M}_{(t+1)} \right) &= {\mathbb{E}_{Q(\bf{z})}}\left[ {\log {p_{{\theta'_{(t + 1)}} \circ \{ {\tilde \theta}_{\mathcal{GI}(1)},\dots,{\tilde \theta}_{\mathcal{GI}(K)}\}
		}}\left( {{\bf{x}}\mid {\bf z} } \right)} \right]  \\&- \sum\limits_{i = 1}^K
	\left(
	\pi_i {{{KL}}\left[ {{q_{{\tilde \omega }_{{\cal G}{\cal I}(i)}} \circ q_{{\omega '}_{(t + 1)}}}\left( {{{\bf{z}}} \mid {\bf{x}}} \right)\mid\mid p\left( {\bf{z}}_i \right)} \right]} \right)
	\label{MixtureElbo}
	\end{aligned}
	\end{equation}
	
	Eq.(\ref{MixtureElbo}) shows that we can implement existing variational inference mechanisms such as using a more expressive posterior \cite{Aux_DGM,VAE_AutoFlow,ImportanceHVAE}, important sampling \cite{IWVAE} and Semi-Implicit Variational Inference \cite{Semi_Implicit_VAE,DoublySemiVAE} in our framework to enable for lifelong learning.
	
	\begin{algorithm}[http]
		\small
		\caption{The training algorithm for DEGM}
		\LinesNumbered 
		\KwIn{All training databases}
		\KwOut{The model's parameters}
		\For{$i < taskCount$}{
			\If{$i == 1$}{
				Build a basic component $B_1$ and added to ${\mathcal{G}}$
				\;
				$isBasic = True$\;
			}
			
			\For{$index < batchCount$}{
				${\bf x}_{batch} \sim Q^S_i$ a batch of images sampled from the $i$-th task\;
				\If{isBasic == True}{
					Update $\{{\omega _i},{\theta _i} \}$ by maximizing ${\mathcal{L}}_{ELBO}({\bf x}_{batch};{\mathcal{M}_i} )$\;
				}\Else{
					Update $\{{\omega' _i},{\theta' _i} \}$ by maximizing ${\mathcal{L}}_{MELBO}({\bf x}_{batch};{\mathcal{M}_i})$\;
				}
				
			}
			\textBF{Expansion mechanism}\;
			${\bf x}_{new} \sim Q_{i+1}^S$ collect 1000 number of samples from the next task\;
			\textBF{Calculate the importance of each basic node to the new task}\;
			\For{$k < K$}{
				$ks_k = \left|\mathcal{L}_{ELBO}(B_k) - {\mathbb E}_{{\bf x} \sim {\bf x}_{new} } \mathcal{L}_{ELBO}({\bf x};B_k )\right|$\;
			}
			$\mathcal{K} = \{ks_1,\dots,ks_K \}$\;
			\If{$\min\{{\mathcal{K}}\} \le \tau$}{
				\textBF{The construction of the basic node}\;
				$V(i+1,\mathcal{GI}(j) ) = 0$, $j=1,\dots,K$\;
				Build a basic component $B_{(K+1)}$ and added to ${\mathcal{G}}$
				\;
				$isBasic = True$\;
				$K=K+1$\;
			}\Else{
				\textBF{The construction of the sub-graph structure}\;
				$V(i+1,\mathcal{GI}(i) ) = (w^*-k s_i)/\sum\nolimits_{j = 1}^K (w^*-k s_j) $, $w^*=\sum\nolimits_{j = 1}^K ks_j$, $i=1,\dots,K$\;
				$\sum\nolimits_{j=1}^{K }  \pi_i
				f_{{{\tilde \omega }_{{\mathcal{GI}(j)}}}} \circ {f_{{{\omega'}_{(i + 1)}}}}({\bf x})$ build the inference model \;
				${\bf z} = \sum\nolimits_{j = 1}^K \pi_j {\bf z}_j$ obtain the latent representation\;
				$\tilde{\bf x} = \sum\nolimits_{j=1}^K \pi_j g_{{\tilde \theta }_{{\mathcal{GI}(j)}}}({\bf z}) $ obtain the intermediate representations from sub-decoder of all basic nodes\; $g_{\theta'_{(i+1)}}(\tilde{\bf x})$ build a sub-decoder \;
				Add ${\mathcal{M}}_{i+1}$ in ${\mathcal{S}}$\;
				$isBasic = False$\;
			}
			
		}
		\label{algorithm1}
	\end{algorithm}
	
	\subsection{The pipeline and algorithm of DEGM}
	
\begin{figure}[htbp]
		\vspace{-5pt}
		\centering
		\includegraphics[scale=0.6]{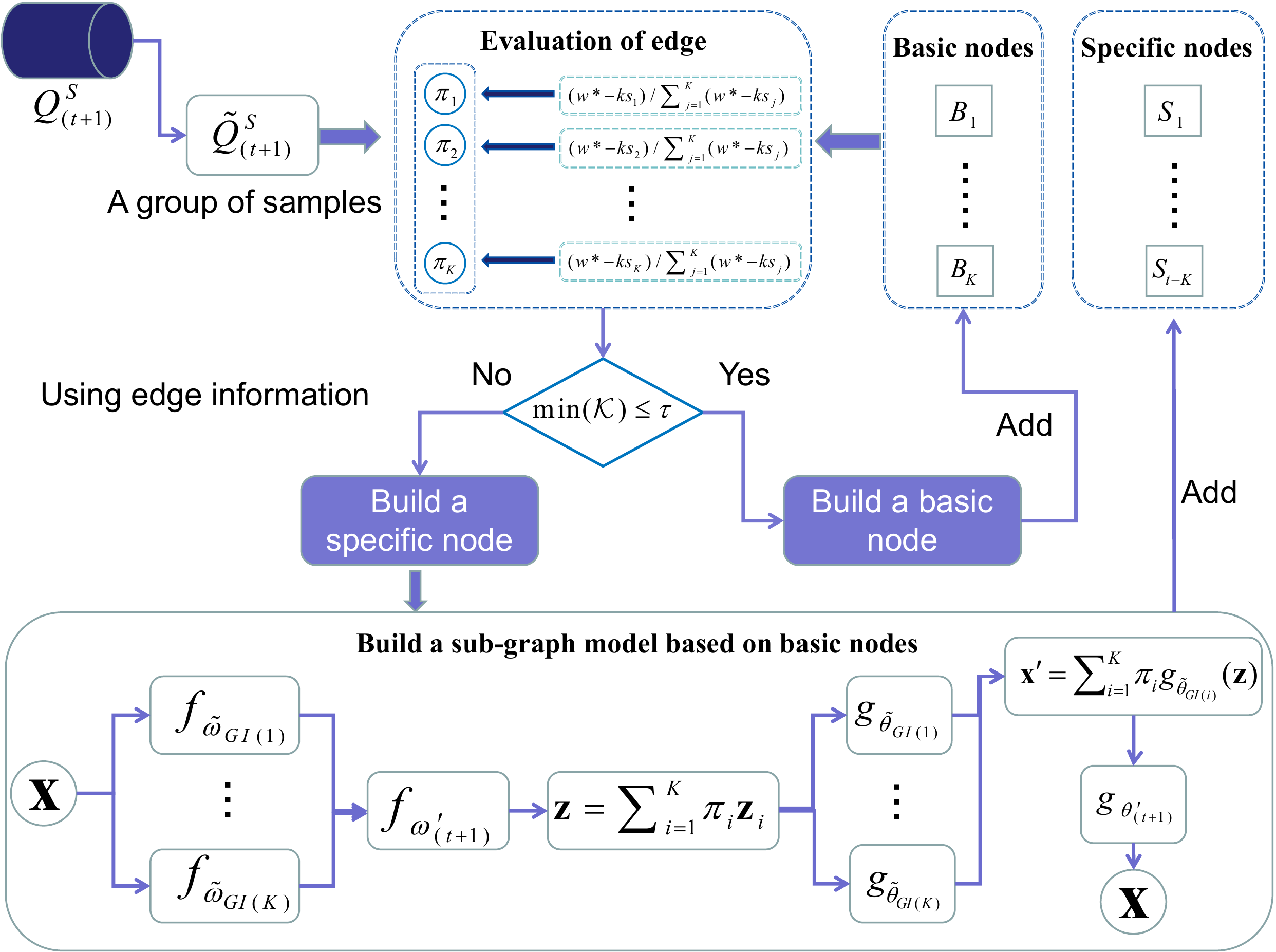}
		\vspace{-2pt}
		\caption{The processing pipeline of the proposed DEGM. }
		\centering
		\label{pipeline_DMix}
	\end{figure}

	In order to conveniently understand the learning procedure of DEGM, we show the processing pipeline in Figure~\ref{pipeline_DMix}. Once the $t$-th task was finished, we get a group of samples, denoted as ${\tilde Q}_{(t+1)}^S$ from $\mathcal{T}_{t+1}$. Then we calculate the edge values, represented by $\{ \pi_1,\dots, \pi_K \}$. If $\min({\mathcal{K}}) > \tau $, then we build a basic node which will be added into $\mathcal{G}$ for learning the $(t+1)$-th task, otherwise, we build a specific node based on members of $\mathcal{G}$ for learning the $(t+1)$-th task. From this pipeline, we observe that the edge values regularize the latent representations and the intermediate data representations in the inference and generation process, respectively. We also provide the pseudocode in Algorithm 1.
	
	\subsection{The component selection for model evaluation during the testing phase}
	Let us consider DEGM model with $t$ trained nodes after LLL. We introduce the cluster assignment ${\bf{u}}$ in the DEGM and the probability density of DEGM on $n$ samples is represented by~:
	\begin{equation}
	\begin{aligned}
	p\left( {\bf x} \right) = \prod\nolimits_{i = 1}^n \sum\nolimits_{j = 1}^t p\left( {\bf x}_i \mid 
	{\bf u}_{(i,j)} \right) p\left( {\bf u}_{(i,j)} \right),\; {\bf u}_{(i,j)} \in \{0,1\}\,.
	\end{aligned}
	\end{equation}
	We particularly focus on the posterior $p({{\bf{u}}_{(i,j)}}|{{\bf{x}}_i})$ which can be rewritten by the Bayes' theorem~:
	\begin{equation}
	\begin{aligned}
	p\left( {\bf u}_{(i,j)} \,|\, {\bf x}_i \right) &=  \frac{p\left( {\bf x}_i\,|\,{\bf u}_{(i,j)} \right)p\left( {\bf u}_{(i,j)} \right)}{p\left( {\bf x}_i \right)}
	= \frac{p\left( {\bf x}_i \,|\, {\bf u}_{(i,j)} \right)p\left( {{{\bf{u}}_{(i,j)}}} \right)}{{\sum\nolimits_{k=1}^t p\left( {\bf u}_{(i,k)} \right)p\left( {\bf x}_i \,|\, {\bf u}_{(i,k)} \right) }},
	\label{Selection}
	\end{aligned}
	\end{equation}
	where the prior is $p({\bf u}_{(i,j)})=1/t$. Since $\log_a$ is a function monotonic increasing if $a>1$, we can replace each 
	$p({\bf x}_i \,|\, {\bf u}_{(i,j)})$ by $\log p({\bf x}_i \,|\, {\bf u}_{(i,j)})$ estimated by 
	$\mathcal{L}_{ELBO}(,{\bf x}_i ;{\cal M}_j )$ for the elements of $\mathcal{G}$ on the sample ${\bf x}$, and by $\mathcal{L}_{MELBO}({\bf x}_i ;\mathcal{M}_j )$ from (\ref{MixtureElbo}) for elements of $\mathcal{S}$, as described in Section 5.3 of the paper. This selection process can allow DEGM to infer a related component without having task labels.
	
	\subsection{Limitations of the proposed theoretical framework}
	
	The primary limitation of the proposed theoretical framework is that it only supports the Gaussian decoder since ELBO can be decomposed by a negative reconstruction error term ($\mathcal{L}$ loss) and the KL divergence term. Our future work will focus on extending the proposed theoretical framework for other types of distributions for the decoder such as Bernoulli distribution. 
	
	One potential negative societal impacts of our work is that the proposed DEGM would cause data leakage. For example, in many lifelong learning tasks, one potential requirement is to allow the data for each task to appear once for privacy protection. 
	
	\subsection{The difference between related work and DEGM}
	
	In this section, we discuss the difference between several related works and DEGM. Firstly, our work is related to \cite{LifelongUnsupervisedVAE} that expands its network architectures on the inference models only. \cite{LifelongUnsupervisedVAE} also uses the generative replay mechanism, which would lead to the degenerated performance on previously learnt samples due to the network forgetting. This is not the case in DEGM since we do not use the generative replay mechanism. To compare with CN-DPM \cite{NeuralDirchlet} that also uses the expansion mechanism only, DEGM can reuse prior knowledge when learning novel samples by forming a graph structure in which the independent basic nodes are used to model entirely different tasks while the specific nodes are built based on the information flow from basic nodes for learning related tasks. However, in both studies \cite{NeuralDirchlet} and \cite{LifelongUnsupervisedVAE}  perform  task-free continual learning, which is not the main task in our work since we address a more general lifelong learning problem where the task boundaries are provided only at the training stage.
	
	Our work is more related to LIMix \cite{LifelongInfinite} that introduces a mixture model with an infinite number of components. LIMix has only a single shared encoder and decoder, which has limitations when learning several entirely different tasks since the shared parameters preserve the knowledge from the first task only. DEGM can have several basic nodes which aim to learn several entirely different tasks and can be used for transferring knowledge when learning novel tasks. Additionally, LIMix introduces a new theoretical analysis for the forgetting behaviour of the model under lifelong learning, which, however, is only applied in the supervised learning task. The proposed theoretical framework in our work studies the forgetting behaviour of the model under the unsupervised lifelong learning setting. 
	
	\section{Unsupervised lifelong learning benchmark}
	
	\subsection{Network architecture and hyperaprameter setting}
	
	For the general network architecture for VAEs, the inference and generator models are implemented by two fully connected networks where each network contains two layers and each layer has 200 hidden units. We also extend VAEs by using two stochastic layers where the latent dimension for the first and second stochastic layer is 100 and 50, respectively. The VAE with two stochastic layers is named as ELBO-GR*. If VAE uses only a single stochastic layer, then the latent dimension is 100. When the important sampling is used in various models such as VAEs, we call it as IWELBO-GR-$K'$ where $K'$ represents the number of weighted samples used in the objective function. The GPU used for the experiment is GeForce GTX 1080. The operating system is Ubuntu 18.04.5.
	
	For the DEGM model, the inference model and the generator in DEGM are implemented by four sub-models and each sub-model has only one layer with 200 units. Therefore, the inference and generator model in each specific node are two sub-models built based on all  basic components members. As similar to a single model, DEGM can be extended with the important sampling framework. For implementation, if a basic component is used for training, we can easily use the IWELBO objective function (See Eq.~\eqref{IWVAE_sample_eq1} in Appendix~\ref{importantSamplingSection}). If a specific node is used for training, we rewrite the objective function as the IWELBO bound form as~:
	\begin{equation}
	\begin{aligned}
	{{\mathcal{L}}_{{MELBO}_{ K'}}}\left({\bf{x}} ; {\mathcal{M}} \right) = {{\mathbb E}_{{{\bf{z}}_1},..,{{\bf{z}}_{K'}} \sim Q\left( {{\bf{z}}} \right)}}\left[ {\log \frac{1}{K'}\sum\limits_{i = 1}^{K'} {\frac{{p\left( {{\bf{x}},{{\bf{z}}_i}} \right)}}{{Q\left( {{{\bf{z}}_i}} \right)}}} } \right]
	\label{iwvae_DMix_eq}
	\end{aligned}
	\end{equation}
	
	The above equation is extended by the IWELBO bound with Eq.~\eqref{ELBO_eq2}. 
	
	CN-DPM is originally designed for the task-free task. In order allow CN-DPM in a more general lifelong unsupervised learning task, we implement a variant of CN-DPM, namely CN-DPM*. We replace the Dirichlet process in CN-DPM* by dynamically creating a new component after the task switch since the Dirichlet process can allow CN-DPM to be a large model. As similar to CN-DPM, CN-DPM* has a single shared module which is frozen after the first task learning and a new component will be built based on this shared module. In order to set the optimal setting for LIMix \cite{LifelongInfinite}, the number of components in LMix matches the number of tasks while each component models a certain task only.
	
	We train various models using Adam \cite{Adam} optimization algorithm with the learning rate of 0.0001. It notes that the decoder models Bernoulli distribution and we use the binary cross entropy as the reconstruction error term in ELBO. The batch size and the number of training epochs for each task learning are 64 and 500, respectively. 
	
	\subsection{Results on cross-domain setting (COFMI)}
	
	We search the threshold $\tau $ of 35-40 for Split MNIST/Fashion and 80-100 for COFMI, respectively. DEGM based methods has four basic nodes and one specific node after Split MNIST and Split Fashion lifelong learning setting. 
	
\begin{figure}[h]
		\centering
		\includegraphics[scale=0.45]{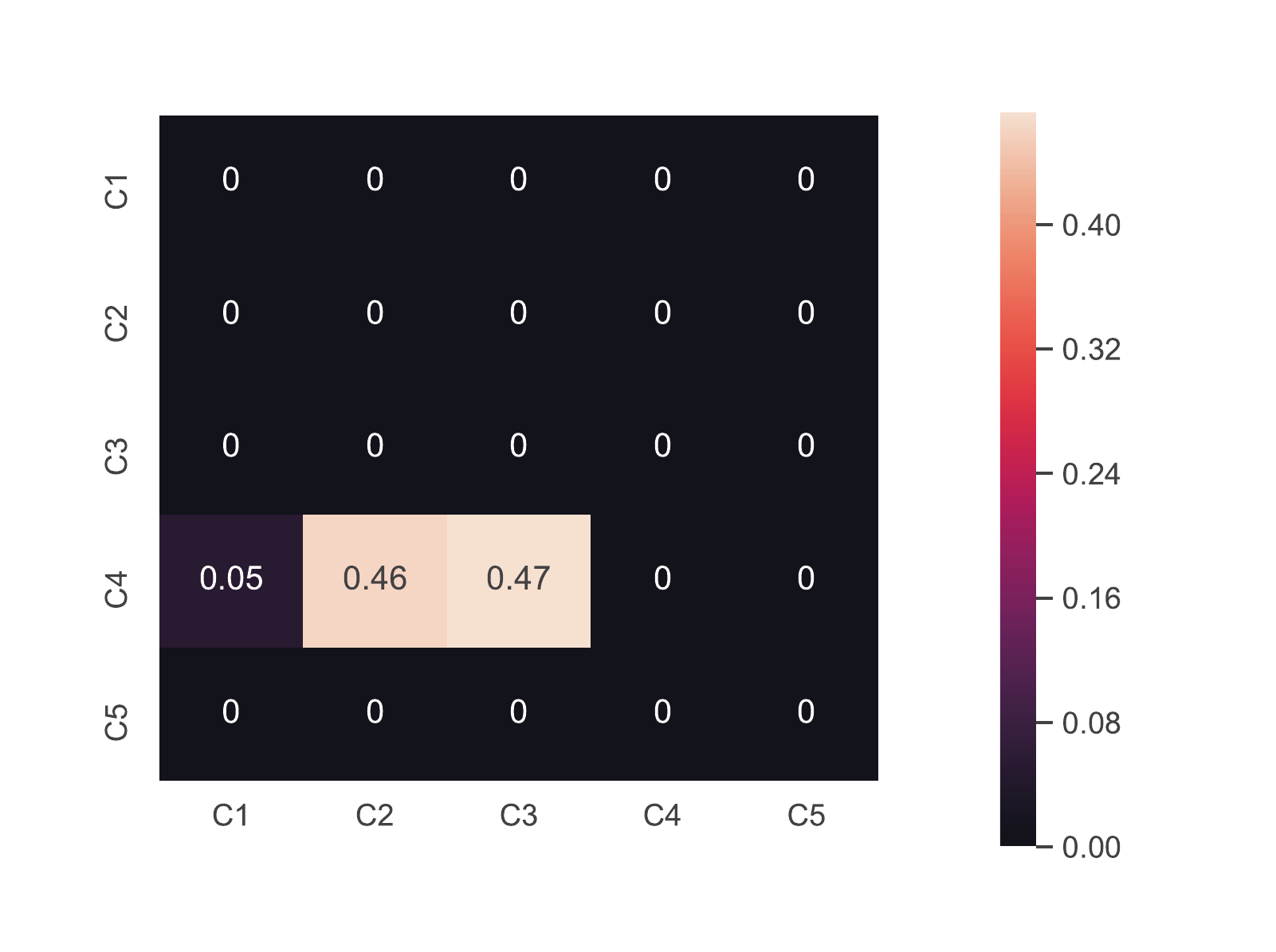}
		\centering
		\caption{$\bf V$ of DEGM-IWVAE-50 under COFMI lifelong learning. "C1" represents the first component and "C4" is a basic node that connects previous three basic nodes ("C1", "C2", "C3"). }
		\label{COFMI_weights}
	\end{figure}

	We report the results under COFMI in Table~\ref{COFMI_tab}. The weight matrix ${\mathcal{V}}$ of DEGM-IWELBO-50 under COFMI lifelong learning is shown in Fig.~\ref{COFMI_weights}. DEGM-IWELBO-50, DEGM-IWELBO-5, DEGM-ELBO has 4 basic and 1 specific nodes after COFMI lifelong learning.
	
	\begin{table}[htb]
		\centering
		\begin{tabular}{lcccccc}
			\toprule
			Mwthods
			&Caltech 101  &OMNIGLOT &Fashion &MNIST&IFashion &Average \\
			\midrule
			ELBO-GR&-163.68&-136.97&-247,91&-101.75&-237,03&-177,47  \\
			IWELBO-GR-50& -153.65&-131.37&-243.62&-97,29&-234.58&-172.10 \\
			IWELBO-GR-5& -166.05&-134.07&-245.78&-99.43&-235.73&-176.21 \\
			ELBO-GR*&-175.10&-140.05&-247.54&-102.73&-237.06&-180.50  \\
			IWELBO-GR*-50&-215.16&-144.42&-246.35&-102.82&-236.12&-188.97 \\
			CN-DPM*-IWELBO-50&-136.22&-150.31&-259.12&-131.34&-243.97&-184.19\\
			LIMix-IWELBO-50&-137.32&-150.79&258.69&-131.25&-243.50&-184.32\\
			\midrule 
			\midrule 
			DEGM-ELBO &-137.72&-116.07&-233.32&-122.33&-234.62&-168.81 \\
			DEGM-IWELBO-50 & -133.93&-112.15&-230.63&-107.44&-232.17& \textBF{ -163.27} \\
			DEGM-IWELBO-5 &-137.18&-113.25&-231.88&-109.21&-233.43&-164.99 \\
			\bottomrule
		\end{tabular}
		\caption{The estimation of the sample log-likelihood under COFMI lifelong learning. }
		\label{COFMI_tab}
	\end{table}
	
	\subsection{Results on Split MNIST and Split Fashion}
	
	In this section, we perform five independent runs for Split MNIST and Split Fashion. After the training, we calculate the average result as well as the standard deviation (std). The results for Split MNIST and Split Fashion are reported in Table~\ref{SplitMnist} and Table~\ref{SplitFashion}, respectively.
	
	\begin{table}[http]
		\centering
		\begin{tabular}{lcccccc}
			\toprule
			Mwthods
			&Run 1  &Run 2 &Run 3 &Run 4&Run 5 &Average \\
			\midrule
			ELBO-GR& -97.97&-98.16&-98.70&-97.91&-98.39&-98.23 $ \pm $ (0.28 std)  \\
			IWELBO-GR-50&-93.61&-93.69&-93.66&-93.64&-93.27&-93.57 $ \pm $ (0.15 std) \\
			IWELBO-GR-5&-95.96&-95.68&-95.95&-95.68&-95.73&-95.80 $ \pm $ (0.12 std)  \\
			ELBO-GR*&-97.92&-98.79&-97.55&-98.45&-99.11&-98.36 $ \pm $ (0.56 std)  \\
			IWELBO-GR*-50& -91.15&-91.33&-91.24&-91.23&-91.18&-91.23  $ \pm $ (0.06 std) \\
			CN-DPM*-IWELBO-50&-95.94&-96.35&-95.90&-95.74&-95.63&-95.91 $ \pm $ (0.24 std) \\
			LIMix-IWELBO-50&-95.57&-95.80&-95.84&-95.84&-95.63&-95.74 $ \pm $ (0.11 std) \\
			\midrule 
			\midrule 
			DEGM-ELBO &-93.85&-87.03&-93.15&-93.71&-93.51&-92.25 $ \pm $ (2.62 std) \\
			DEGM-IWELBO-50 &-89.41&-89.58&-89.60&-85.78&-85.83& \textBF{ -88.04} $ \pm $ (1.82 std) \\
			DEGM-IWELBO-5 &-91.54&-91.70&-91.42&-91.21&-91.32&-91.44  $ \pm $ (0.17 std) \\
			\bottomrule
		\end{tabular}
		\caption{Five independent runs for the estimation of the sample log-likelihood under Split MNIST setting.}
		\label{SplitMnist}
	\end{table}
	
	\begin{table}[http]
		\centering
		\begin{tabular}{lcccccc}
			\toprule
			Mwthods
			&Run 1  &Run 2 &Run 3 &Run 4&Run 5 &Average \\
			\midrule
			ELBO-GR& -241.00&-240.38&-240.26&-240.82&-240.44&-240.58  $ \pm $ (0.28 std)   \\
			IWELBO-GR-50& -236.48&-236.65&-236.98&-236.70&-236.49&-236.66 $ \pm $ (0.18 std) \\
			IWELBO-GR-5& -237.86&-238.19&-238.31&-238.00&-238.00&-238.08 $ \pm $ (0.15 std)   \\
			ELBO-GR*& -242.54&-242.27&-244.76&-242.38&-247.61&-243.91  $ \pm $ (2.06 std)  \\
			IWELBO-GR*-50& -236.71&-236.77&-237.04&-236.88&-237.08&-236.90 $ \pm $ (0.14 std)  \\
			CN-DPM*-IWELBO-50 &-237.60&-237.13&-237.81&-237.38&-237.43&-237.47$ \pm $ (0.22 std) \\
			LIMix&-237.28&-237.50&-237.39&-237.37&-237.87&-237.48$ \pm $ (0.20 std) \\
			\midrule 
			\midrule 
			DEGM-ELBO & -240.59&-237.68&-239.45&-237.60&-237.39&-238.54 $ \pm $ (1.26 std)  \\
			DEGM-IWELBO-50 &-234.17&-232.72&-234.13&-233.15&-234.63& \textBF{-233.76} $ \pm $ (0.70 std)  \\
			DEGM-IWELBO-5 &-236.51&-234.46&-236.59&235.00&-237.08&-235.93 $ \pm $ (1.00 std) \\
			\bottomrule
		\end{tabular}
		\caption{Five independent runs for the estimation of the sample log-likelihood under Split Fashion setting.}
		\label{SplitFashion}
	\end{table}
	
	\clearpage
	
	\section{Additional experiment results}

	\subsection{Quantitative evaluation}
	
	\noindent \textBF{Baselines:} We introduce several baselines and one baseline is the proposed mixture model but we use few epochs (five epochs) for training when the new node is built on other basic nodes, namely DEGM-1. The second baseline is dynamically creating a new VAE to adapt to a new task, namely DEGM-2, which is a strong baseline and would achieve the best performance for each task.
	
	\noindent- \textBF{Hyperparameter setting and GPUs}
	The SGD optimization algorithm for each model is using the Adam \cite{Adam}, with a learning rate of 0.0002 while the other hyperparameters are set to their default. The GPU used for the experiment is GeForce GTX 1080.

	\noindent \textBF{Setting.} To measure the reconstruction quality, we calculate Squared Loss (SL) as the criterion. We train various models under MNIST \cite{MNIST}, SVHN \cite{SVHN}, Fashion \cite{FashionMNIST}, InverseFashion (IFashion), Rated MNIST (RMNIST) and Cifar10 \cite{CIFAR10}, (MSFIRC) lifelong learning setting. We also evaluate the performance of various models under CelebA \cite{Celeba}, CACD \cite{CACD}, 3D-Chair \cite{3DChairs}, Ommiglot \cite{Omniglot}, ImageNet* \cite{ImageNet}, Car \cite{CompCars}, Zappos \cite{shore_Dataset}, CUB \cite{CUB_Birds} (CCCOSCZC) lifelong learning setting.
	
	In the following, we provide details about the setting in the experiments. For lifelong learning setting of MSFIRC and CCCOSCZC, we train model for each task for 20 training epochs. The SGD optimization algorithm for each model is using the Adam \cite{Adam} where the learning rate is 0.0002 and other hyperparameters are default. In the following, we provide the details for datasets. For CelebA, CACD, we randomly choose 10000 samples as the testing set and the remaining samples are used as the training set. For 3D-chair, we randomly choose 1000 samples as the testing set and the remaining samples as the training set. For ImageNet, we randomly choose 10000 and 50000 samples as the testing set and the training set, respectively. For CUB, we randomly choose 1000 samples as the testing set and the remaining samples as the training set.
	
	We consider the threshold of 400 and 600 for DEGM on the MSFIRC and CCCOSCZC, respectively. We report the results in Table~\ref{unsupervised1} and Table~\ref{unsupervised2}, respectively. The interference can be observed by BE and CN-DPM* which gives the degenerated performance on the last task since Cifar10 is different from the first task. In contrast, DEGM avoids interference by building a basic component to learn Cifar10 database. 
	
	\begin{figure}[h]
		\centering
		\subfigure[${\bf V}$ of DEGM after the MSFIRC lifelong learning.]{
			\centering
			\includegraphics[scale=0.45]{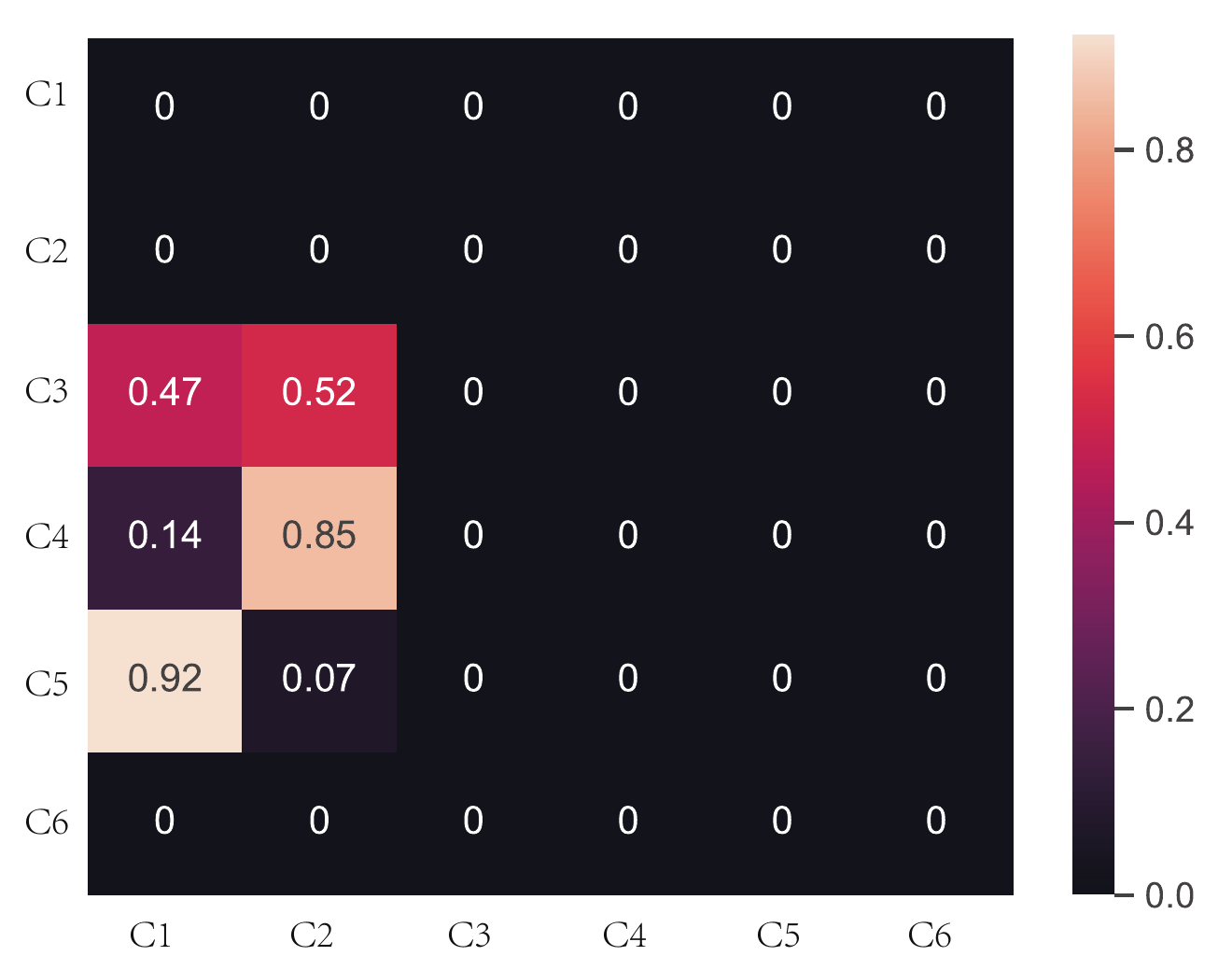}
		}
		\subfigure[$\bf{V}$ of DEGM after the CCCOSCZC lifelong learning.]{
			\centering
			\includegraphics[scale=0.45]{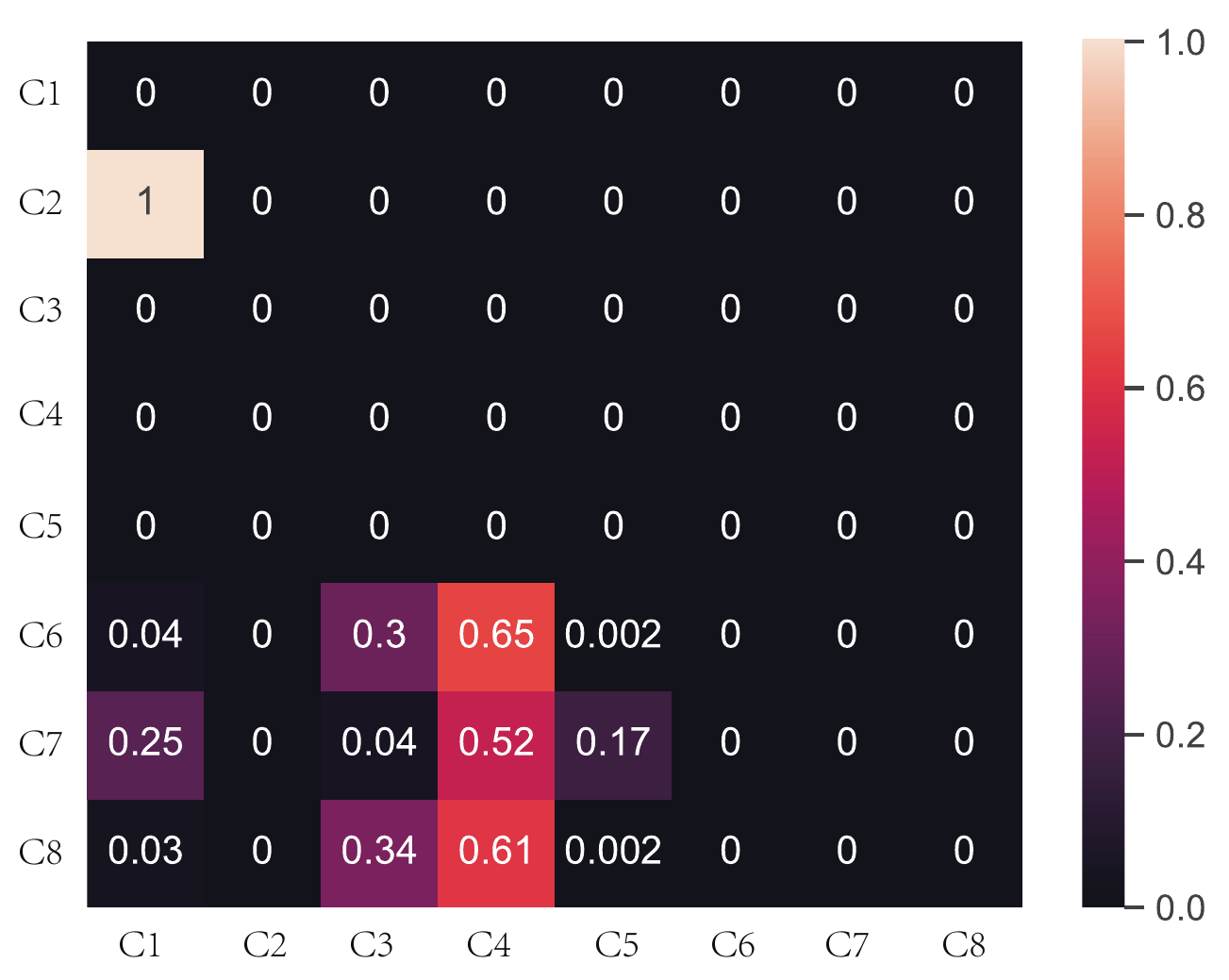}
		}
		\centering
		\caption{The graph adjacency matrix ${\bf V}$, characterizing the connections between different nodes   of DEGM after lifelong learning.}
		\label{selectedResults}
	\end{figure}
	
	\begin{figure}[h]
		\centering
		\subfigure[The edges of DEGM after the MSFIRC lifelong learning.]{
			\centering
			\includegraphics[scale=0.45]{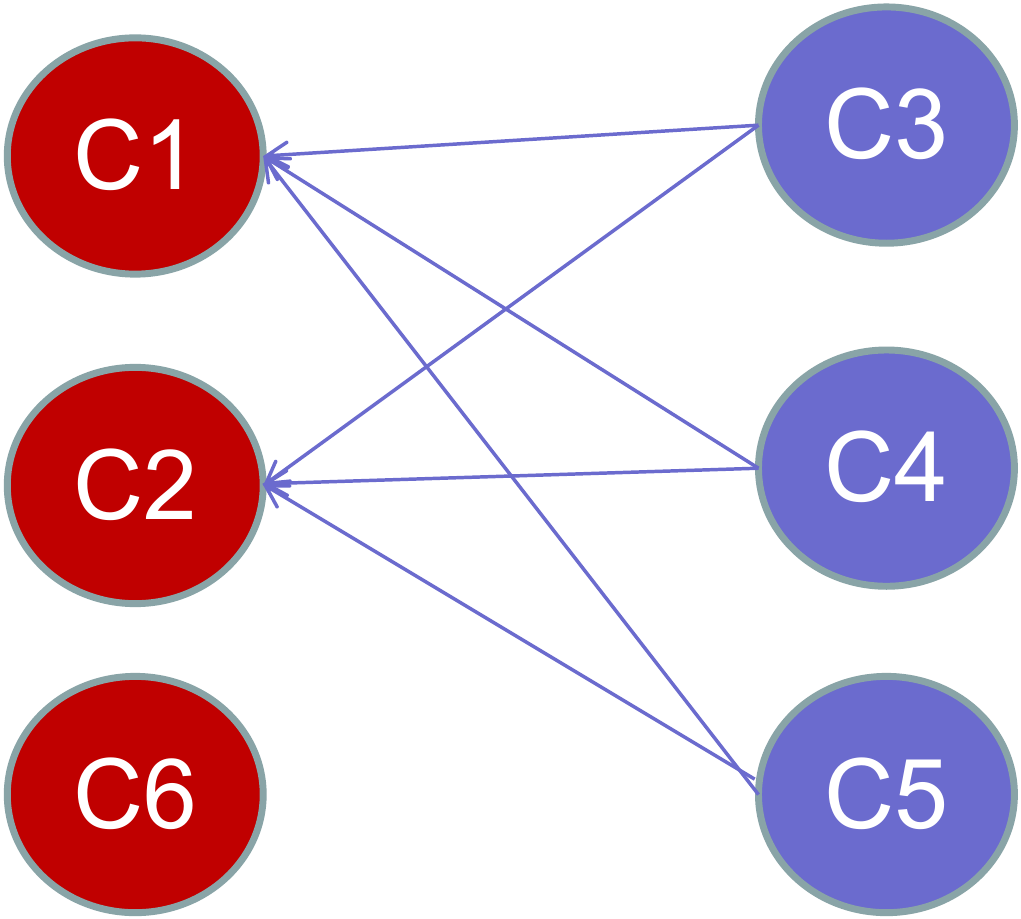}
		}
		\hspace{20pt}
		\subfigure[The edges of DEGM after the CCCOSCZC lifelong learning.]{
			\centering
			\includegraphics[scale=0.45]{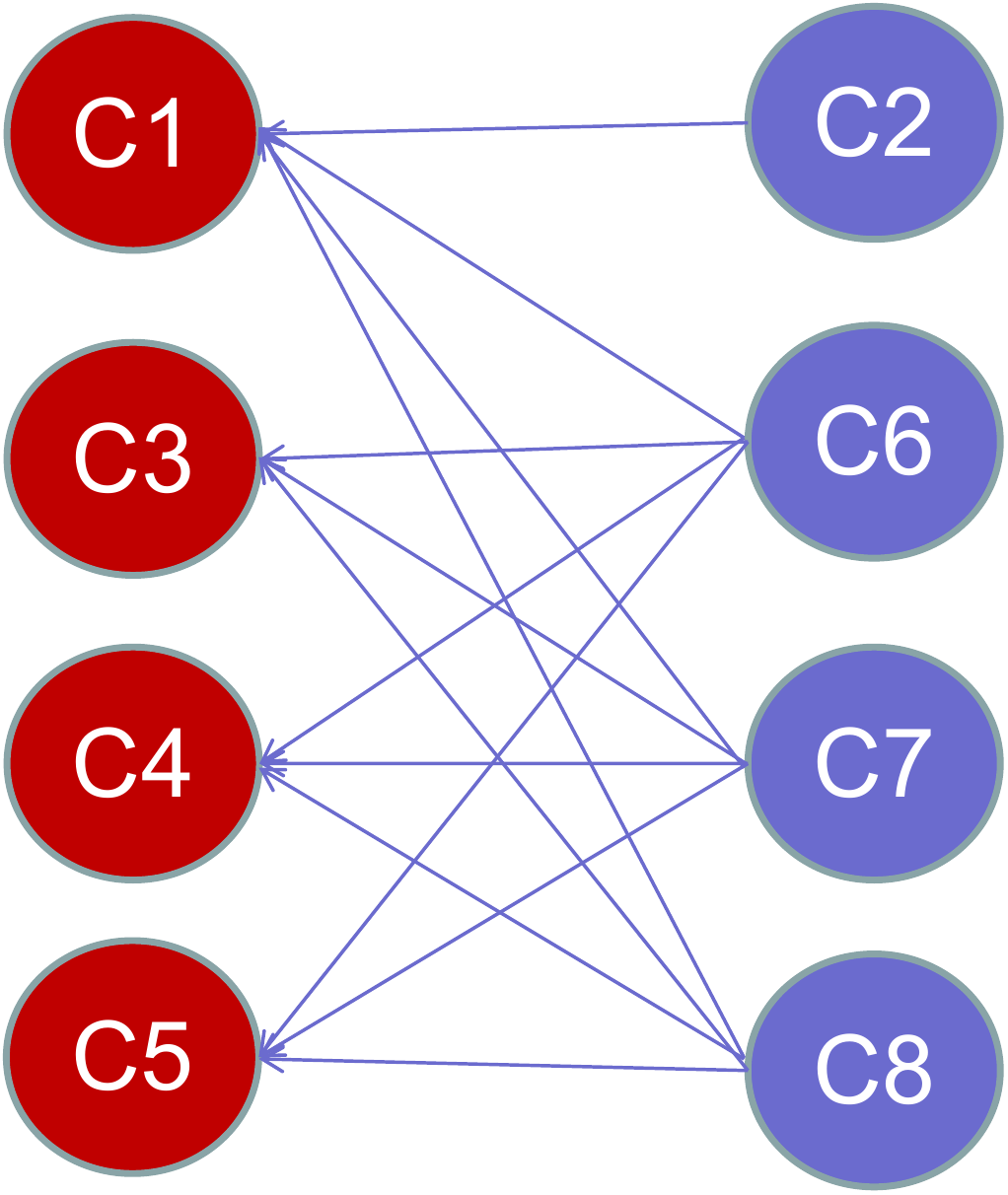}
		}
		\centering
		\caption{Edge information of DEGM after lifelong learning. The red and blue represent the basic and specific node, respectively.}
		\label{selectedResults2}
	\end{figure}
	
	\begin{table}[htb]
		\centering
		\begin{tabular}{lcccccc}
			\toprule
			Criteria
			& Dataset & BE  & LGM  &DEGM & DEGM-2 &CN-DPM* \\
			\midrule
			\multirow{7}{*}{SL} &MNIST &26.3&685.3&22.3 &22.3&21.9
			\\
			&SVHN &47.0&941.7&30.1 &29.0&39.3
			\\
			&Fashion &43.8&663.4&37.7 &27.4&36.6
			\\
			&IFashion &45.9&1148.4&35.6 &27.4&38.4
			\\
			&RMNIST &27.9&704.2&20.2 &22.1&25.3
			\\
			&Cifar10 &994.4&1241.1&615.3 &608.1&892.1
			\\
			&Average &197.5&897.4&126.9 &122.7&175.6
			\\
			\bottomrule
		\end{tabular}
		\vspace{2pt}
		\caption{The results under MSFIRC lifelong learning.}
		\label{unsupervised1}
	\end{table}
	
	\begin{table}[h]
		\centering
		\begin{tabular}{lcccccc}
			\toprule
			Criteria
			& Dataset & BE  & LGM &DEGM & DEGM-2 &CN-DPM* \\
			\midrule
			\multirow{7}{*}{SL}& CelebA &213.9&535.6&229.2&217.0&215.4
			\\
			&CACD &414.9&814.3&368.3&281.95&347.3
			\\
			&3D-Chair &649.1&2705.9&324.0&291.46&513.8
			\\
			&Omniglot &875.1&5958.9&225.6&195.7&343.2
			\\
			&ImageNet* &758.4&683.1&689.6&652.8&769.1
			\\
			&Car &745.1&583.7&588.8&565.9&709.8
			\\
			&Zappos &451.1&431.2&263.4&275.8&280.7
			\\
			&CUB &492.0&330.2&461.3&569.6&638.6
			\\
			&Average &575.0&1505.4&393.8&381.3&477.2
			\\
			\bottomrule
		\end{tabular}
		\vspace{2pt}
		\caption{The results under CCCOSCZC lifelong learning.}
		\label{unsupervised2}
		\vspace{10pt}
	\end{table}
	
	We show the adjacency matrix of the nodes from the graph ${\bf V}$ of DEGM after MSFIRC lifelong learning in Fig.~\ref{selectedResults}a where "C1" represents the first node. We can observe that DEGM creates three basic nodes and three specific nodes, respectively. Three specific nodes have edges from the first and second basic nodes. We also show ${\bf V}$ of DEGM after CCCOSCZC lifelong learning in Fig.~\ref{selectedResults}b. DEGM creates basic nodes when learning the first, third, fourth and fifth task. We also show the edge information between members of $\mathcal{S}$ and members of $\mathcal{G}$ in Fig.~\ref{selectedResults2} where red colour represents the basic nodes and blue colour represents the specific nodes.
	
	\subsection{Results for generalization bounds}
	\label{Appendix_Results_GB}
	
	To estimate the discrepancy, we train an auxiliary model on the distribution $\mathbb{P}^i \otimes {\tilde{\mathcal{P}}_{(i+1)}}$ for each $(i+1)$-th task learning. Then we calculate the discrepancy by using Definition 3 of the paper for each training epoch.
	
	In the following, we train a single model with GR under MNIST, Fashion and IFashion (MFI) lifelong learning. The images from all databases have pixel values within the range $[0,255]$. We implement the decoder by a neural network that outputs the mean vector of a Gaussian distribution with the diagonal covariance matrix (diagonal element is $1.0$). We estimate the reconstruction error term by using~:
	
	\begin{equation}
	\begin{aligned}
	{\log {p_\theta }\left( {{\bf x}\,|\,{\bf z}} \right)} =  - \frac{1}{{2\sigma  ^2}}{\left\| {{\bf{x}} - {\mu _\theta }\left( {\bf{z}} \right)} \right\|^2} - \frac{1}{2}\log 2\pi \sigma  ^2
	\end{aligned}
	\end{equation}
	
\begin{figure*}[h]
		\centering
		\subfigure[Evaluation on all tasks.]{
			\centering
			\includegraphics[scale=0.5]{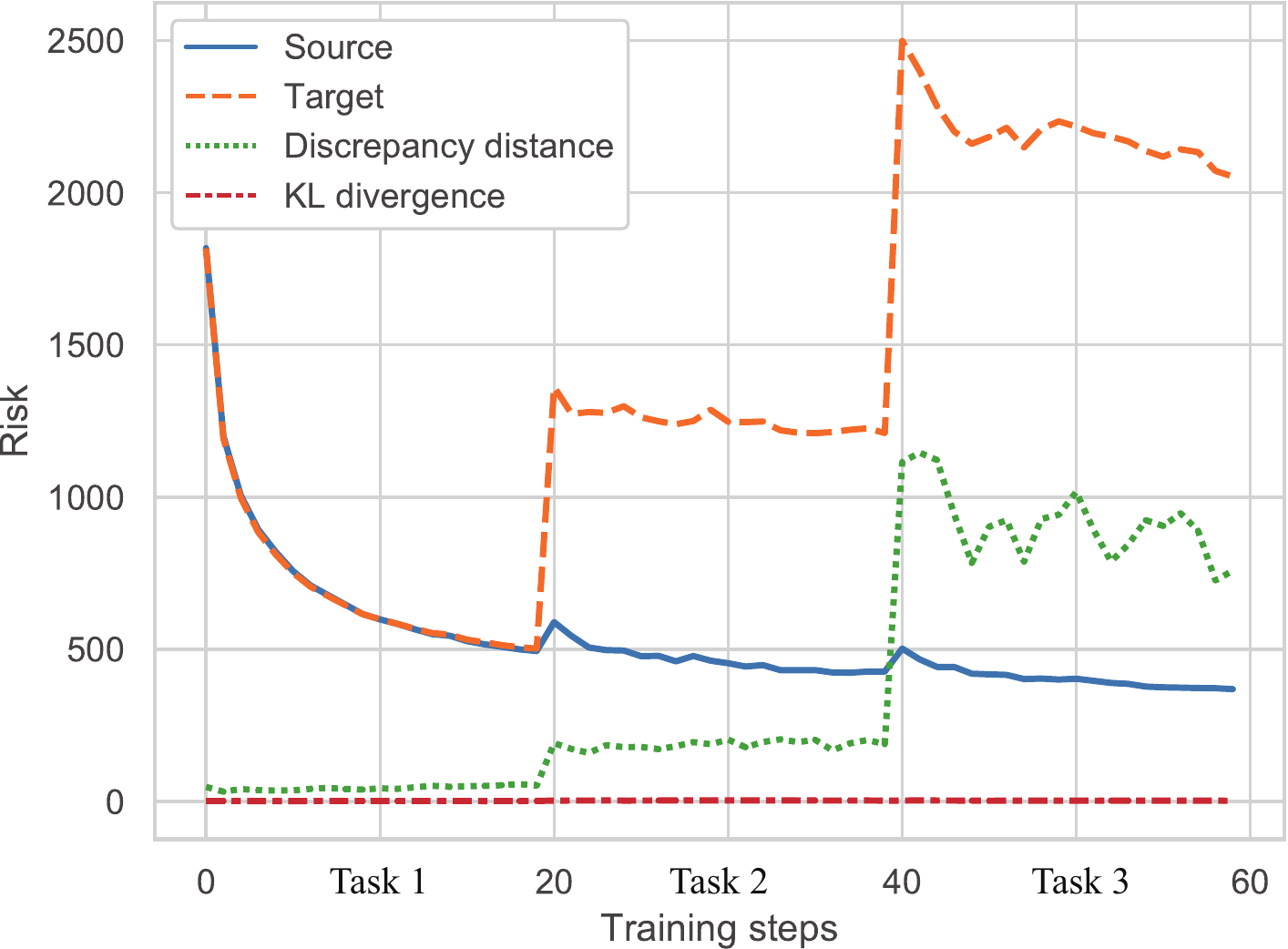}
		}
		\subfigure[Evaluation on MNIST]{
			\centering
			\includegraphics[scale=0.5]{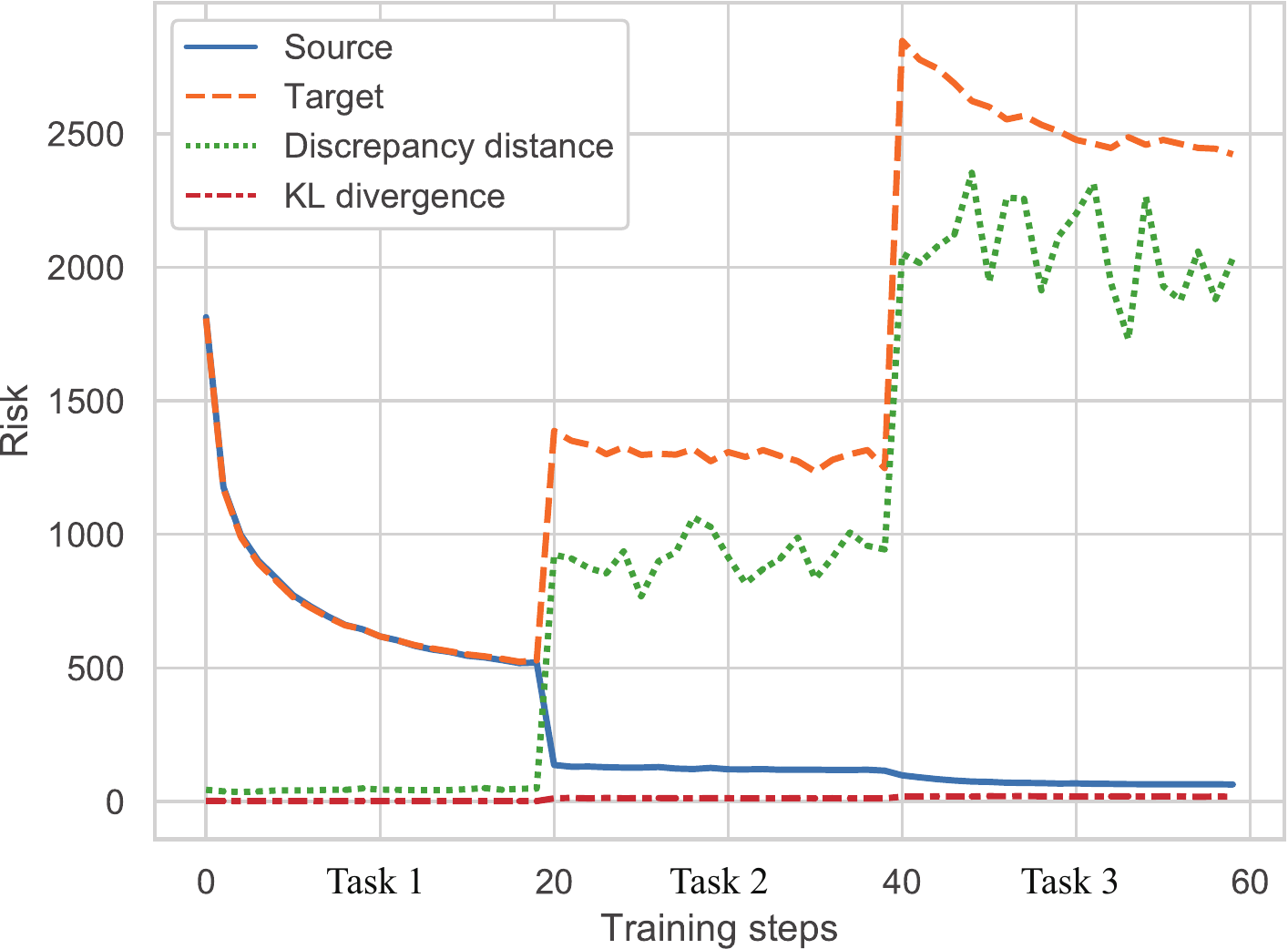}
		}
		\centering
		\caption{ The risk, $|KL_1 - KL_2|$ and discrepancy distance estimated by a single VAE model with GR under MFI lifelong learning. }
		\label{Theorem2_elbo_mse}
	\end{figure*}

	It can be noted that we also normalize this reconstruction error by dividing $28 \times 28$, as done in \cite{VLAE}. We evaluate the average risk and the discrepancy distance for each training epoch (See details in Lemma 1 from the paper). To calculate $|KL_1 - KL_2 |$ and discrepancy distance, we randomly choose 10000 number of samples from ${\mathcal{P}}_{(1:t)}$ and ${\mathbb P}^{t-1} \otimes {\tilde{\mathcal{P}}}_t$ at the $t$-th task learning, respectively. The results are presented in Fig.~\ref{Theorem2_elbo_mse}a where the source risk is continuously decreased while the discrepancy is increased as learning more tasks. We also evaluate the risk, $|KL_1 - KL_2|$ and discrepance distance on MNIST, shown in Fig.~\ref{Theorem2_elbo_mse}b. In order to select the generated images that are belonging to MNIST, we train a task-specific classifier that predicts the task label for giving samples. 
	
\begin{figure}[h]
		\centering
		\includegraphics[scale=0.5]{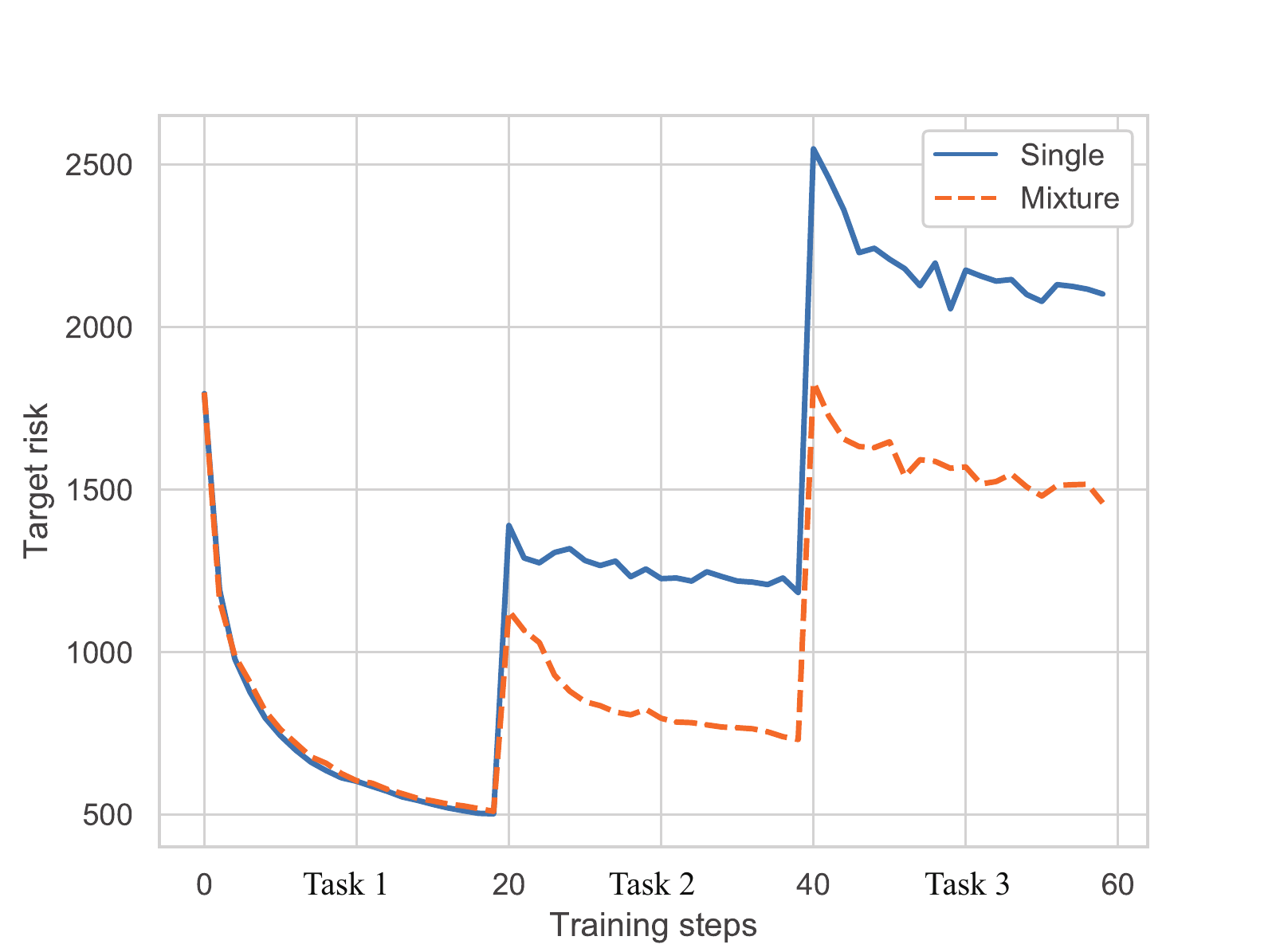}
		
		\centering
		\caption{ Target risk on the single model and the mixture model.  }
		\label{Lemma2_results}
	\end{figure}

	In the following, we investigate the results for Lemma 2 of the paper. We consider a sequence of MNIST, Fashion, IFashion (MFI) learning tasks. We consider a mixture model ${\bf M} = \{{\mathcal{M}}_1,{\mathcal{M}}_2  \}$ consisting of two components after LLL in which ${\mathcal{M}}_1$ is fixed after the first task learning while ${\mathcal{M}}_2$ is used to learn Fashion and IFashion, respectively. We also consider to learn a single model ${\mathcal{M}}$ with GR under MFI lifelong learning. We evaluate the average target risk (NLL estimated by ELBO) for each training epoch and the results are reported in Fig.~\eqref{Lemma2_results} where "single" and "mixture" represent ${\mathcal{M}}$ and ${\bf M}$, respectively. As shown from the results, the expansion mechanism can achieve a tight GB, as discussed in Lemma 2 of the paper.
	
	In the following, we provide additional empirical results for the theoretical analysis. It notes that in the following experiments, we resize all databases as $32 \times 32 \times 3$ and the pixel values of all images are normalized as $[0,1]$. 
	
\begin{figure*}[h]
		\centering
		\hspace{2pt}
		\subfigure[]{
			\centering
			\includegraphics[scale=0.35]{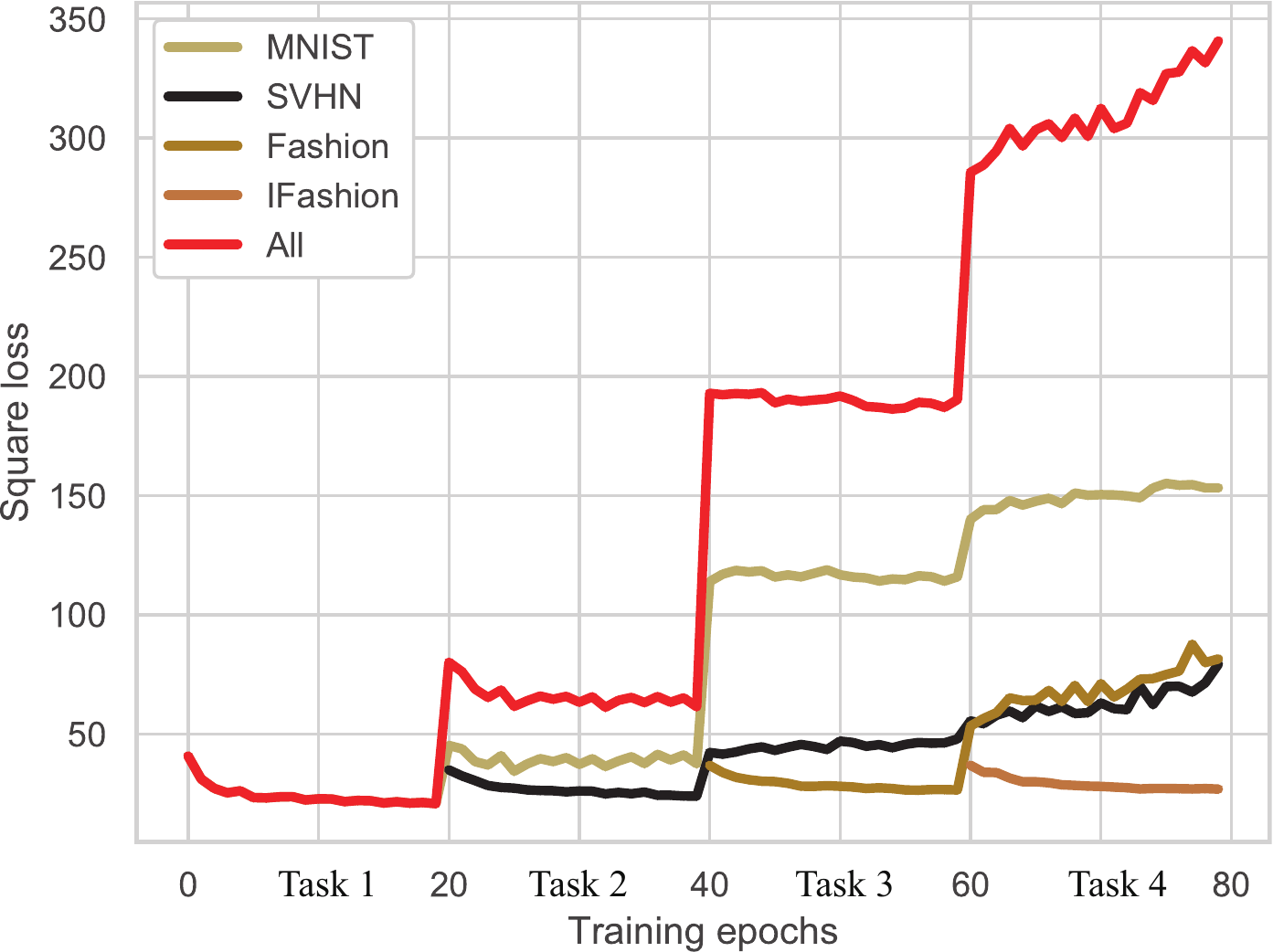}
		}
		\hspace{-10pt}
		\subfigure[]{
			\centering
			\includegraphics[scale=0.35]{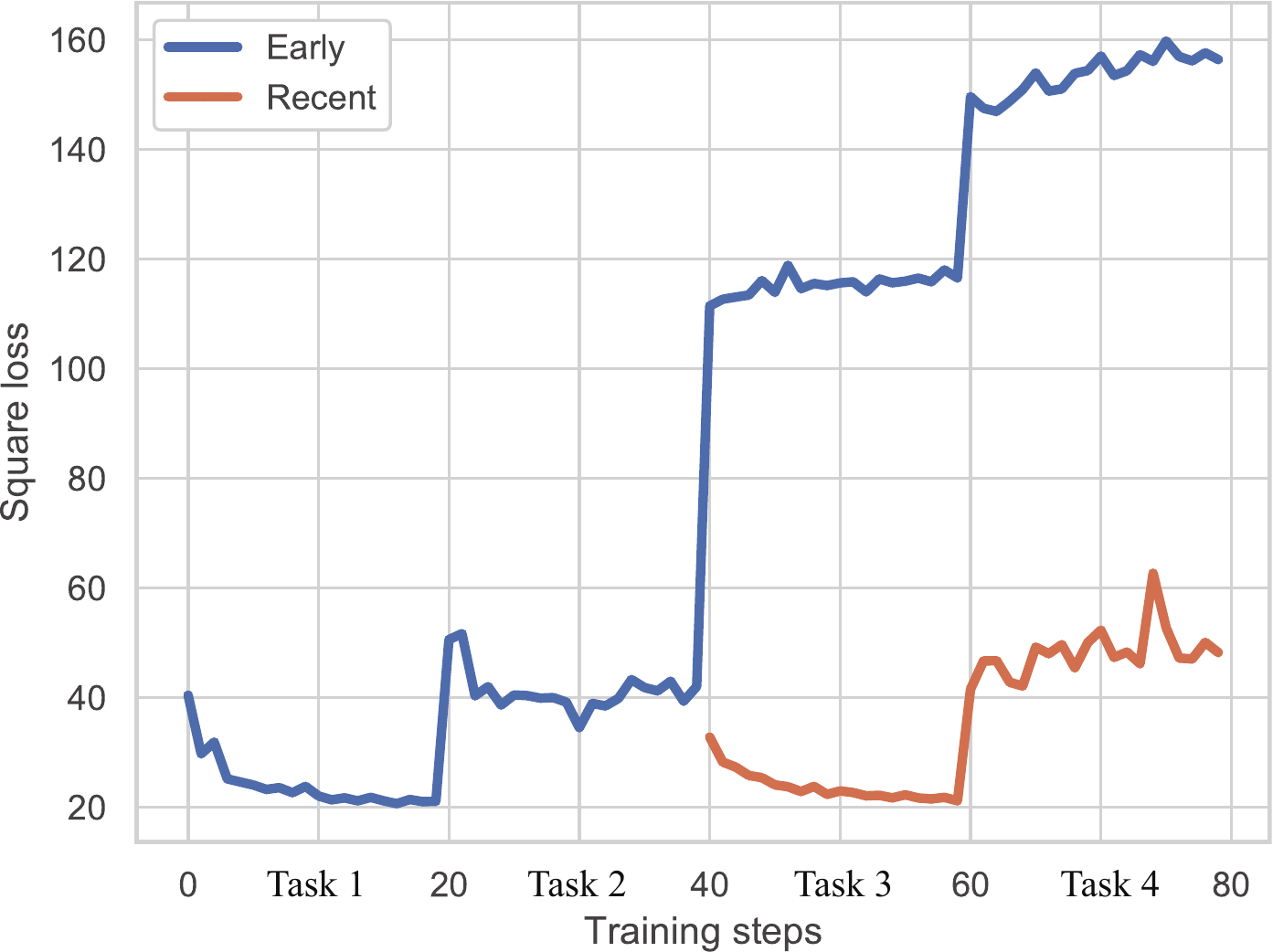}
		}
		\hspace{-10pt}
		\subfigure[]{
			\centering
			\includegraphics[scale=0.35]{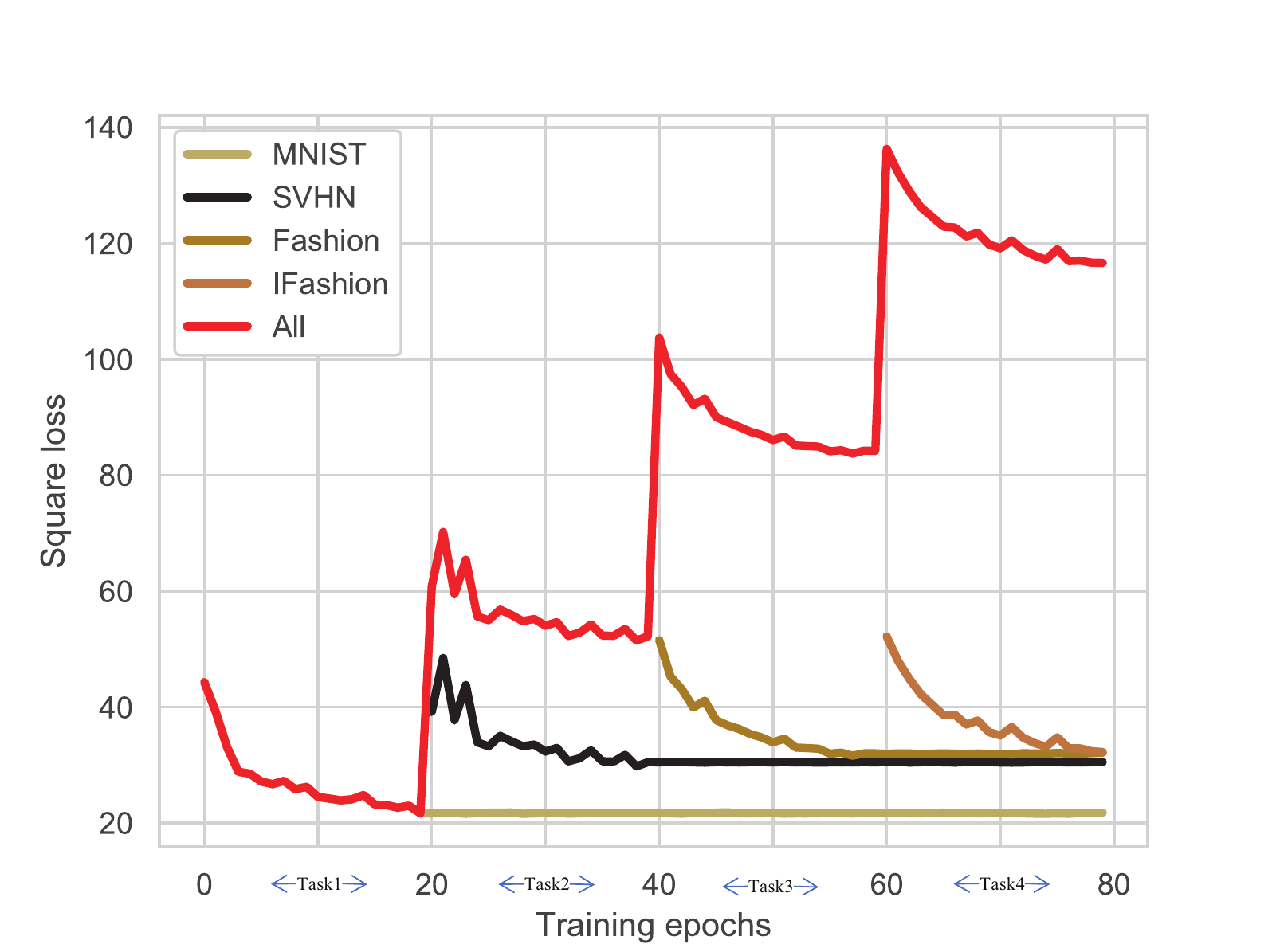}
		}
		\centering
		\caption{ "a" shows the target risk (LHS of Eq.(7) in Theorem-2 of the paper)) across four tasks under MSFI lifelong learning. "b" shows the target risk (LHS of Eq.(7) in Theorem-2 of the paper) on MNIST under MSFI lifelong learning.  "c" shows the target risk (LHS of Eq.(7) in Theorem-2 of the paper) estimated by the proposed mixture model under (MSFI) lifelong learning. It notes that all results are the accumulated target error and we do not calculate the average result.}
		\label{theoremEvaluation1}
		\vspace{10pt}
	\end{figure*}

	We evaluate the target risk (square loss) of a single model on four datasets (MNIST, SVHN, Fashion, IFashion) under MSFI (See Theorem-2). We plot the results in Fig.~\ref{theoremEvaluation1}-a where "All" represents the accumulated target risks for all tasks (left hand side of Equation-7 in the paper). In the next, we train a single model under MSFI and SFMI setting, respectively. We plot the target risk on MNIST in Fig.~\ref{theoremEvaluation1}-b where "Early" and "Recent" denote that the MNIST is used as the first and third task, respectively. It observes that the model tends to forget early tasks than recent tasks, as demonstrated in Equation-14 of the paper and discussed in Theorem 3. We also evaluate the accumulated target risk estimated by DEGM under MSFI lifelong learning, which is shown in Fig.~\ref{theoremEvaluation1}-c. It observes that there have no accumulated errors for each task during lifelong learning since the number of components match the number of task, as discussed in Theorem 3. 
	
\begin{figure*}[htbp]
		\centering
		\subfigure[]{
			\centering
			\includegraphics[scale=0.35]{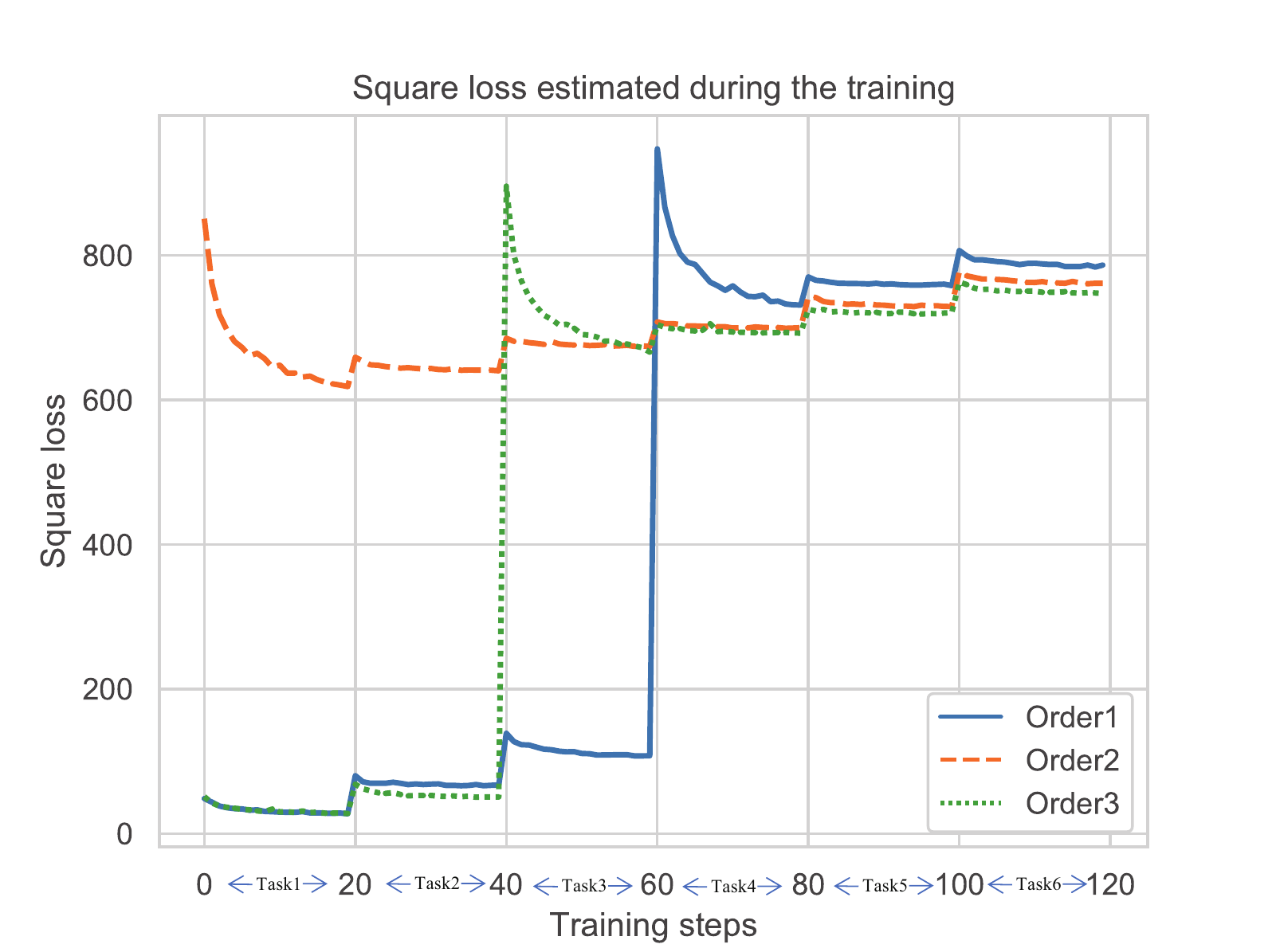}
		}
		\hspace{-10pt}
		\subfigure[]{
			\centering
			\includegraphics[scale=0.35]{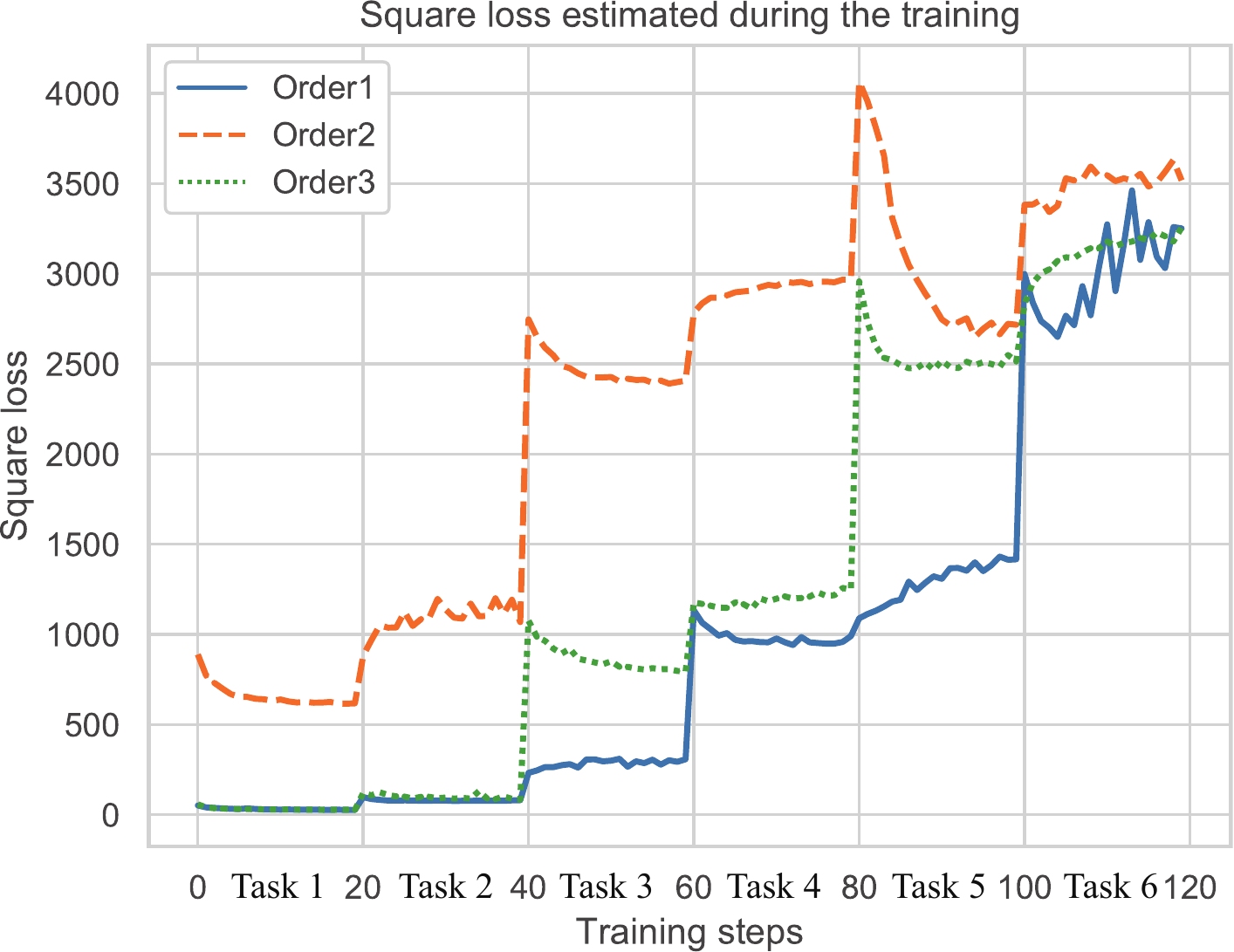}
		}
		\subfigure[]{
			\centering
			\includegraphics[scale=0.35]{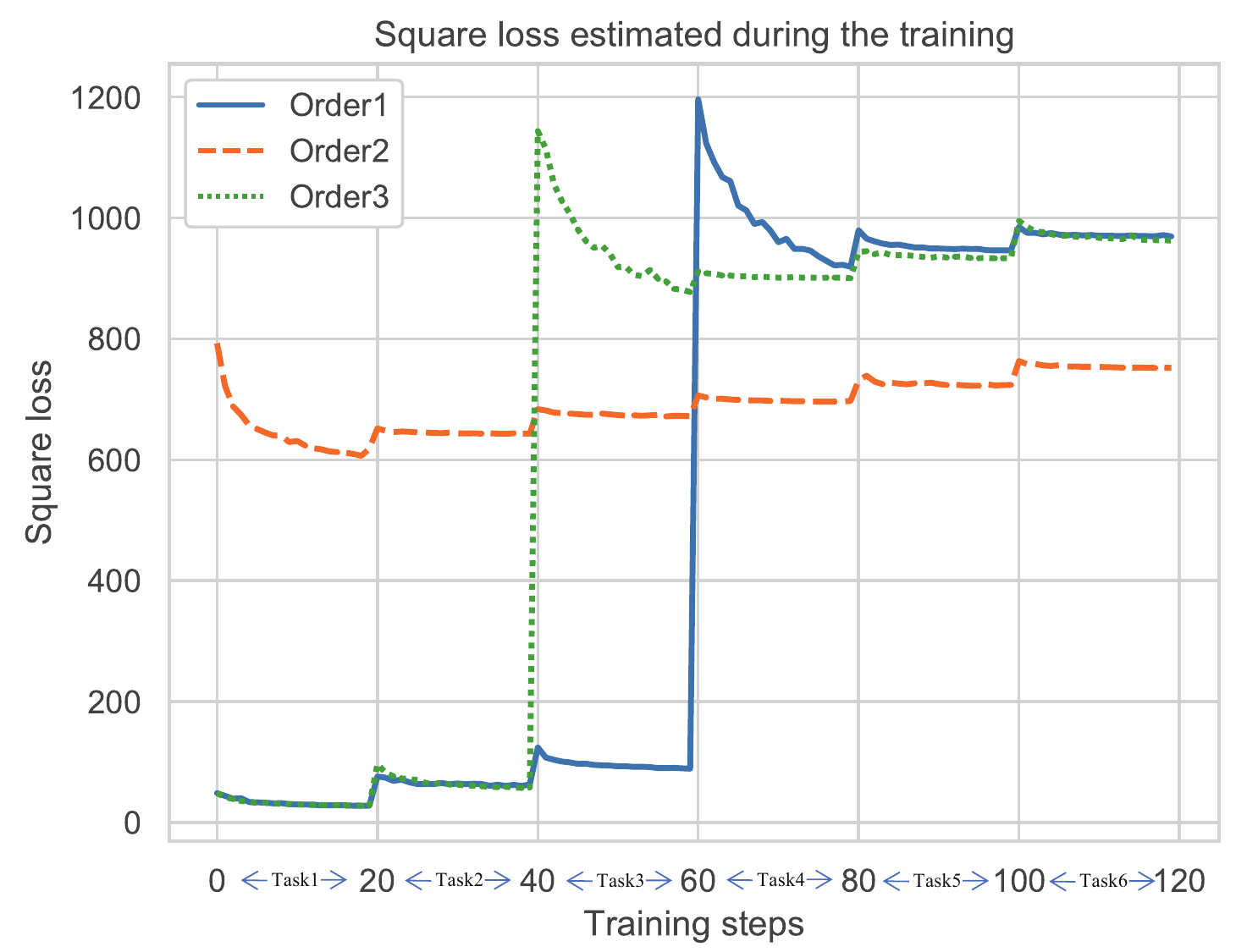}
		}
		\centering
		\caption{The accumulated target risks of DEGM, a single model and CN-DPM* with different orders of tasks. “a”, "b" and "c" represent the results achieved by DEGM, a single model and CN-DPM*, respectively.}
		\label{orderChanges}
	\end{figure*}

	In the following, we investigate the performance of DEGM and a single model when changing the order of tasks. We randomly generate three different orders~: Fashion, SVHN, MNIST, Cifar10, IMNIST, IFashion (FSMCII); Cifar10, IMNIST, Fashion, MNIST, SVHN, IFashion (CIFMSI); IFashion, IMNIST, Cifar10, Fashion, SVHN, MNIST (IICFSM). We evaluate the accumulated target risk for all tasks and the results are presented in Fig.~\ref{orderChanges} where "order1", "order2" and "order3" denote FSMCII, CIFMSI and IICFSM, respectively. It observes that a single model is sensitive to the choice of orders of tasks since the final accumulated target risk for all tasks are different when training the model under different orders of tasks. Although CN-DPM can avoid the forgetting during the training, the performance is still changed when changing the order of tasks. This is mainly because newly created components only reuse the transferable information from the first task and would lead to negative transfer when learning an entire different task. However, the proposed DEGM is robust to the change of orders of tasks.
	
	\clearpage
	\subsection{Evaluation by using other criterion}
	In addition to SL and NLL, we introduce to use the structural similarity index measure (SSIM) \cite{Reconstruction_criteria} and the Peak-Signal-to-Noise Ratio (PSNR) \cite{Reconstruction_criteria} as criteria. We report the results evaluated by the above criterion in Table~\ref{Unsupervised1} and Table~\ref{Unsupervised2}.
	
	\begin{table*}[h]
		\vspace{-5pt}
		\tiny
		\centering
		\begin{tabular}{c|ccccc|ccccc|ccccc}
			\toprule
			\multirow{2}{*}{Criteria} & \multicolumn{5}{c|}{SL} 
			&  \multicolumn{5}{c|}{SSMI}&
			\multicolumn{5}{c}{PSNR}
			\\
			& BE & LGM & DEGM & DEGM-2& CN-DPM*  
			& BE & LGM & DEGM & DEGM-2& CN-DPM*
			& BE & LGM & DEGM & DEGM-2& CN-DPM*
			\\
			\midrule
			MNIST &26.3&685.3&22.3 &22.3&21.9
			&0.88&0.19&0.90 &0.90&0.90
			&21.0&7.0&21.8 &21.8&21.8
			\\
			SVHN &47.0&941.7&30.1 &29.0&39.3
			&0.58&0.20&0.66 &0.67&0.61
			&13.7&5.0&15.5 &15.7&14.3
			\\
			Fashion&43.8&663.4&37.7 &27.4&36.6
			&0.68&0.15&0.72 &0.79&0.73
			&18.4&3.7&19.0 &20.6&19.2
			\\
			IFashion &45.9&1148.4&35.6 &27.4&38.4
			&0.72&0.11&0.76 &0.81&0.76
			&18.2&5.0&19.4 &20.6&19.1
			\\
			RMNIST &27.9&704.2&20.2 &22.1&25.3
			&0.87&0.20&0.91 &0.90&0.89
			&16.5&7.0&22.2 &21.8&21.2
			\\
			Cifar10 &994.4&1241.1&615.3 &608.1&892.1
			&0.29&0.23&0.49 &0.50&0.34
			&16.5&15.4&18.9 &18.9&17.0
			\\
			Average &197.5&897.4&126.9 &122.7&175.6
			&0.67&0.18&0.74 &0.76&0.70
			&18.1&7.2&19.5 &19.9&18.8
			\\
			\bottomrule
		\end{tabular}
		\caption{The performance of various models under the MSFIRC learning setting.}
		\label{Unsupervised1}
	\end{table*}
	
	\begin{table*}[h]
		\vspace{-5pt}
		\tiny
		\centering
		\begin{tabular}{c|ccccc|ccccc|ccccc}
			\toprule
			\multirow{2}{*}{Criteria} & \multicolumn{5}{c|}{SL} 
			&  \multicolumn{5}{c|}{SSMI}&
			\multicolumn{5}{c}{PSNR}
			\\
			& BE & LGM & DEGM & DEGM-2& CN-DPM*  
			& BE & LGM & DEGM & DEGM-2& CN-DPM*
			& BE & LGM & DEGM & DEGM-2& CN-DPM*
			\\
			\midrule
			CelebA &213.9&535.6&229.2&217.0&215.4
			&0.69&0.48&0.66&0.69&0.69
			&23.5&19.3&23.2&23.4&23.5
			\\
			CACD &414.9&814.3&368.3&281.95&347.3
			&0.57&0.47&0.62&0.68&0.63
			&20.6&17.33&21.2&22.4&21.4
			\\
			3D-Chair &649.1&2705.9&324.0&291.46&513.8
			&0.73&0.42&0.84&0.86&0.79
			&19.0&13.54&22.4&23.1&20.5
			\\
			Omniglot &875.1&5958.9&225.6&195.7&343.2
			&0.73&0.22&0.92&0.93&0.89
			&17.9&9.2&24.0&24.6&22.1
			\\
			Sub-ImageNet &758.4&683.1&689.6&652.8&769.1
			&0.37&0.42&0.41&0.43&0.37
			&18.5&18.9&19.0&19.2&18.5
			\\
			Car &745.1&583.7&588.8&565.9&709.8
			&0.39&0.48&0.47&0.49&0.42
			&18.0&19.0&19.0&19.2&18.2
			\\
			Zappos &451.1&431.2&263.4&275.8&280.7
			&0.68&0.60&0.75&0.74&0.73
			&20.0&20.2&22.4&22.3&22.1
			\\
			CUB &492.0&330.2&461.3&569.6&638.6
			&0.35&0.48&0.45&0.43&0.35
			&19.0&20.9&19.3&18.6&18.0
			\\
			Average &575.0&1505.4&393.8&381.3&477.2
			&0.60&0.45&0.64&0.66&0.61
			&19.6&17.3&21.3&21.6&20.5
			\\
			\bottomrule
		\end{tabular}
		\caption{The performance of various models under the CCCOSCZC learning setting.}
		\label{Unsupervised2}
	\end{table*}

	\subsection{Ablation study}
	
\begin{figure}[htbp]
		\centering
		\subfigure[]{
			\centering
			\includegraphics[scale=0.48]{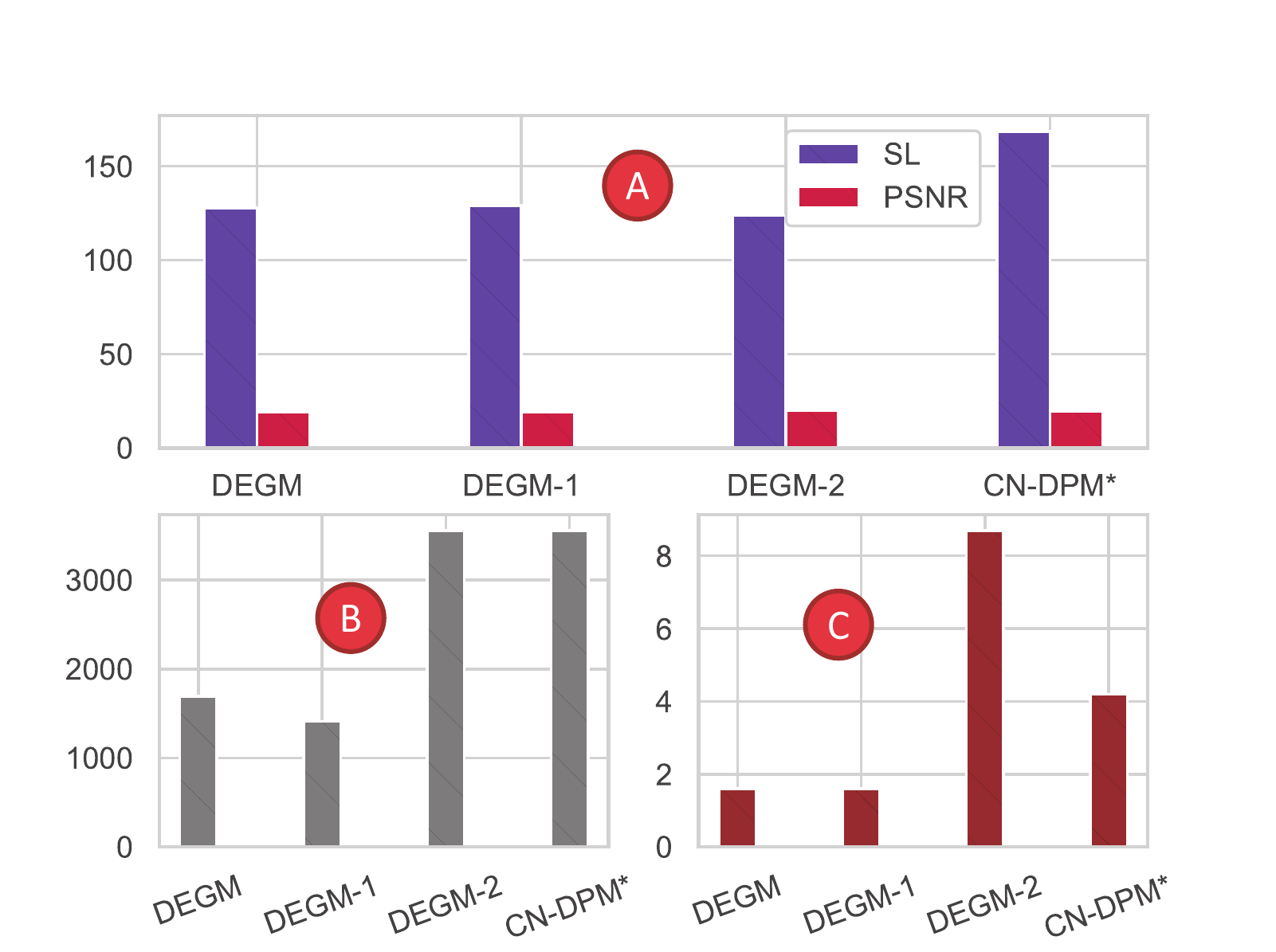}
		}
		\hspace{-12pt}
		\subfigure[]{
			\centering
			\includegraphics[scale=0.5]{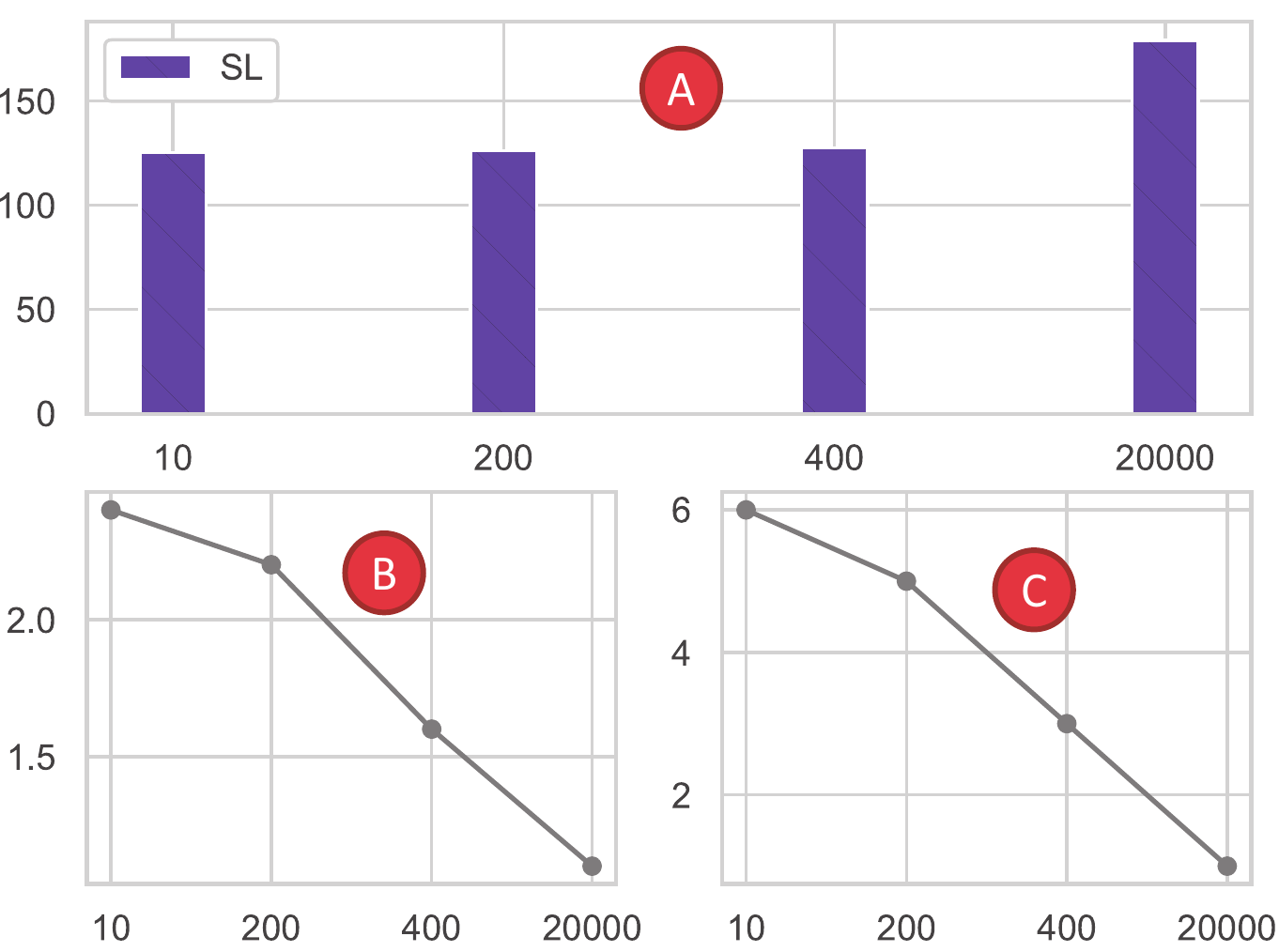}
		}
		\centering
		\caption{ The results for various baselines under MSFIRC. In the left chart, “A” represent the average square loss and PSNR for all tasks. “B” represent the overall training times (seconds). “C” represents the number of parameters ($10^8$) for various baselines. In the right chart, “A” represent the average square loss for all tasks. “B” represent number of parameters ($10^8$). “C” represents the number of basic components. X-axis represents different threshold values $thr$}
		\label{ablationStudy}
		\vspace{5pt}
	\end{figure}

	First, we evaluate the effectiveness of the proposed dynamic expansion mechanism used in our model. We train all models under MSFIRC lifelong learning and present the results in Fig~\ref{ablationStudy}a where we compare the average score, the training times and the memory use. It observes that even if the DEGM uses few training epochs but does not degenerate the performance. In addition, DEGM outperforms CN-DPM* that only transfers features from a single shared model. Furthermore, the proposed DEGM can achieve a similar performance as the baseline that trains individual VAEs for each task. 
	
	\noindent \textBF{The effects of thresholds in ORVAE} In the following, we investigate the performance of DEGM with different thresholds $\tau $ under MSFIRC lifelong learning and report results in Fig.~\ref{ablationStudy}b. As reduce $\tau$, DEGM tends to increase the number of basic components and gradually improve performance. It observes that the choice of $\tau = 400$ can trade-off between the performance and the model's complexity since it does not significantly improve the performance when decreasing $thr$ from 400.   
	In the next, in order to evaluate the effects of the proposed expansion mechanism and the adaptive weight, we introduce several new baselines in the following. 
	
	\textBF{DEGM-4:} This baseline generates information flows from all learned components to a new component. For instance, if a new component receives the information flow from members of $\mathcal{S}$, we will sum up the latent codes and intermediate representations from the sub-inference and sub-decoder of these components. DEGM-4 does not use the adaptive weight.
	
	\textBF{DEGM-5:} We implement this baseline by considering to create the edges without using the adaptive weight. The training process for this baseline is described as follows: Once the $t$-th task learning was finished, As similar done to DEGM, we have a set of $K$ measures, denoted by $KS = \{ k{s_1},\dots,k{s_K}\}$, which can be used to build the edges from a new node to members of $\mathcal{G}$. We set a threshold $\tau$ which is used to update $\bf{V}$ such that if each $ks_i < \tau$, then ${\bf{V}}(t+1,{\mathcal{GI}(i)} ) = 1$, otherwise ${\bf{V}}(t+1,{\mathcal{GI}(i)}) = 0$. This means that if $ks_i > \tau$, then the construction of a new component does not reuse the information and parameters from the $i$-th component in DEGM. 
	
	\textBF{DEGM-6:} For this baseline, we consider the adaptive weight for each edge is equal. This means that the importance of each basic component is treated as the same for a new task.
	
	\textBF{DEGM-7:} We implement this baseline by considering to create only a single edge for a new component to a certain basic component that has the maximum sample log-likelihood for data of the new task. 
	
	\textBF{CN-DPM*-1;} This baseline builds a new components and creates connections with previously learned components, as similar to DEGM-4.
	
	\textBF{CN-DPM*-2:} We implement this baseline by using the large model which contains $1.3 \times 10^9$ number of parameters.
	
	We report the results in Table~\ref{unsupervised_4}. It observes that the adaptive weights in the expansion mechanism can further improve the performance. We also find that although CN-DPM*-2 uses the more parameters, our DEGM still outperforms CN-DPM* by a large margin.
	
	\begin{table}[h]
		\centering
		\begin{tabular}{lcccccccc}
			\toprule
			Criteria
			& Dataset & DEGM  &DEGM-4&CN-DPM*-1&CN-DPM*-2&DEGM-5&DEGM-6&DEGM-7  \\
			\midrule
			\multirow{7}{*}{SL} &MNIST &22.30&22.18&22.12&22.67&22.88&21.48&22.35
			\\
			&SVHN &30.18 &30.73&40.53&38.74&30.56&29.44&29.20
			\\
			&Fashion &37.73&41.22&45.03&38.51&37.65&37.26&41.29
			\\
			&IFashion &35.62&49.26&36.19&36.90&41.27&37.15&43.95
			\\
			&RMNIST &20.23&60.72&24.79&24.39&27.86&25.73&25.78
			\\
			&Cifar10 &615.34&610.38&929.55&877.35&617.34&617.36&614.48
			\\
			&Average &\textBF{ 126.90}&135.58&183.03&173.09&129.59&128.07&129.51
			\\
			\bottomrule
		\end{tabular}
		\caption{The results of various models under MSFIRC lifelong learning.}
		\label{unsupervised_4}
	\end{table}
	
	\subsection{The number of parameters used in various methods}
	
	We list the number of parameters used in various methods in Table~\ref{Unsupervised_1} and Table~\ref{Unsupervised_2}, respectively. We can observe that the proposed DEGM architecture requires less parameters than other parameters.
	
	\begin{table}[ht]
		\centering
		\begin{tabular}{l|cccccc}
			\hline
			Model & LGM  &BE   & DEGM& DEGM-2&CN-DPM*  \\
			\hline
			N &${1.5 \times 10^8}$ &${9.4 \times 10^8}$& ${1.6 \times 10^8}$&${8.7 \times 10^8}$&${4.2 \times 10^8}$ \\
			\hline
		\end{tabular}
		\caption{The number of parameters of various models under MSFIR learning setting.}
		\label{Unsupervised_1}
	\end{table}
	
	\begin{table}[ht]
		\centering
		\begin{tabular}{l|cccccc}
			\hline
			Model & LGM  &BE   & DEGM& DEGM-2&CN-DPM* &LIMix  \\
			\hline
			N &${1.9 \times 10^9}$ &${3.9 \times 10^9}$&${3.2 \times 10^8}$ &${1.3 \times 10^9}$&${9.4 \times 10^9}$&${9.4 \times 10^9}$ \\
			\hline
		\end{tabular}
		\caption{The number of parameters of various models under CCCOSCZC learning setting.}
		\label{Unsupervised_2}
	\end{table}
	
	\newpage
	\subsection{Visual results}
	
\begin{figure}[htbp]
		\vspace{-10pt}
		\centering
		\subfigure[MNIST.]{
			\centering
			\includegraphics[scale=0.25]{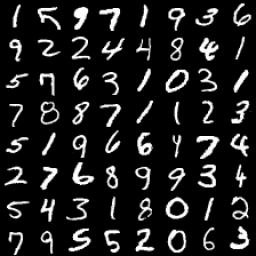}
		}
		\subfigure[SVHN.]{
			\centering
			\includegraphics[scale=0.25]{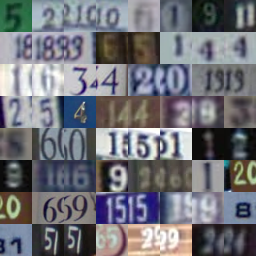}
		}
		\subfigure[Fashion.]{
			\centering
			\includegraphics[scale=0.25]{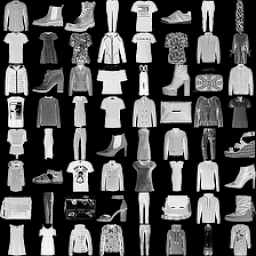}
		}
		\subfigure[IFasion.]{
			\centering
			\includegraphics[scale=0.25]{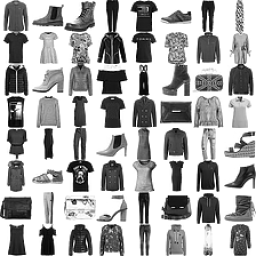}
		}
		\subfigure[RMNIST.]{
			\centering
			\includegraphics[scale=0.25]{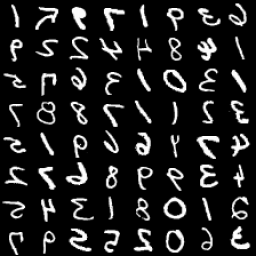}
		}
		\subfigure[Cifar10.]{
			\centering
			\includegraphics[scale=0.25]{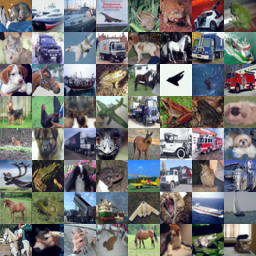}
		}
		\centering
		\subfigure[Task 1.]{
			\centering
			\includegraphics[scale=0.25]{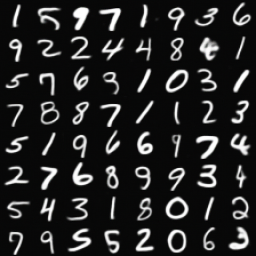}
		}
		\subfigure[Task 2.]{
			\centering
			\includegraphics[scale=0.25]{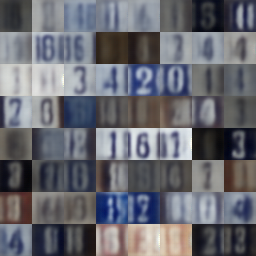}
		}
		\subfigure[Task 3.]{
			\centering
			\includegraphics[scale=0.25]{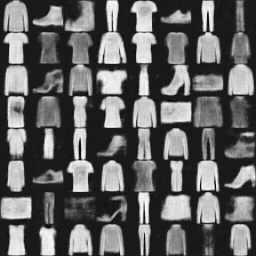}
		}
		\subfigure[Task 4.]{
			\centering
			\includegraphics[scale=0.25]{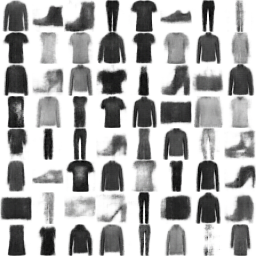}
		}
		\subfigure[Task 5.]{
			\centering
			\includegraphics[scale=0.25]{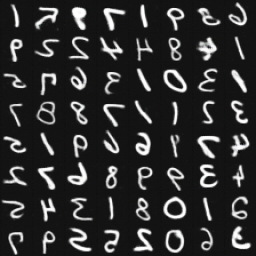}
		}
		\subfigure[Task 6.]{
			\centering
			\includegraphics[scale=0.25]{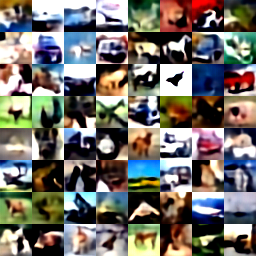}
		}
		\centering
		\caption{ Image reconstructions when using DEGM under the MSFIRC lifelong learning. The first row represents testing data samples and the second row are their reconstructions using DEGM. }
		\label{Reco_MSFIR}
	\end{figure}

\begin{figure}[htbp]
		\vspace{-10pt}
		\centering
		\subfigure[Task1.]{
			\centering
			\includegraphics[scale=0.3]{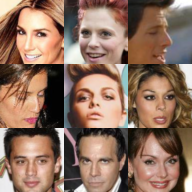}
		}
		\subfigure[Task2.]{
			\centering
			\includegraphics[scale=0.3]{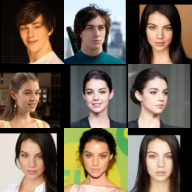}
		}
		\subfigure[Task3.]{
			\centering
			\includegraphics[scale=0.3]{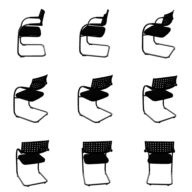}
		}
		\subfigure[Task4.]{
			\centering
			\includegraphics[scale=0.3]{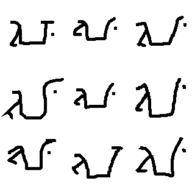}
		}
		\subfigure[Task5.]{
			\centering
			\includegraphics[scale=0.3]{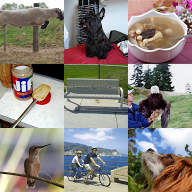}
		}
		\subfigure[Task6.]{
			\centering
			\includegraphics[scale=0.3]{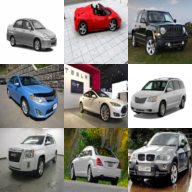}
		}
		\centering
		\subfigure[Task7.]{
			\centering
			\includegraphics[scale=0.3]{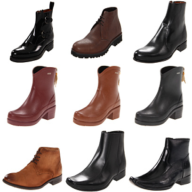}
		}
		\subfigure[Reconstruction of Task1.]{
			\centering
			\includegraphics[scale=0.3]{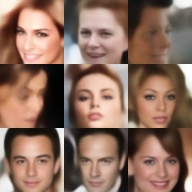}
		}
		\subfigure[Reconstruction of Task2.]{
			\centering
			\includegraphics[scale=0.3]{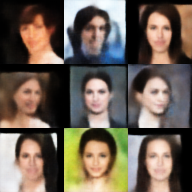}
		}
		\subfigure[Reconstruction of Task3.]{
			\centering
			\includegraphics[scale=0.3]{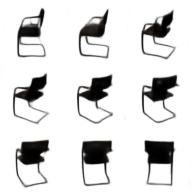}
		}
		\subfigure[Reconstruction of Task4.]{
			\centering
			\includegraphics[scale=0.3]{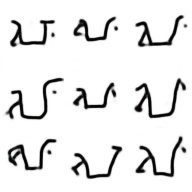}
		}
		\subfigure[Reconstruction of Task5.]{
			\centering
			\includegraphics[scale=0.3]{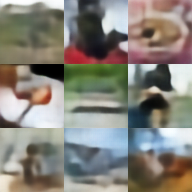}
		}
		\subfigure[Reconstruction of Task6.]{
			\centering
			\includegraphics[scale=0.3]{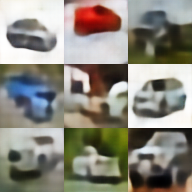}
		}
		\subfigure[Reconstruction of Task7.]{
			\centering
			\includegraphics[scale=0.3]{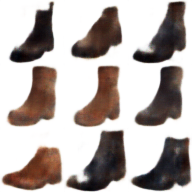}
		}
		\subfigure[Task8.]{
			\centering
			\includegraphics[scale=0.3]{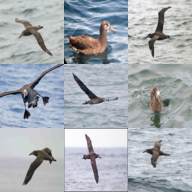}
		}
		\subfigure[Reconstruction of Task8.]{
			\centering
			\includegraphics[scale=0.3]{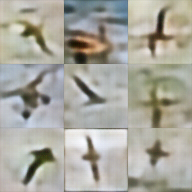}
		}
		\centering
		\caption{ Image reconstructions when using DEGM under the CCCOSCZC lifelong learning. The first row represents testing data samples and the second row are their reconstructions using DEGM. }
		\label{Reco_CCCOSCZC}
	\end{figure}

	We show the reconstructions from DEGM under MSFIRC and CCCOSCZC lifelong learning in Fig.~\ref{Reco_MSFIR} and Fig.~\ref{Reco_CCCOSCZC}, respectively.
	
	\clearpage

	
\end{document}